\documentclass{article}

\usepackage{microtype}
\usepackage{graphicx}
\usepackage{booktabs} 
\usepackage{multirow}
\usepackage{amsmath}
\usepackage{amssymb}  

\usepackage{tcolorbox}
\tcbuselibrary{listings}
\tcbuselibrary{skins} 
\usepackage{algorithmic}
\usepackage{subfig}
\usepackage{wrapfig}
\usepackage{pifont}
\usepackage{dsfont}

\usepackage{colortbl}  
\usepackage{xcolor}

\usepackage{hyperref}
\usepackage{enumitem}

\renewcommand{\algorithmiccomment}[1]{\hfill// #1}

\usepackage[accepted]{mlsys2025}

\mlsystitlerunning{Submission and Formatting Instructions for MLSys 2025}

\begin{document}

\twocolumn[
\mlsystitle{AERO: Entropy-Guided Framework for Private LLM Inference}





\mlsyssetsymbol{equal}{*}

\begin{mlsysauthorlist}
\mlsysauthor{Nandan Kumar Jha}{ed}
\mlsysauthor{Brandon Reagen}{ed}
\end{mlsysauthorlist}

\mlsysaffiliation{ed}{New York University (NYU)}

\mlsyscorrespondingauthor{Nandan Kumar Jha}{nj2049@nyu.edu}

\mlsyskeywords{Machine Learning, MLSys}

\vskip 0.3in

\begin{abstract}
Privacy-preserving computation enables language model inference directly on encrypted data yet suffers from prohibitive latency and communication overheads, primarily due to nonlinear functions. Removing nonlinearities, however, can trigger one of two failure modes restricting the potential for nonlinearity removal: entropy collapse in deeper layers, which destabilizes training, and entropic overload in early layers, causing under-utilization of attention heads. To address these challenges, we introduce AERO, an entropy-guided framework  to strategically eliminates costly nonlinear operations from transformer architectures, which employs an adaptive recalibration through a head-wise entropy regularizer with learnable per-head strengths, enabling each head to adjust its entropy level while penalizing extreme entropies and fostering functional diversity through a tolerance margin. Experiments show AERO can save 3.4$\times$ communication and 1.4$\times$  latency,  without any performance penalty. 
\end{abstract}

]



\printAffiliationsAndNotice{\mlsysEqualContribution} 

\section{Introduction}

The popularity of large language model (LLMs) has raised privacy concerns as more sensitive user data (e.g., prompts) are sent to the cloud \cite{staab2024beyond,mireshghallah2024can,priyanshu2023chatbots,ChatGptWiredArticle}. 
Private inference (PI) leverages cryptography to allow users to interact with service providers' models without revealing inputs, ensuring data privacy and model weight protection \cite{carlini2024stealing}. However, deploying private transformer-based LLMs is challenging due to prohibitive communication and latency overheads. For instance, BOLT \citep{pang2023bolt} requires 59.6 GB for a single private inference on BERT-base model, while Bumblebee \citep{lu2023bumblebee}, despite leveraging homomorphic encryption (HE) for nonlinear operations, still requires 15.5 GB for BERT-large. 

Moreover, our experimental study, performed using Bumblebee  private LLM inference framework, showed that generating one output token in a private GPT-2 (125M) inference takes 8.2 minutes and 25.3 GBs of communication (Table \ref{tab:GPT2CLen128}) with 128 input tokens,
and at 512 inputs 30.7 minutes and 145.2 GBs of communication (Table \ref{tab:LanguiniGPT2}).

\begin{figure} [t]
\centering
\includegraphics[width=.49\textwidth]{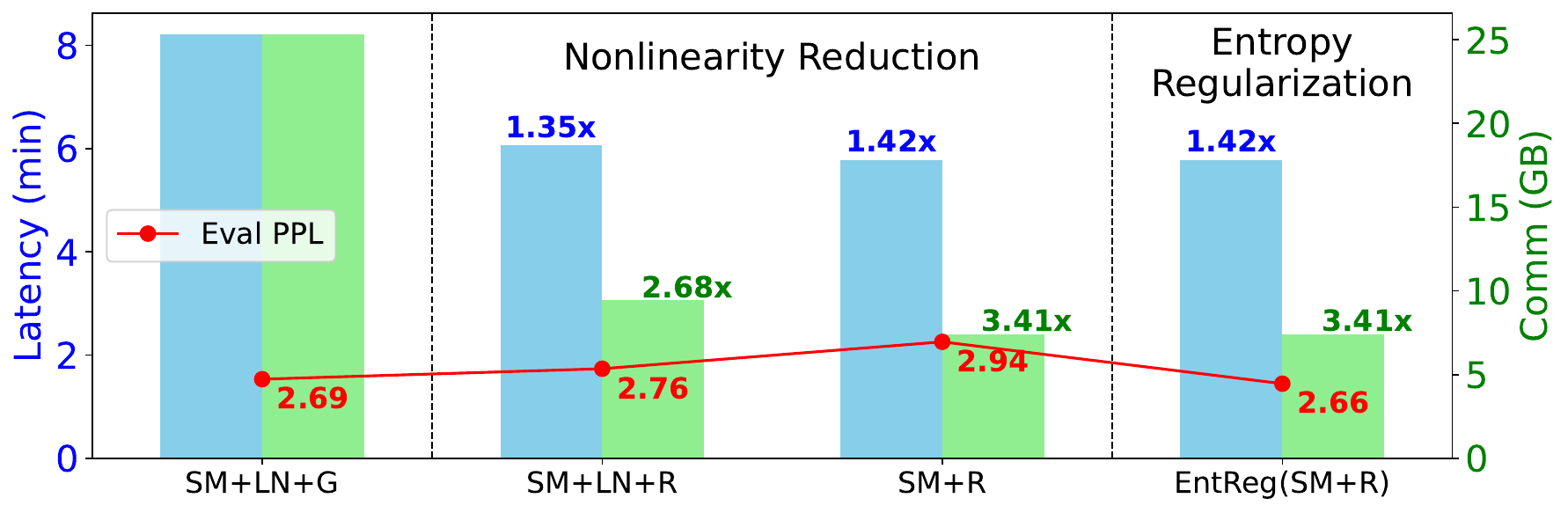}  \vspace{-2.5em}
\caption{Latency and communication savings through nonlinearity reduction, and performance improvement through entropy regularization when AERO is applied on GPT-2 (125M),  trained from scratch on CodeParrot dataset (detailed breakdown in Table \ref{tab:GPT2CLen128}).} 
\label{fig:GPT2vsSOTA}
\end{figure}


Recent work \cite{lu2023bumblebee,hou2023ciphergpt} attribute these overheads primarily to the expensive \emph{nonlinear} operations (e.g., GELU and LayerNorm). While these operations are crucial for model stability and performance, they rely on expensive cryptographic primitives with require numerous rounds of interaction, leading to disproportionate latency and communication costs. For instance, a {\em single} GELU operation in BERT-base model requires 3.9$\times10^6$ operations, each involving multiple secure multiplications and communication rounds,  adding 1-2 KB per operation  \cite{lu2023bumblebee}. 
Thus, GELU and LayerNorm contribute 59\% of communication and 49\% of latency in private GPT-2  inference with 256 input tokens \cite{hou2023ciphergpt}.

Designing LLMs with fewer nonlinear operations offers a promising direction for enabling cryptographically secure private LLM inference. However, this
requires a deeper understanding of the fundamental role that nonlinearities in shaping learning dynamics and internal representations---which remains largely unexplored. The state-of-the-art  private LLM systems optimize for how efficiently inference is performed under cryptography settings {\em rather} than which nonlinearity the model requires. Specifically, THOR \citep{moon2024thor} accelerates HE execution using diagonal-major packing and tailored the protocols for nonlinear operations, NEXUS \citep{zhang2024secure} makes PI non-interactive under CKKS FHE, and BLB \citep{xu2025breaking} reduces the number of truncations and conversion in a hybrid system by fine-grained operator splitting and fusion.

Nonetheless, some recent studies have examined the  LLM failure modes such as training instability \cite{wortsman2024smallscale,rybakov2024methods} and rank collapse \cite{bao2024self,noci2022signal}. 
However, the critical questions remain, specifically:
(i) How does removing LLM nonlinearities affect learning dynamics and information flow? 
(ii) Do we need all types of nonlinearity or can we use only those vital, e.g., Softmax for self-attention?


To address this, we introduce an information-theoretic framework to systematically analyze the role of nonlinearities in transformer-based LLMs.
Using Shannon's entropy as a quantitative lens, we uncover the {\bf dual} significance of  nonlinearities: 
(i) they preserve the representational diversity in attention heads, particularly in early layers, fostering head-wise specialization; and
(ii) maintain stable training by preventing entropy collapse in deeper layers.

We observed a new phenomenon, {\em entropic overload}, where many attention heads remain stuck in greater entropy states during training when {\em crucial} nonlinearities are removed. 
Our empirical results suggest that entropic overload disrupts information flow, impairing the model's ability to learn diverse representations and  increases perplexity (Table \ref{tab:NormFreePythiaGPT}).

To mitigate entropic overload, we propose an entropy-guided attention mechanism paired with a novel entropy regularization scheme that strategically penalizes the deviation from well-behaved entropy distribution through two key innovations: 
(i) {\em Headwise learnable thresholds} to dynamically adjust regularization strength for each attention head;
(ii) {\em Tolerance margins} to prevent over-regularization by allowing only a small fraction of attention heads to attain higher entropy values.
These assure diversity while controlling excessive penalization in the absence of nonlinearities.

Furthermore, we establish  the effectiveness of PI-friendly static alternatives to LayerNorm, such as weight and spectral normalization \cite{salimans2016weight,miyato2018spectral}, which do not requires nonlinear operation at inference. Unlike stabilization methods that heavily rely on LayerNorm layers, such as QK-LayerNorms \cite{dehghani2023scaling,wortsman2024smallscale,muennighoff2024olmoe} or the positional variants like MixLN \citep{li2025mixln}, static normalization methods maintain training stability by preventing entropy collapse in deeper layers, and avoid the cost of inverse square-root operation at inference.

To this end, we summarize our {contributions} as follows. 
\begin{itemize} [noitemsep,nolistsep,leftmargin=0.5cm] \vspace{-1em}
\item {\bf Conceptual Insights}: Using Shannon’s entropy as a quantitative lens, we analyze how nonlinearities regulates internal information flow and stabilize training in transformer-based LLMs, offering design insights for tailoring existing LLM architecture for efficient PI (\S \ref{sec:NonlinearCharacterization}).
\item {\bf Framework}: We propose AERO, a three-stage design framework that systematically reduces nonlinearities in LLMs while maintaining training stability, showing that models remain trainable even with Softmax as the only nonlinearity (\S \ref{sec:AERO}).
\item {\bf Architectural}: We explore static alternatives to LayerNorm which effectively prevent entropy collapse and maintain training stability without incurring the overheads of nonlinear operations at inference (\S \ref{subsec:StaticNorm}). 
\item {\bf Algorithmic}: We introduce an entropy-guided attention mechanism paired with a novel entropy regularization scheme which prevents entropic overload in the absence of nonlinearities. We use entropy as an actionable cause and penalize extreme entropy values while adapting the  regularization strength to the head-specific role (\S \ref{subsec:entropyreg}).
\item {\bf Empirical}: We demonstrate the scalability and efficacy of AERO across context sizes (128, 256, 512), model depths (12 and 18 layers), and training scales (1.2B to 4.8B tokens) on the CodeParrot and Languini book datasets (\S \ref{sec:evaluation}). AERO achieves  {\bf 3.4}$\times$ communication and {\bf 1.42}$\times$ latency reduction with no loss in performance (Figure \ref{fig:GPT2vsSOTA}). Even under the most-stringent stringent nonlinearity setting, where Softmax is the only remaining nonlinearity, entropy regularization improves perplexity by {\bf 6\%}–{\bf 8\%}. We also conducted extensive benchmarking against state-of-the-art method \cite{he2024simplifying}.
\end{itemize}



\section{Related Work}

{\bf Role of nonlinearity in LLMs}
Prior work used a simplified one-layer transformer architecture---a single Softmax-based self-attention head and a ReLU-based feed-forward network(FFN)---to investigate the role of attention and FFN nonlinearity for in-context learning tasks~\cite{linonlinear}. 
Other explores a broader range of nonlinear architectures and in-context learning tasks, showing that the optimal activation function can vary depending on the specific function class the model is attempting to learn~\cite{cheng2023transformers}. 

To the best of our knowledge, the most closely related work is by \citet{brody2023expressivity}, which showed the LayerNorm's geometric regularization effect on attention expressivity. However, their analysis is specific to LayerNorm, while our study takes a more holistic approach---quantifying how each nonlinearity affects the entropy dynamics of attention mechanism using a unified ablation framework.

{\bf Simplifying nonlinearity in transformer models} 
Existing approaches to simplifying nonlinearities in transformer models generally follow two main strategies.
The first centers on architectural optimizations, particularly the design of normalization-free models. While these models remove LayerNorms, they retain nonlinearities such as GELU \cite{he2024simplifying,noci2024shaped,he2023deep}, or even replace LayerNorm with \texttt{tanh} \citep{Zhu2025DyT}--both of which remain expensive to compute under encryption. 
Consequently, their efficiency in plaintext settings often fails to translate effectively to cryptographic settings. 

The second approach approximates nonlinear functions with polynomials, mainly targeting GELU and inverse-square-root in LayerNorms \cite{zimerman2023converting,dhyani2023privit}. 
While promising, these approximations come with their own set of challenges: they tend to work only within narrow input ranges and can make training unstable, particularly when high-degree polynomials are involved \cite{knott2021crypten,zimerman2023converting}. 
Additionally, a few prior on private  transformer-based models  overlook the cost of LayerNorm \citep{li2023mpcformer,zeng2022mpcvit,Zhang_2023_ICCV,chen2023rna}.  

These findings, while valuable, do not address the comprehensive role of nonlinearities in maintaining model stability and fostering attention head diversity in a multi-layer LLM, limiting their utility for private LLM design. We address this gap with an information-theoretic framework that  analyzes how nonlinearities shape learning dynamics at {\em pre-training}. Refer to Appendix \ref{AppendixSec:extnd_related_work} for additional related work.

\section{Background}

We first give an overview of the Transformer architecture in decoder-only LLMs, followed by the threat model and cryptographic protocols details. We also discuss our methodological considerations for characterizing nonlinearities.

\subsection{Transformer Architecture  in Decoder-only LLMs}

{\bf Notations}
We denote the number of layers as $L$, number of heads as $H$, model dimensionality as $d$, head dimension as $d_k$ (where \(d_k = \frac{d}{H}\)), and context length as $T$.  


{\bf An overview of transformer-based decoder-only architecture.}
A transformer-based LLM is constructed by sequentially stacking \(L\) transformer blocks, where each block is composed of two sub-blocks: an attention mechanism and a feed-forward network (FFN), both having their own residual connections and normalization layers (PreLN \citep{xiong2020layer}). Formally, transformer blocks take an input sequence \(\mathbf{X}_{\text{in}} \in \mathbb{R}^{T \times d}\), consisting of \(T\) tokens of dimension \(d\), and transform it into \(\mathbf{X}_{\text{out}}\) as follows:

\vspace{-2em}

\begin{equation} \label{eqn:ffn_mha}
\begin{aligned}
\mathbf{X}_{\text{out}} &= \hat{\mathbf{X}}_{\text{SA}} + \text{FFN}_{\text{GELU}}(\text{LayerNorm}_2(\hat{\mathbf{X}}_{\text{SA}})), \\
&\text{where} \; \hat{\mathbf{X}}_{\text{SA}} = \mathbf{X}_{\text{in}} + \text{MHA}(\text{LayerNorm}_1(\mathbf{X}_{\text{in}}))
\end{aligned}
\end{equation}

The Multi-Head Attention (MHA) sub-block enables input contextualization by sharing information between individual tokens. The self-attention mechanism in MHA computes the similarity score of each token with respect to all other tokens in the sequence. In particular, self-attention mechanism transform the input sequence \(\mathbf{X}\) into  \(\mathbf{Attn}(\mathbf{X})\) as follows:

\vspace{-1.5em}

\begin{equation} \label{eqn:attn_softmax}
\text{Attn}(\mathbf{X}) = \Big(\text{Softmax}\Big(\frac{(\mathbf{X}\mathbf{W}^Q)(\mathbf{X}\mathbf{W}^K)^\top}{\sqrt{d_k}} + \mathbf{M}\Big)\Big)\mathbf{X}\mathbf{W}^V
\end{equation}

Here, each token generates query($Q$), key($K$), and value($V$) vectors through the linear transformations \(\mathbf{W}^Q, \mathbf{W}^K, \; \text{and} \; \mathbf{W}^V \in \mathbb{R}^{d \times d_h}\), respectively. Then, similarity scores are computed by taking the dot product of $Q$ and $K$, scaled by the inverse square root of the $K$ dimension, and passed through a softmax function to obtain the attention weights. These attention scores are  used for computing a weighted sum of the $V$ vectors, producing the output for each token. For auto-regressive models, mask \(\mathbf{M} \in \mathbb{R}^{T\times T}\),  with values in \(\{0, -\infty\}\) with \(\mathbf{M}_{i,j} = 0 \,\text{iff} \, {i \geq j}\), is used for masking the information from future tokens.

The MHA sub-block employs a self-attention mechanism across all the heads, each with its own $Q$, $K$, and $V$. This allows the attention heads to focus on different parts of the input sequence, capturing various aspects of the input data simultaneously. The outputs from all heads are concatenated and linearly transformed ($\mathbf{W}^O\in\mathbb{R}^{d\times d}$) to produce the final MHA output as follows: 

\vspace{-2em}

\begin{equation} \label{eqn:mha_concat}
\text{MHA}(\mathbf{X}) = \text{Concat}\big(\text{Attn}_1(\mathbf{X}),  \dots, \text{Attn}_H(\mathbf{X})\big) \mathbf{W}^O
\end{equation}

\vspace{-0.5em}

Following the MHA sub-block, the FFN sub-block transforms each token independently. The FFN sub-blocks have a single hidden layer whose dimension is a multiple of \(d\) (e.g., \(4d\) in GPT-2). Specifically, the FFN sub-block first applies a linear transformation to the input \(\mathbf{X}\) using \(\mathbf{W}^{\text{ffn}}_{\text{in}} \in \mathbb{R}^{d \times 4d}\), followed by a non-linear transformation using an activation function such as GELU. This is then followed by another linear transformation using \(\mathbf{W}^{\text{ffn}}_{\text{out}} \in \mathbb{R}^{4d \times d}\), as follows:

\vspace{-1.5em}

\begin{equation} \label{eqn:ffn_gelu}
\text{FFN}(\mathbf{X}) = (\text{GELU}(\mathbf{X} \mathbf{W}^{\text{ffn}}_{\text{in}}))\mathbf{W}^{\text{ffn}}_{\text{out}}
\end{equation}



\subsection{Threat Model and Cryptographic Protocols} 

\begin{figure} [t]
\centering
\includegraphics[width=0.5\textwidth]{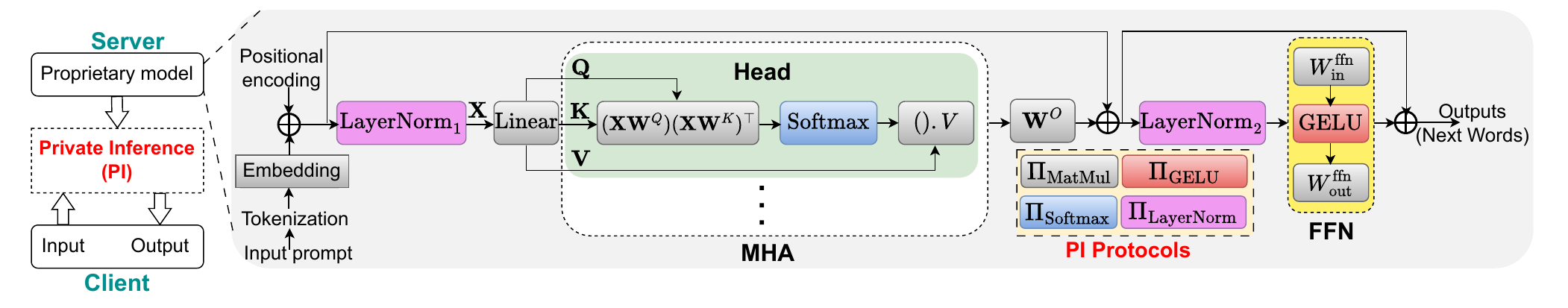} \vspace{-2em}
\caption{An illustration of threat model for  private LLM inference.} 
\label{fig:GptBlockDiagram}
\end{figure}

{\bf Threat model for private LLM Inference}
We use standard two-party (2PC) client-server setting, which provides security against semi-honest (honest-but-curious) adversaries bounded by probabilistic polynomial time \cite{xu2025breaking,zhang2024secure,lu2023bumblebee,pang2023bolt,hou2023ciphergpt}.  Both parties follow protocol specifications but may attempt to gain additional information from their outputs about the other party's input. In this 2PC  setting (see Figure \ref{fig:GptBlockDiagram}), the server holds the propriety LLM, and the client queries the model with a prompt. The protocols ensure that server does not know anything about the client's input and the output of their queries, and client does not know anything about the server's model except its architecture.

Next, we describe the cryptographic protocols used for linear and nonlinear operations during private inference. In particular, we use BumbleBee \cite{lu2023bumblebee} private inference framework for end-to-end private inference evaluation.

{\bf Linear Operations (MatMul)}
For privacy-preserving matrix multiplication operations, BumbleBee  leverages homomorphic encryption with a novel ciphertext compression strategy. The protocol implements Oblivious Linear Transform with efficient packing techniques that reduce communication costs by 80-90\% compared to previous approaches. Testing on BERT-base model showed 92\% less communication compared to IRON \cite{hao2022iron} and 90\% less than BOLT \cite{pang2023bolt}. The protocol seamlessly handles both scenarios where one matrix is in plaintext and another is secret-shared, and when both matrices are secret-shared.

{\bf GELU} To efficiently compute GELU activation, the protocol employs strategic polynomial approximations (with degree 3 and 6)  across input ranges. The implementation optimizes branch selection through batched comparisons and leverages mixed bitwidth arithmetic, and maintains model accuracy within 1\% of plaintext evaluation.

{\bf ReLU} The private ReLU is implemented through a composition of secure comparison and multiplexer operations. Using optimized OT(oblivious transfer)-based comparison protocols and Boolean-to-arithmetic conversions, achieving significant efficiency gains. The protocol leverages Ferret OT \cite{yang2020ferret} instead of traditional IKNP OT \cite{ishai2003extending}, contributing to the overall improvement in communication efficiency while maintaining the simplicity.

{\bf LayerNorm} The LayerNorm computation breaks down into optimized sub-components: secure mean calculation, variance computation using an efficient square protocol, and normalization using reciprocal square root. The implementation takes advantage of squaring operations costing half of general multiplication in secure computation. This optimization, combined with efficient reciprocal computation, shows significant improvement over IRON's approach. 

{\bf Softmax} For Softmax computation, the protocol implements a numerically stable approach using max normalization. The exponential function computation is optimized specifically for negative inputs using Taylor approximation. The division operation is restructured to use reciprocal followed by multiplication, leads to 80\% saving in communication costs while maintaining numerical precision \citep{lu2023bumblebee}.

\subsection{Methodological Consideration}

{\bf Why training from scratch to study nonlinearities?} 
Isolating the functional role of architectural components, such as activation functions (GELU, ReLU) and LayerNorms, requires examining their impact from initialization, uninfluenced by prior optimization biases. Training from scratch provides the necessary experimental control:
\begin{enumerate}[noitemsep,nolistsep,leftmargin=0.5cm] \vspace{-0.8em}
\item \textit{Capturing fundamental learning dynamics:} Nonlinearities shape gradient flow and optimization trajectories from the first weight update. Training from scratch shows how these components influence convergence behavior and stability during critical early-stage learning, when attention patterns and representations first emerge.
\item \textit{Clean ablations at feasible scales:} Starting from random initialization eliminates confounding factors from pre-trained weights, ensuring observed effects stem solely from architectural modifications. Moreover, controlled model scales (e.g., 125M parameters) enable exhaustive ablation  across multiple architectural variants, which is infeasible with larger models.
\item \textit{Limitations of fine-tuning approaches:} Fine-tuning from pre-trained checkpoints cannot reveal how nonlinearities affect early optimization since the trajectory is already determined, and learned representations mask architectural effects. Granular entropy analysis requires observing complete training trajectories from initialization. 
\end{enumerate}


{\bf Why perplexity is a reliable metric for evaluating LLMs at short context lengths?}
Perplexity \citep{jelinek1977perplexity} remains the standard metric for evaluating the pre-training quality, capturing token-level predictive performance, in autoregressive LLMs. However,  to ensure its comparability across architectural variants, factors such as tokenizer design, vocabulary size and quality must remain consistent \citep{hutchins2022block}. Else,  it can obscure the true impacts of architectural simplification. Thus, in all our experiments, we control these factors by adopting a uniform tokenizer and maintaining a fixed vocabulary {\em within} each dataset.

Nonetheless, perplexity does have {\em known limitations at very long contexts} (e.g., 32k): it weakly correlates with the true contextual understanding \citep{hu2024can,fang2025what}. At very large context, only a small fraction of tokens (often $<$10 \%) benefit from the extended attention window, leading to a dilution effect where perplexity mostly measures local predictability rather than long-range reasoning.

Since private LLM inference primarily focuses on smaller context lengths due to computational constraints  \citep{zhang2024secure,lu2023bumblebee,zimerman2023converting,pang2023bolt,gupta2023sigma,hou2023ciphergpt}, we strategically use {\em shorter context length} (128-512), where most tokens lie well within the effective attention span. Thus, perplexity remains informative and  provides a reliable basis for evaluating the effect of architectural simplifications.


{\bf Why use Shannon's entropy to evaluate the impact of nonlinearities in LLMs?} 
While perplexity (PPL) is the standard metric for evaluating the language modeling quality in decoder-only LLMs, it primarily reflects the model’s end-to-end behavior. As such, PPL conflates multiple error sources such as tokenization artifacts, sampling noise, and compounding effects across layers, making it {\em ill-suited} for diagnostic. It obscures {\em where} and {\em why} a model fails. 

In contrast, Shannon’s entropy of the attention score distribution provides a fine-grained view of information flow, functional specialization, and failure modes during training. Hence, it is widely used for understanding the effect of architectural design on representation dynamics
 \cite{zhang2024attention,lee2025mitigating,zhang2024the,nahshan2024linear,vig2019analyzing,ghader2017does}.

Below, we outline key reasons for using entropy as an interpretable metric for analyzing nonlinearities in our work:
\begin{enumerate} [noitemsep,nolistsep,leftmargin=0.5cm] \vspace{-0.6em}
\item {\em Quantifying information flow and head-specific diagnostics:} Analyzing per-head entropy statistics characterizes information flow across layers. A higher entropy corresponds to a more diffuse attention distribution, indicating exploration, whereas lower entropy signify concentration on a few key tokens, suggesting exploitation.
\item {\em Mechanistic specificity:} Entropy isolates the behavior of the self-attention mechanism without interference from the FFN sub-blocks, which allows us to attribute representational instability directly to the attention collapse. 
\item {\em Actionable causality rather than correlation:}
Recent studies demonstrate that attention entropy is an actionable cause---not merely a symptom---of degeneracy in transformer models. \citet{velickovic2025softmax} proved a monotone temperature-entropy law, showing that entropy directly governs model efficiency, and  \citet{zhai2023stabilizing} linked entropy collapse to the spectral norm of the score matrix, which controls gradient stability. 
\end{enumerate}

\section{Characterizing LLM Nonlinearity Through Entropy Dynamics} \label{sec:NonlinearCharacterization}


{\bf Setup.} To analyze the role of LLM nonlinearities, we progressively remove them from the architecture and observe their entropic dynamics at {\em pre-training}. Shannon's entropy measures the uncertainty in a probability distribution, defined as \( \mathbf{E}(P) = -\sum_{i} P(x_i) \log P(x_i) \),  reflecting the amount of information needed to describe a stochastic system \cite{shannon1948mathematical,jaynes1957information}. In Softmax-based attention mechanisms, entropy quantifies the sharpness or spread of the attention scores for each query position: higher entropy indicates a uniform distribution, while lower entropy indicates a focused attention on specific input tokens.


Let \(\mathbf{A}^{(h,l)} \in \mathbb{R}^{T\times T} \) be the attention matrix of $h$th head in $l$th layer, and each element in the attention matrix (\( a_{ij}^{(l, h)} \)) are attention score for the \(i\)th query and \(j\)th key:

\vspace{-2em}

\begin{equation*}
\mathbf{A}^{(l, h)} = \left[ a_{ij}^{(l, h)} \right]_{T \times T}, \; a_{ij}^{(l, h)} \geq 0 \; \text{and} \; \sum_{j=1}^{T} a_{ij}^{(l, h)} = 1 
\end{equation*}

\vspace{-1em}

This square matrix is the outcome of Softmax operation over the key length for each query position as follows 

\vspace{-2em}

\begin{equation*} 
\mathbf{A}^{(h,l)}(\mathbf{X}) = \text{Softmax}\Big(\frac{1}{\sqrt{d_k}} (\mathbf{X} \mathbf{W}^Q) (\mathbf{X}{\mathbf{W}^K})^\top \Big)
\end{equation*}

\vspace{-1em}

Following \citet{zhai2023stabilizing}, we compute the average entropy across all query positions within an attention head to obtain its headwise entropy. Entropy \( \mathbf{E}^{(l, h)} \) for the \( h \)th head in \( l \)th layer of an attention matrix is given by:

\vspace{-2em}

\begin{equation*} 
\begin{aligned}
\mathbf{E}^{(l, h)} &= -\frac{1}{T} \sum_{i=1}^{T} \sum_{j=1}^{T} a_{ij}^{(l, h)} \log (a_{ij}^{(l, h)}), \\
\text{where} \; a_{ij}^{(l, h)} &= \frac{\exp\left(\frac{1}{\sqrt{d_k}} (\mathbf{X}_i \mathbf{W}^Q) (\mathbf{X}_j \mathbf{W}^K)^\top \right)}{\sum_{k=1}^{T} \exp\left(\frac{1}{\sqrt{d_k}} (\mathbf{X}_i \mathbf{W}^Q) (\mathbf{X}_k \mathbf{W}^K)^\top \right)}
\end{aligned}
\end{equation*}

\begin{table}[t]
\caption{Architectural configurations of nonlinearities in LLMs, illustrating the combinations of Softmax (SM), LayerNorm (LN), GELU (G), and ReLU (R) functions (see Eq. \ref{eqn:ffn_mha}, \ref{eqn:attn_softmax}, \ref{eqn:mha_concat} and \ref{eqn:ffn_gelu}). } 
\label{tab:ArchConfigGPT2}
\centering 
\resizebox{0.49\textwidth}{!}{
\begin{tabular}{l|c} \toprule 
Abbreviation & Architectural configuration \\ \toprule 
\textcolor{blue}{SM} + \textcolor{violet}{LN} + \textcolor{red}{G} & $\mathbf{X}_{\text{out}} = \text{FFN}_{\text{\textcolor{red}{GELU}}}(\text{\textcolor{violet}{LayerNorm}}_{\textcolor{violet}{2}}(\text{MHA}(\text{Attn}_{\text{\textcolor{blue}{Softmax}}}(\text{\textcolor{violet}{LayerNorm}}_{\textcolor{violet}{1}}(\mathbf{X}_{\text{in}})))))$ \\
\textcolor{blue}{SM} + \textcolor{violet}{LN} + \textcolor{red}{R} & $\mathbf{X}_{\text{out}} = \text{FFN}_{\text{\textcolor{red}{ReLU}}}(\text{\textcolor{violet}{LayerNorm}}_{\textcolor{violet}{2}}(\text{MHA}(\text{Attn}_{\text{\textcolor{blue}{Softmax}}}(\text{\textcolor{violet}{LayerNorm}}_{\textcolor{violet}{1}}(\mathbf{X}_{\text{in}})))))$ \\ 
\textcolor{blue}{SM} + \textcolor{violet}{LN} & $\mathbf{X}_{\text{out}} = \text{FFN}_{\text{Identity}}(\text{\textcolor{violet}{LayerNorm}}_{\textcolor{violet}{2}}(\text{MHA}(\text{Attn}_{\text{\textcolor{blue}{Softmax}}}(\text{\textcolor{violet}{LayerNorm}}_{\textcolor{violet}{1}}(\mathbf{X}_{\text{in}})))))$ \\ 
\textcolor{blue}{SM} + \textcolor{red}{G} & $\mathbf{X}_{\text{out}} = \text{FFN}_{\text{\textcolor{red}{GELU}}}(\text{MHA}(\text{Attn}_{\text{\textcolor{blue}{Softmax}}}(\mathbf{X}_{\text{in}})))$ \\ 
\textcolor{blue}{SM} + \textcolor{red}{R} & $\mathbf{X}_{\text{out}} = \text{FFN}_{\text{\textcolor{red}{ReLU}}}(\text{MHA}(\text{Attn}_{\text{\textcolor{blue}{Softmax}}}(\mathbf{X}_{\text{in}})))$ \\ 
\textcolor{blue}{SM} & $\mathbf{X}_{\text{out}} = \text{FFN}_{\text{Identity}}(\text{MHA}(\text{Attn}_{\text{\textcolor{blue}{Softmax}}}(\mathbf{X}_{\text{in}})))$ \\ \bottomrule
\end{tabular} }
\end{table}


\begin{figure} [t]
\centering
\includegraphics[width=0.49\textwidth]{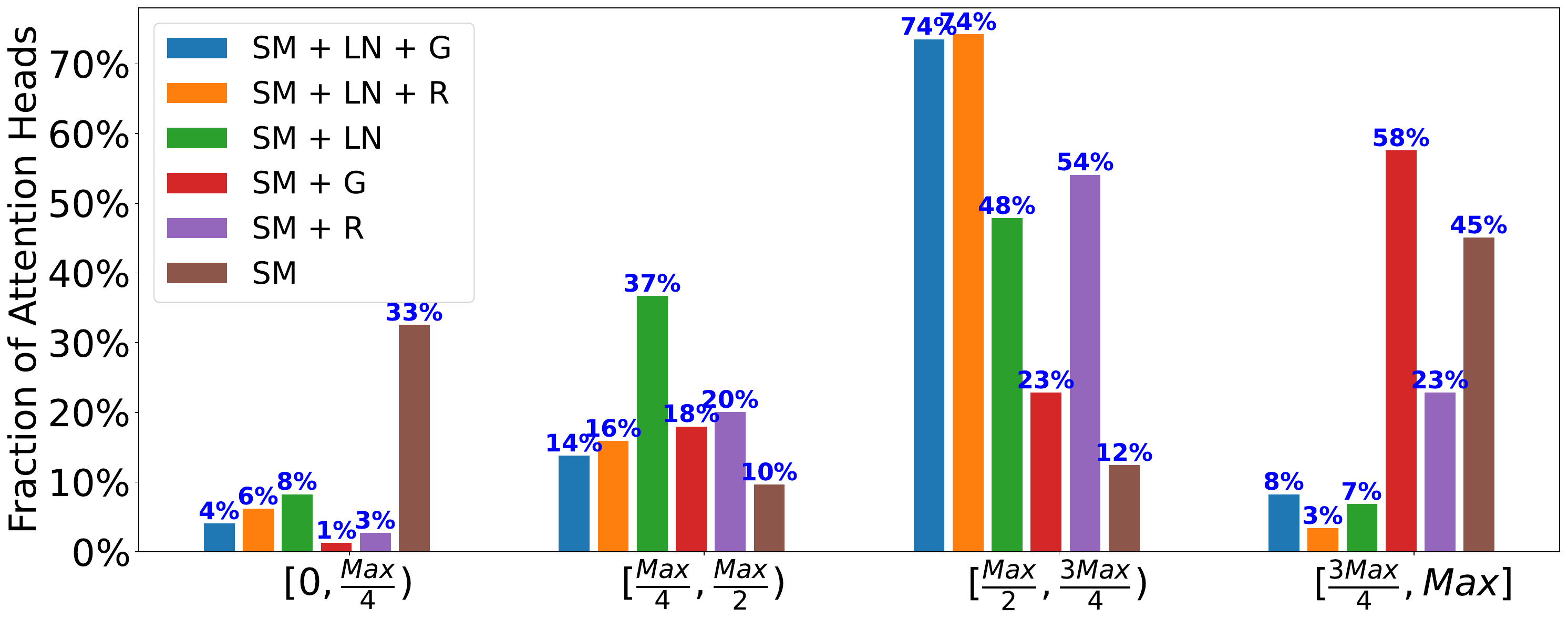} \vspace{-2.2em}
\caption{Distribution of attention heads (\%) across entropy ranges for different model configurations (Table \ref{tab:ArchConfigGPT2}) trained from scratch, showing the concentration of heads in specific entropy intervals.} 
\label{fig:FractionAttentionHeads}
\end{figure}

{\bf Well-behaved entropy distribution for LLMs.} 
First, we analyze the headwise entropy distribution of baseline architecture with GELU and ReLU in the FFN, i.e., configurations ${\tt SM + LN + G}$ and ${\tt SM + LN + R}$, respectively. We find that the majority of heads ($\approx$90\%) possess entropy values between $\frac{\text{max}}{4}$ and $\frac{\text{3max}}{4}$, where ${\tt max}$ is maximum observed entropy value among all heads (see Figure \ref{fig:FractionAttentionHeads}).  
This concentration in the mid-entropy range, while avoiding extremes, shows a well-behaved distribution, providing a benchmark for assessing the impact of nonlinearities on model behavior.


{\bf Entropic overload in LLMs.}
We observed that in the absence of crucial nonlinearities, a disproportionately large fraction of the attention heads exhibit higher entropy values (between $\frac{3\text{max}}{4}$ and ${\tt max}$). We term this phenomenon as entropic overload and hypothesize that this imbalance results in {\em under-utilization} of multi-head attention (MHA), as too many heads engaged in exploration, impairing the model's ability to learn diverse and specialized attention patterns.

To better understand this behavior, we analyze how entropy evolves throughout training. In models with all nonlinearities intact, attention heads typically start with high entropy, reflecting an initial phase of exploration, and gradually transition toward a balance between exploration and exploitation (see Figure \ref{subfig:sm_ln_g} and \ref{subfig:sm_ln_r}). However, when key nonlinearities are removed, this balance breaks down and any attention heads in the early layers remain stuck in persistently higher entropy states (see Figure \ref{subfig:sm_g}), which impedes learning and ultimately degrades model performance.

{\bf Entropy collapse in LLMs.}
This is characterized by extremely-smaller (near-zero) entropy values and recognized as a key indicator of training instability in transformer architectures \cite{zhai2023stabilizing,he2024understanding}. 



\subsection{Activation Functions in Normalization-Free LLMs}

While GELU is typically favored in transformer models for its smoothness and optimization benefits, we observe the {\em opposite trend} in normalization-free settings: ReLU outperforms GELU by achieving a {\bf 8.2}\% lower perplexity, which is consistent with the 
normalization-free Pythia models across various context lengths (Table \ref{tab:NormFreePythiaGPT}).

\begin{table}[t]
\caption{Performance comparison of baseline and normalization-free GPT-2 (12$L$, 12$H$, 768$d$) and Pythia-70M (6$L$, 8$H$, 512$d$) models, each with GELU and ReLU variants, trained on CodeParrot. While GELU outperforms ReLU in baseline models, the normalization-free models exhibit the {\em opposite trend}. }
\label{tab:NormFreePythiaGPT}
\centering
\resizebox{0.49\textwidth}{!}{
\begin{tabular}{lcccccc}
\toprule
& \multicolumn{2}{c}{GPT-2 ($T$=128)} & \multicolumn{2}{c}{Pythia-70M ($T$=128)} & \multicolumn{2}{c}{Pythia-70M ($T$=256)} \\
\cmidrule(lr){2-3} \cmidrule(lr){4-5} \cmidrule(lr){6-7}
& Eval PPL & +$\Delta$(\%) & Eval PPL & +$\Delta$(\%) & Eval PPL & +$\Delta$(\%) \\ 
\toprule
SM+LN+G & 2.69 & 0.00 & 3.51 & 0.00 & 3.05 & 0.00 \\
SM+LN+R & 2.76 & 2.53 & 3.59 & 2.22 & 3.11 & 1.73 \\
SM+G 
 & \cellcolor{green!15}3.20 & \cellcolor{green!15}18.92 
 & \cellcolor{green!15}4.09 & \cellcolor{green!15}16.35 
 & \cellcolor{green!15}3.57 & \cellcolor{green!15}16.87 \\
SM+R & 2.94 & 9.20 & 3.74 & 6.36 & 3.27 & 7.17 \\
\bottomrule
\end{tabular}}
\end{table}

\begin{figure} [t]
\centering
\subfloat[SM + LN + G \label{subfig:BaselineGELU}]{\includegraphics[width=.125\textwidth]{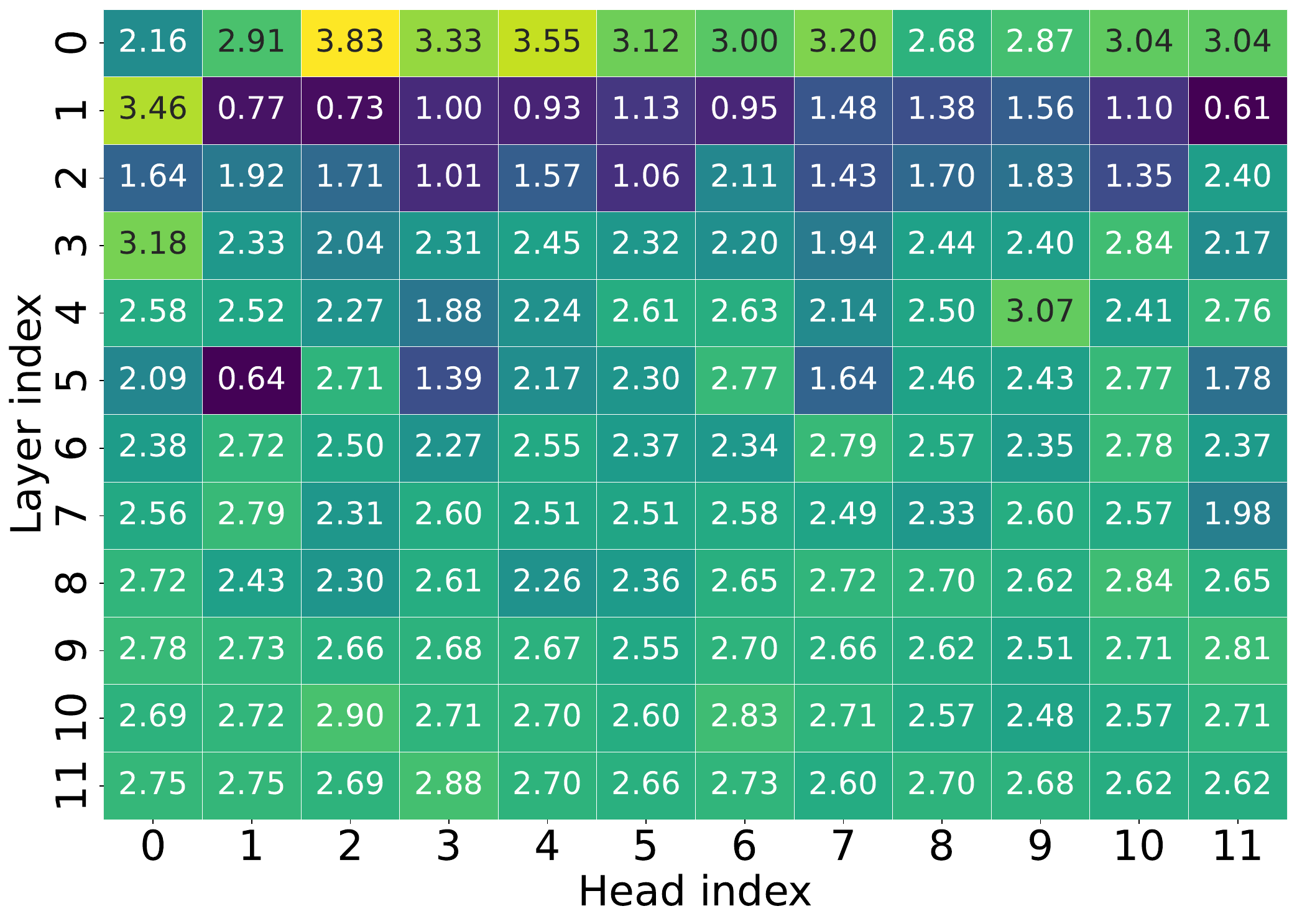}}
\subfloat[SM + LN + R \label{subfig:BaselineReLU}]{\includegraphics[width=.125\textwidth]{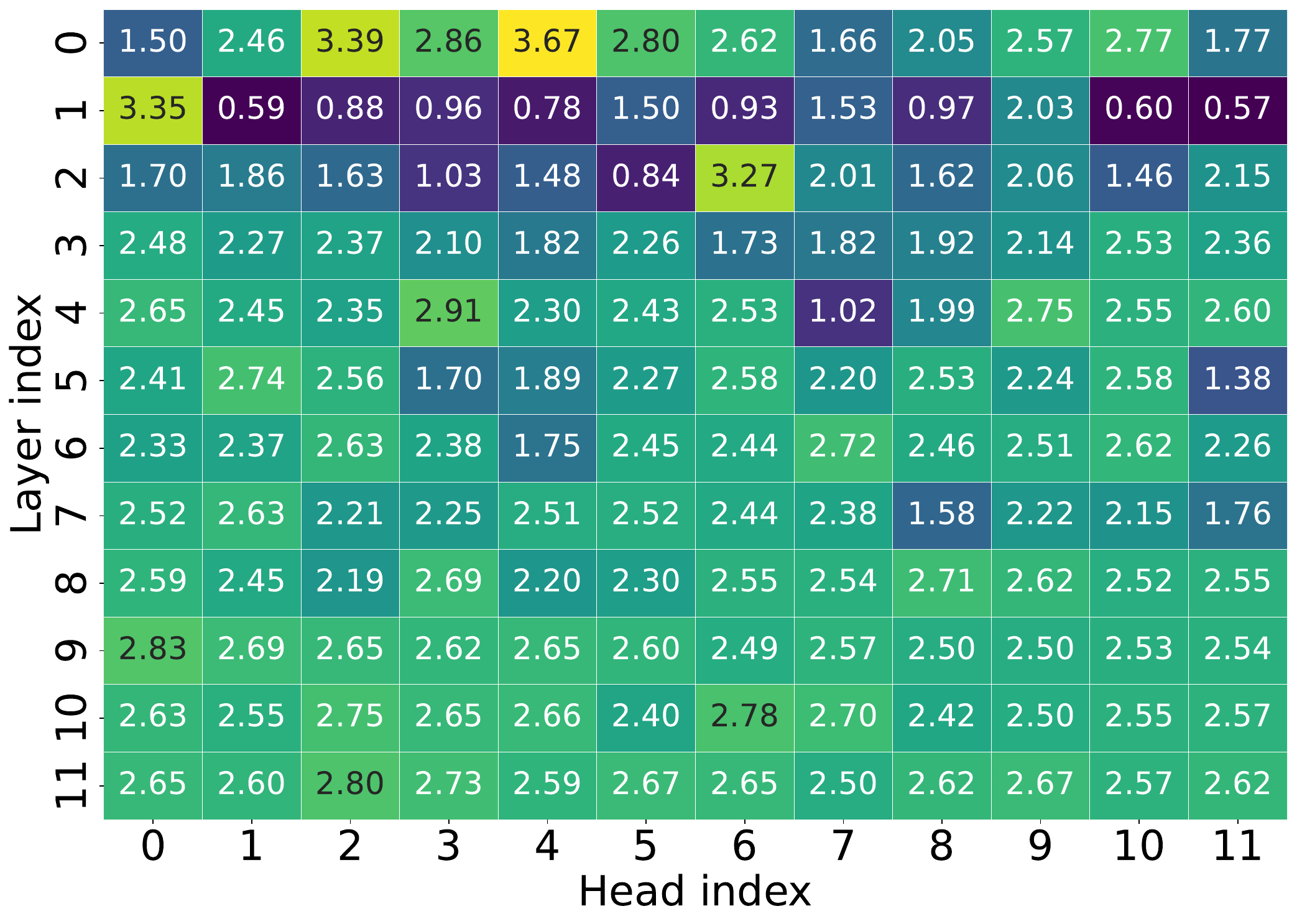}}  
\subfloat[SM + G \label{subfig:LNFreeGELU}]{\includegraphics[width=.125\textwidth]{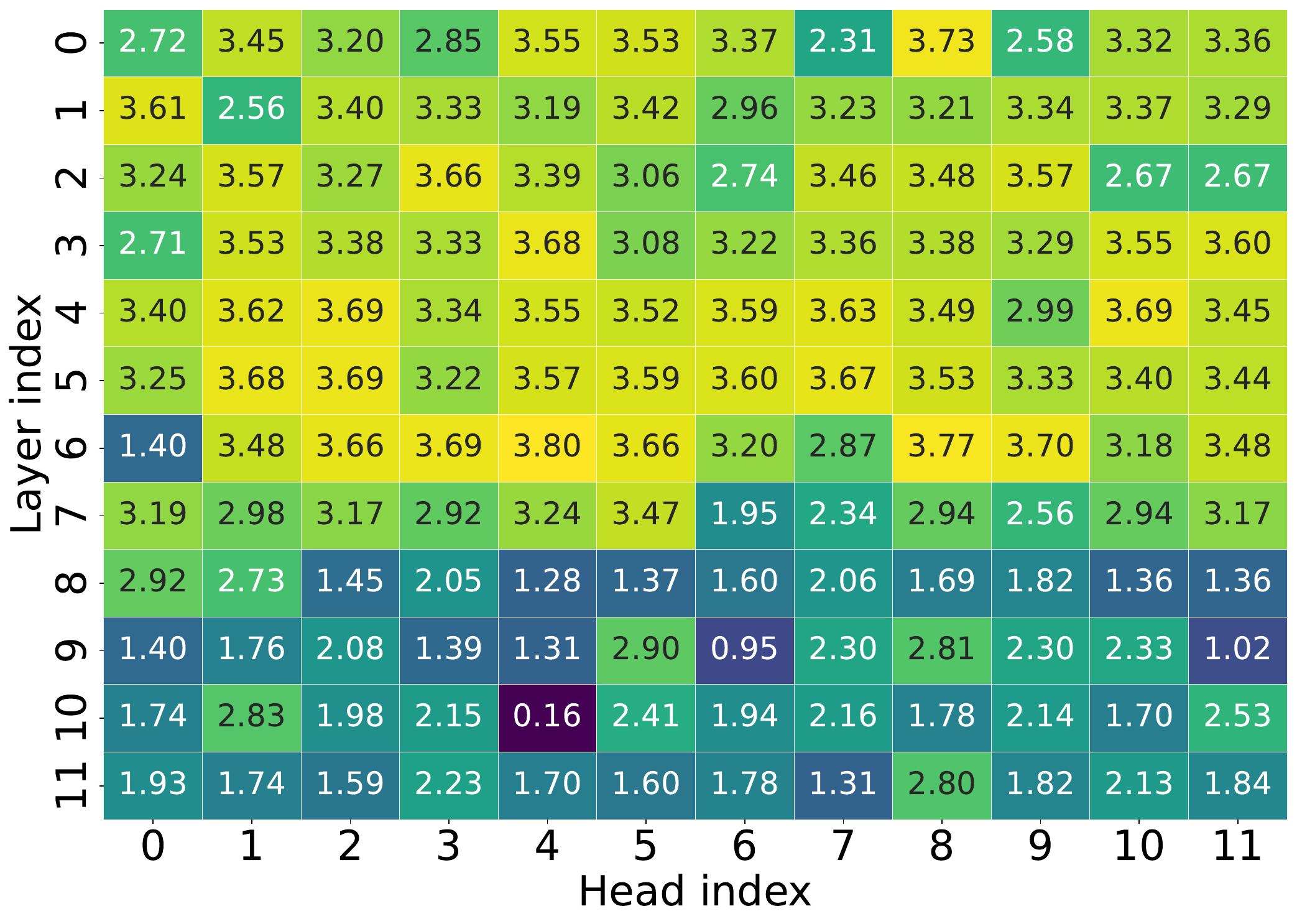}} 
\subfloat[SM + R \label{subfig:LNFreeReLU}]{\includegraphics[width=.125\textwidth]{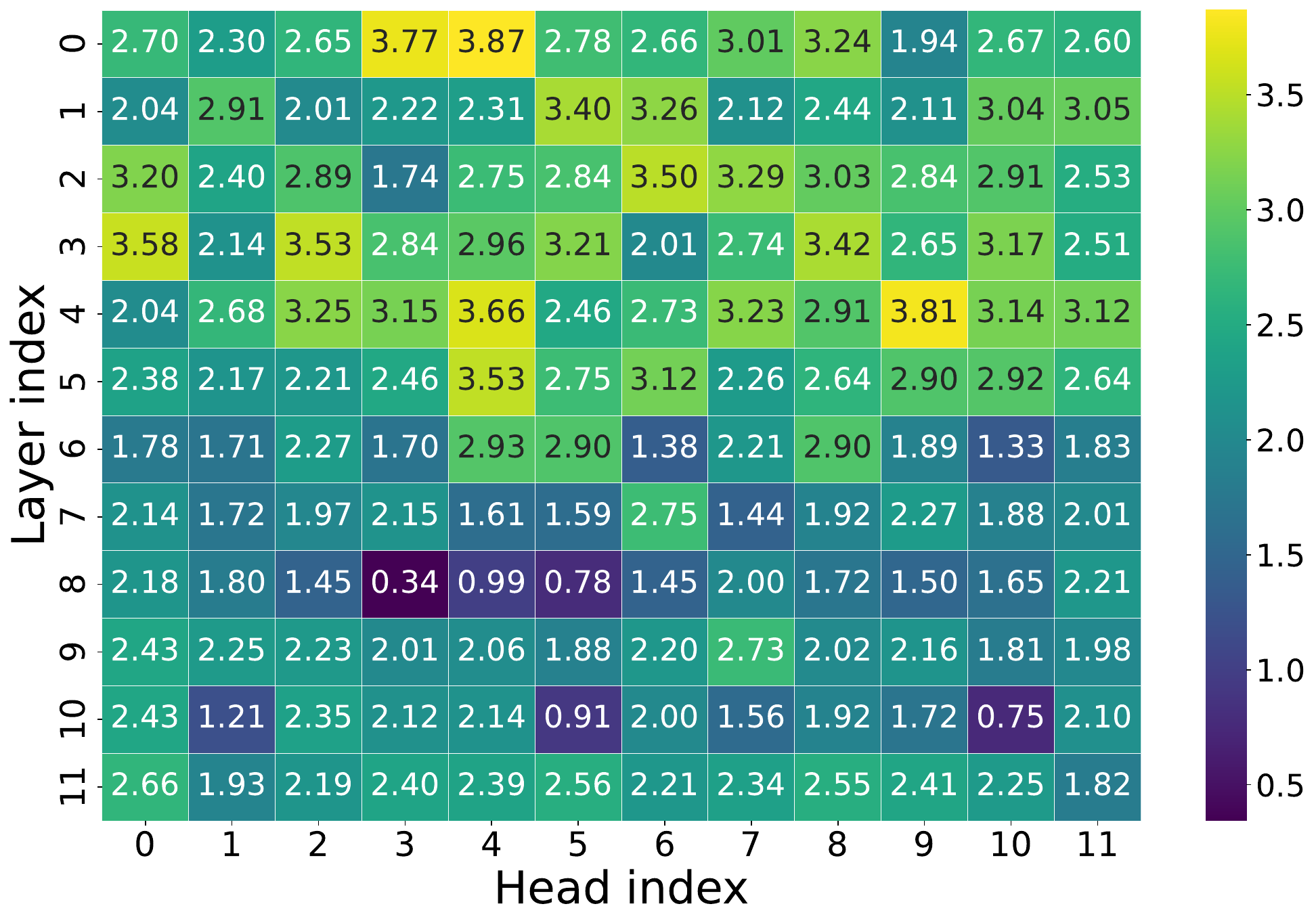}} 
\vspace{-1.5em}
\caption{Entropy heatmaps of GPT-2 with GELU and ReLU in the FFN (a, b) and their normalization-free variants (c, d). Without LayerNorm, {\em GELU causes significantly higher entropic overload.}}
\label{fig:AttnEntHeatMaps}
\end{figure}

{\bf Observation 1: In normalization-free LLM, ReLU outperforms GELU due to lesser entropic overload.}
Figure \ref{fig:FractionAttentionHeads} shows that 58\% of attention heads in the normalization-free GELU (SM+G) model have entropy values in the range $\frac{\text{3max}}{4}$ and ${\tt max}$, compared to only 23\% in the normalization-free ReLU model (SM+R). Moreover, very few heads in the latter approach maximum entropy, unlike  GELU variant (see yellow regions in Figure \ref{subfig:LNFreeReLU} \& \ref{subfig:LNFreeGELU}). This over-concentration  of high-entropy heads in SM+G suggests a lack of  attention diversity and specialization, which we believe is a key reason for its higher perplexity compared to SM+R.

\begin{figure} [t]
\centering
\subfloat[Layerwise learnable slope]{\includegraphics[width=.24\textwidth]{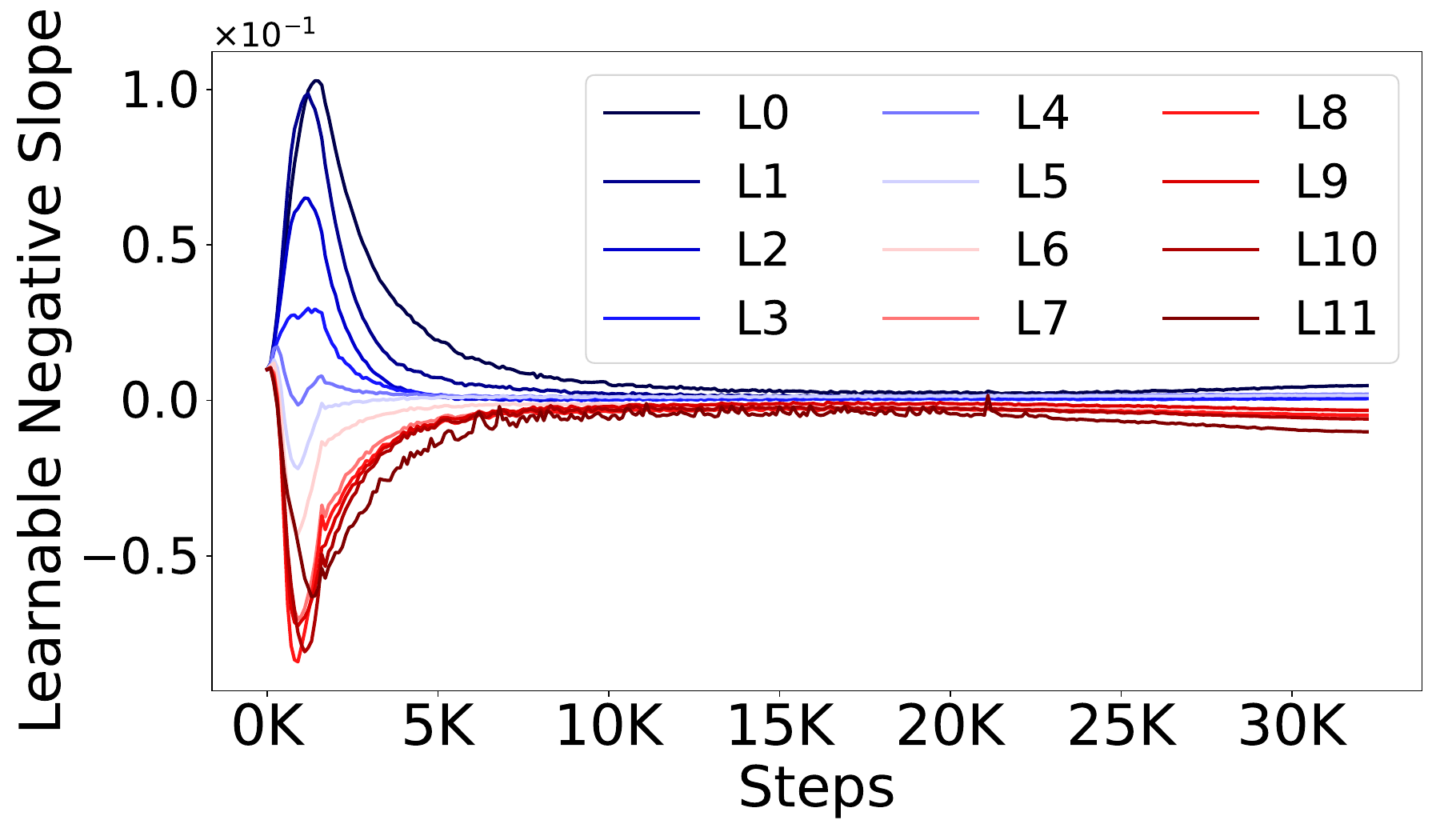}} 
\subfloat[Global learnable slope]{\includegraphics[width=.24\textwidth]{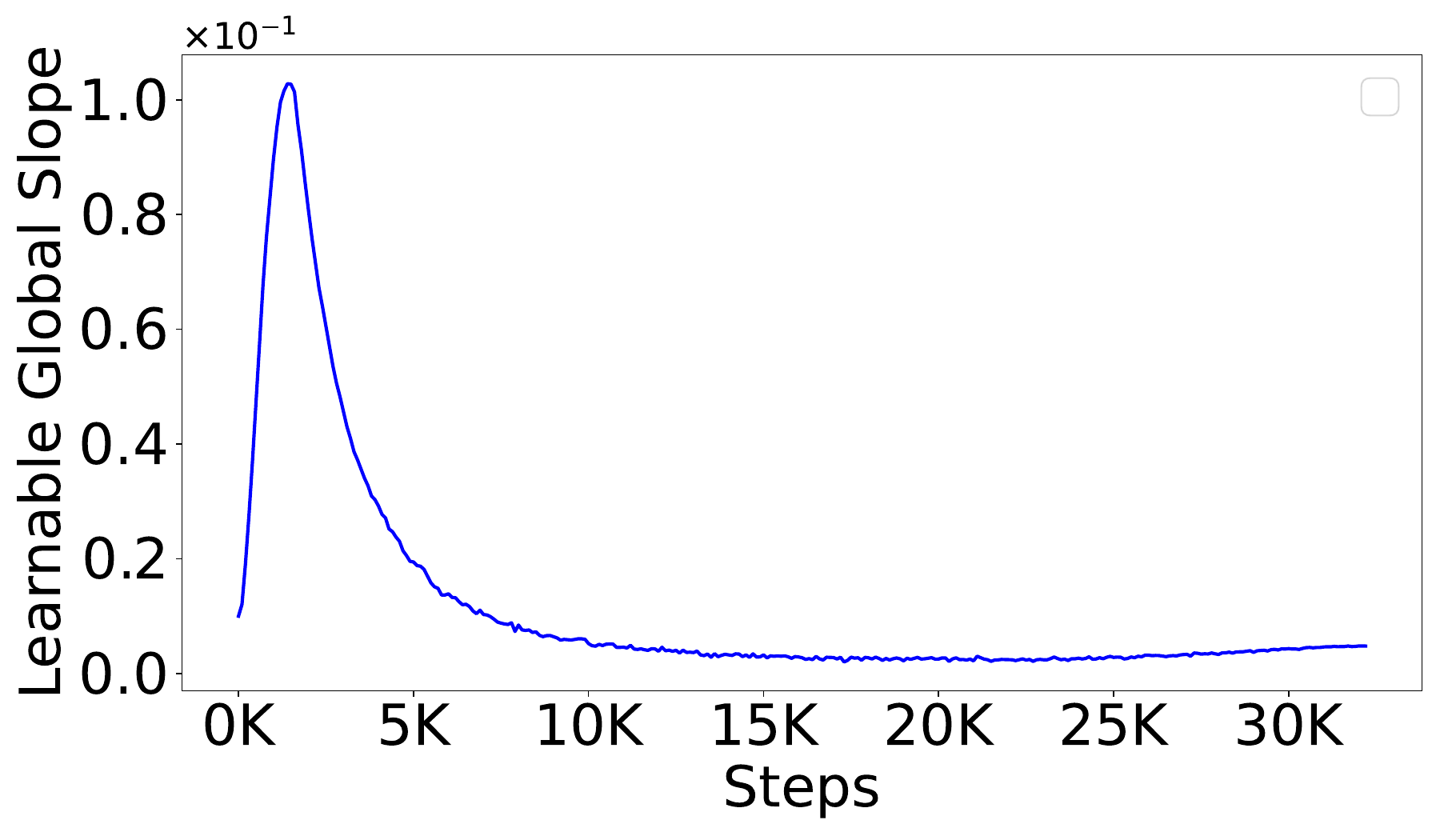}} \vspace{-0.8em}
\caption{Normalization-free GPT-2 with LeakyReLU having learnable negative slope. In both settings, (a) layerwise and (b) global, slope values  converge toward zero during training, indicating a preference for zero negative slope--A ReLU-like behavior.  } 
\label{fig:LearnableNegSlope}
\end{figure}

{\bf Observation 2: Normalization-free models with learnable activations naturally converge to ReLU-behavior}
To investigate the {\em inductive bias} favoring ReLU over GELU in normalization-free settings, we parameterize the Leaky ReLU activation with a learnable negative slope and evaluate two configurations: (1) layer-wise, where each layer has an independently learnable slope parameter, and (2) global, where a single slope is shared across all layers. Results in Figure~\ref{fig:LearnableNegSlope} shows that, in the layer-wise setting, early layers initially learn positive slopes while deeper layers exhibit negative ones; however, all slope parameters converge toward zero. In the global setting, the shared slope exhibits an initial drift toward positive values before approaching zero.

\begin{figure} [t]
\centering
\subfloat[{\footnotesize Learnable slope} \label{subfig:EntLearnable}]{\includegraphics[width=.16\textwidth]{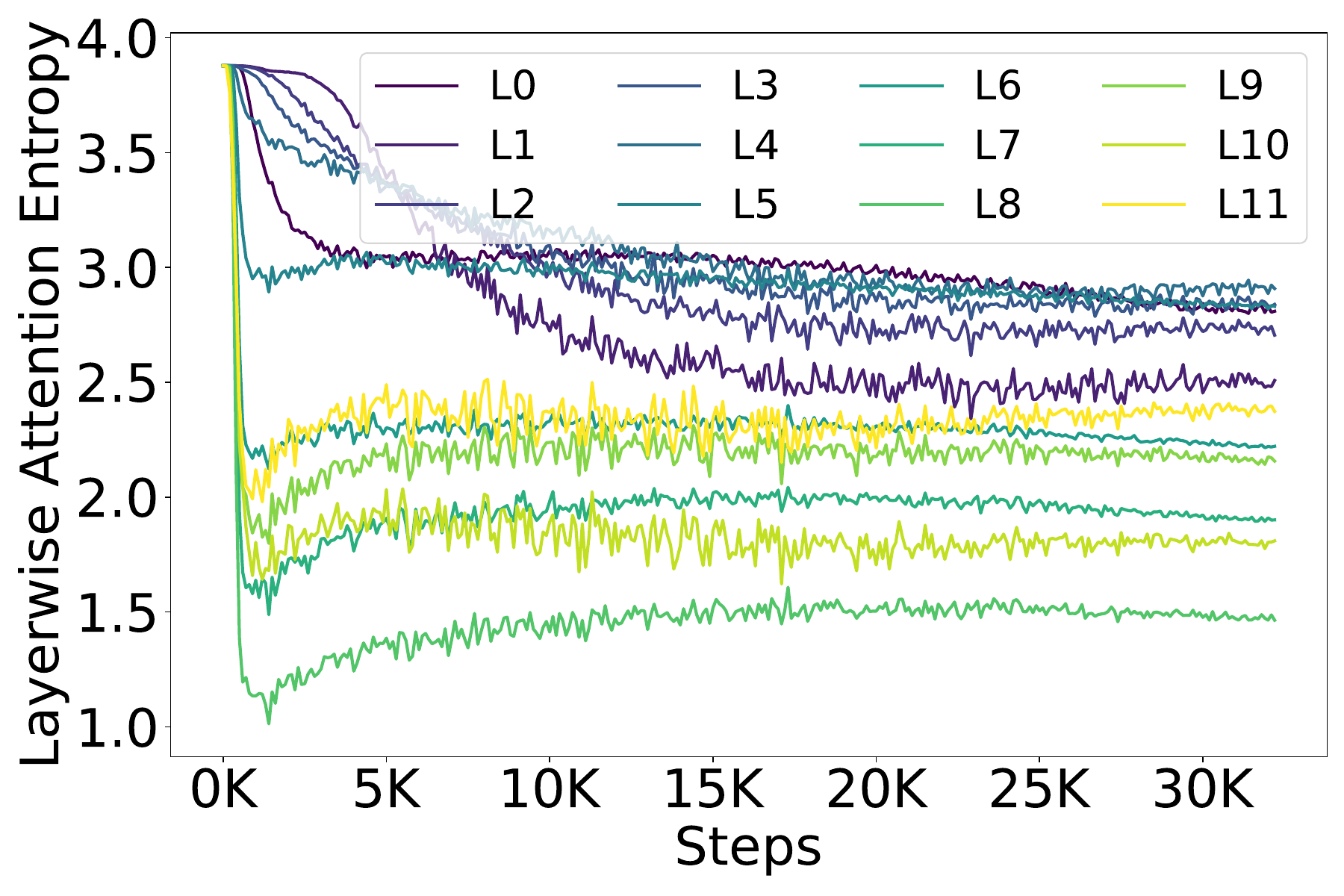}} 
\subfloat[{\footnotesize Fixed slope $5e$-2} \label{subfig:Ent5e2}]{\includegraphics[width=.16\textwidth]{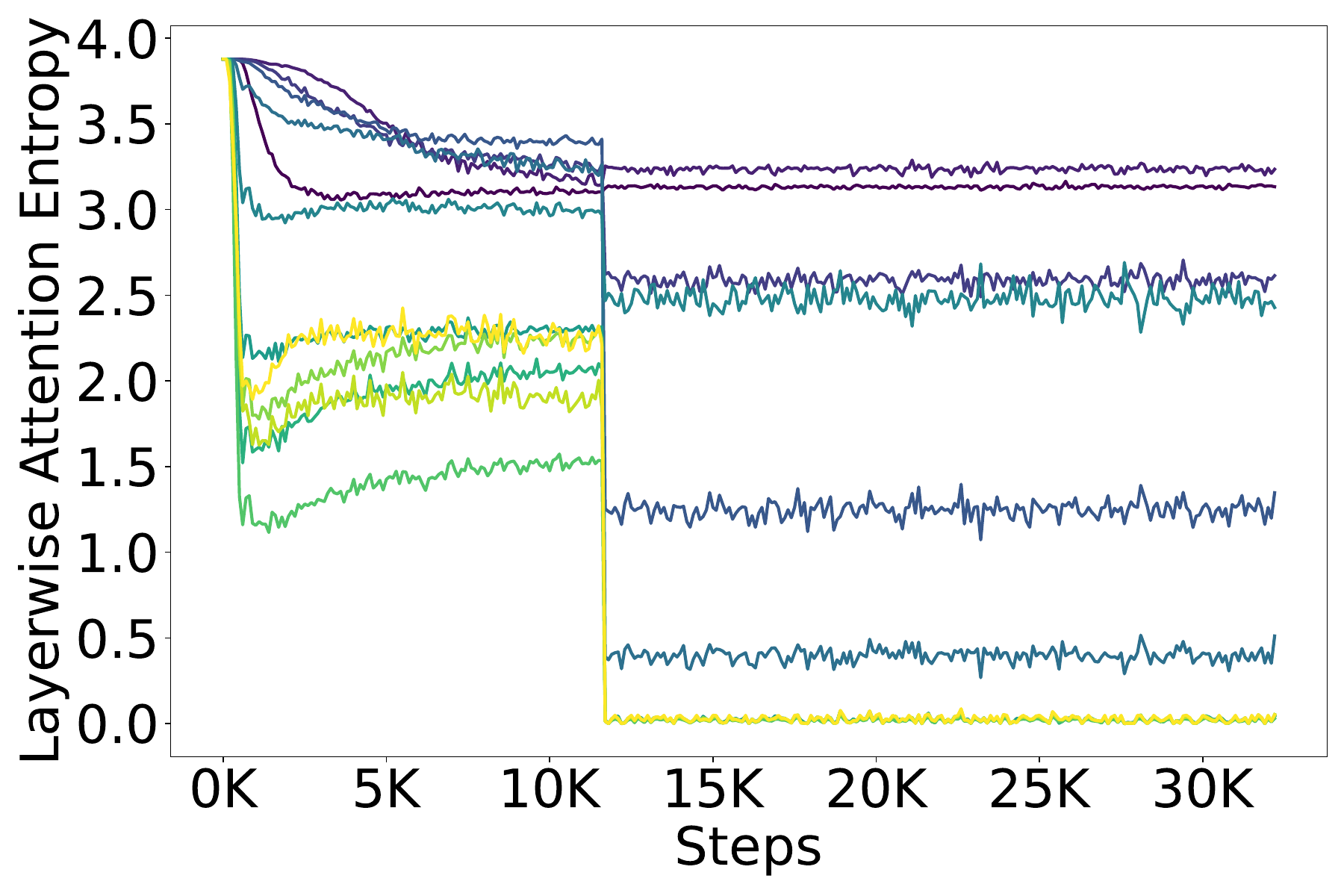}} 
\subfloat[{\footnotesize Fixed slope $1e$-1} \label{subfig:Ent1e1}]{\includegraphics[width=.16\textwidth]{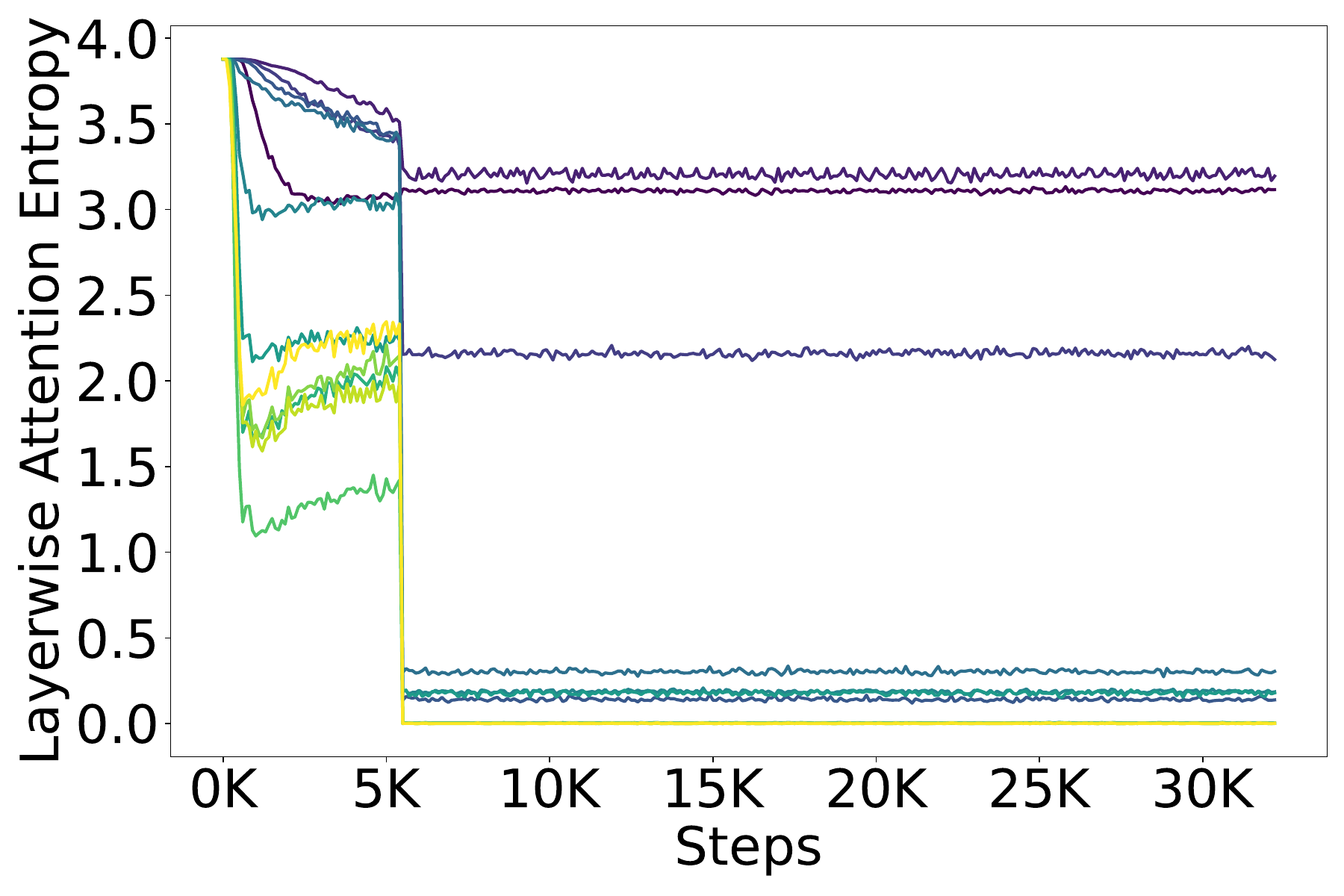}} \vspace{-0.6em}
\caption{Entropy dynamics in normalization-free GPT-2 (125M) with LeakyReLU having: learnable negative slopes (a)  and fixed slopes (b,c). While fixed slopes lead to earlier entropy collapse, learnable configuration mimic the stable entropy profile of ReLU. } 
\label{fig:NegSlopeEntropyDyamics}
\end{figure}


Further, we analyzed the entropy dynamics under learnable slope configurations and compared them to fixed negative slope settings in Leaky ReLU (with slopes of $5e$-2 and $1e$-1). Interestingly, the learnable configurations exhibit entropy dynamics that closely resembles those of normalization-free ReLU (Figure \ref{fig:NegSlopeEntropyDyamics}). In contrast, fixed-slope Leaky ReLU exhibits a clear trend: larger negative slopes induce earlier entropy collapse. This suggests that increasing the negative slope reduces the window of stable training. These findings further reinforce the inductive bias toward ReLU-like activations in normalization-free LLMs


\subsection{Selective Removal of Activations in FFNs}

While completely removing FFN activations degrades performance by {\bf +25.6\%} (3.376 vs 2.688 PPL, Table \ref{tab:PartialActsRemoval}), this degradation may not be uniform across layers, motivating us to explore selective removal strategies. Specifically, we ask: {\em do FFN layers exhibit location-dependent criticality?}. 

To this end, we explored three strategies: (1) 25\% activation, where we divide the network into four stages and remove activations from three; (2) 50\% activation, where activations are removed from alternating layers or half the network; and (3) fractional GELU, where GELU is applied to only a fraction of neurons within each layer.

{\bf Observations 3: Deeper FFNs require activations more critically than early layers.}
Across all partial configurations, removing activations from deeper layers consistently leads to higher perplexity, indicating that deeper FFNs require activations more than the early layers (Table \ref{tab:PartialActsRemoval}). For instance, 25\% GELU in layers $L$0-$L$2 yields 3.126 PPL versus 3.054 PPL when placed in $L$9-$L$11. This observation is consistent for both GELU and ReLU activations. Future private LLM architectures could exploit this asymmetry for reducing the overheads of FFN activations.


\begin{table}[t]
\centering
\setlength{\tabcolsep}{4pt}
\renewcommand{\arraystretch}{1.02}
\caption{Partial removal of FFN activations: perplexity (PPL) and selective FFN layers with activations 
in GPT-2 (125M) variants. Removing activations from deeper layers results in worse PPL. }
\label{tab:PartialActsRemoval}
\centering 
\resizebox{0.49\textwidth}{!}{
\begin{tabular}{lcccc}
\toprule
Activation & Config 1 & Config 2 & Config 3 & Config 4 \\
\midrule
Full-GELU      & \multicolumn{4}{c}{PPL: 2.688} \\
Full-ReLU      & \multicolumn{4}{c}{PPL: 2.757} \\
\midrule
50\% & \cellcolor{green!15}PPL: 2.903 & \cellcolor{green!15}2.889 & \cellcolor{green!15}2.885 & \cellcolor{green!15}2.877 \\
GELU & L(0,1,2,3,4,5) & L(6,7,8,9,10,11) & L(0,2,4,6,8,10) & L(1,3,5,7,9,11) \\
\cmidrule(lr){2-5}
50\% & \cellcolor{green!15}PPL: 2.961 & \cellcolor{green!15}2.943 & \cellcolor{green!15}2.944 & \cellcolor{green!15}2.931 \\
ReLU & L(0,1,2,3,4,5) & L(6,7,8,9,10,11) & L(0,2,4,6,8,10) & L(1,3,5,7,9,11) \\ \midrule
25\% & \cellcolor{green!15}PPL: 3.126 & \cellcolor{green!15}3.071 & \cellcolor{green!15}3.061 & \cellcolor{green!15}3.054 \\
GELU & L(0,1,2) & L(3,4,5) & L(6,7,8) & L(9,10,11) \\
\cmidrule(lr){2-5}
25\% & \cellcolor{green!15}PPL: 3.147 & \cellcolor{green!15}3.097 & \cellcolor{green!15}3.095 & \cellcolor{green!15}3.093 \\
ReLU & L(0,1,2) & L(3,4,5) & L(6,7,8) & L(9,10,11) \\
\midrule
Fractional & \cellcolor{green!15}PPL: 2.875 & \cellcolor{green!15}3.040 & \cellcolor{green!15}3.161 & \cellcolor{green!15}3.243 \\
GELU & 50\% neurons & 25\% neurons & 12.5\% neurons & 6.25\% neurons \\ \midrule
No-Acts        & \multicolumn{4}{c}{PPL: 3.376} \\ 
\bottomrule
\end{tabular}}
\end{table}


\subsection{Softmax-only Architecture}

Having shown the viability of selective activation removal, we now examine the extreme minimal-nonlinearity configuration: eliminating all activations and normalizations, except the Softmax in attention sub-block. This allows us to verify architectural and algorithmic solutions to linearize private LLMs  under the most constrained nonlinear regime.

\begin{figure} [t]
\centering
\subfloat[Layerwise NaNs \label{subfig:LayerwiseNaNs}]{\includegraphics[width=.24\textwidth]{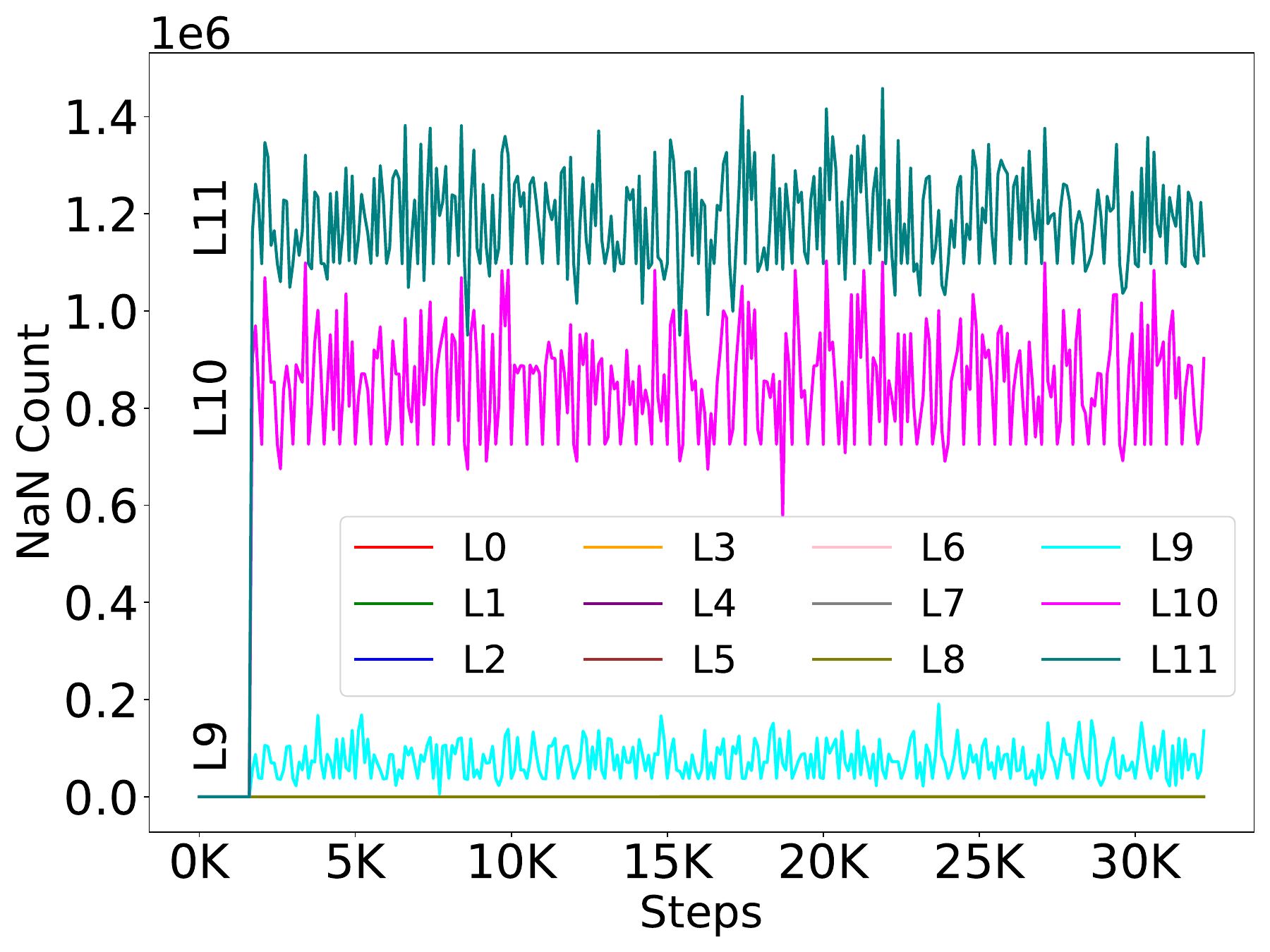}} 
\subfloat[ Entropy heatmap \label{subfig:EntHmapSM}]{\includegraphics[width=.24\textwidth]{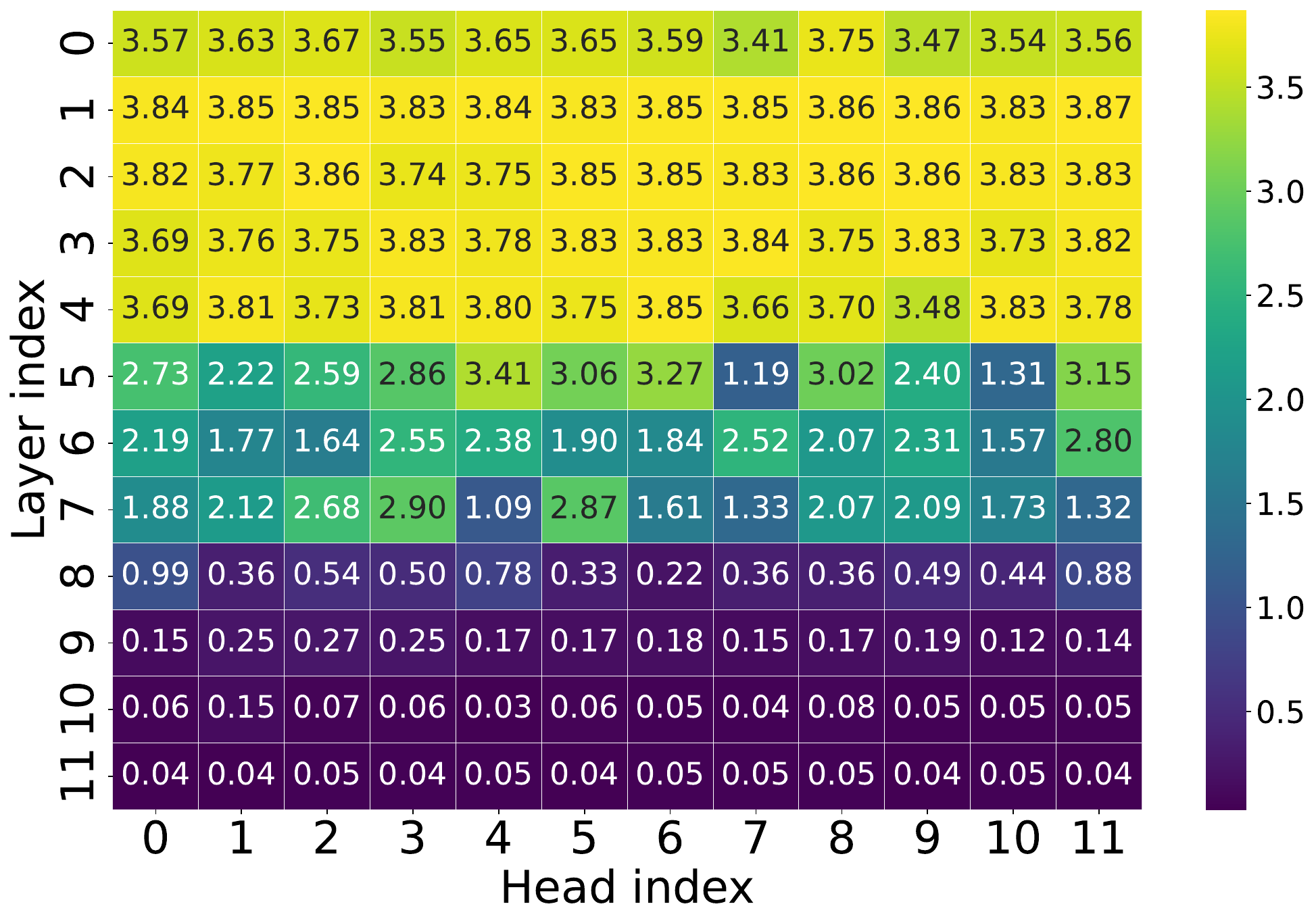}} 
\vspace{-1em}
\caption{Training collapses in softmax-only GPT-2 model.} 
\label{fig:SoftmaxNaN}
\end{figure}

{\bf Observation 4: The softmax-only model exhibits severe entropic overload in the early layers and entropy collapse in the deeper layers.} 
When FFN nonlinearity is removed in a normalization-free architecture, leaving Softmax as the sole source of nonlinearity and the FFN purely linear, the resulting design is referred to as a Softmax-only model. However, training such a model is challenging, as loss values in deeper layers quickly diverge to NaNs  (Figure \ref{subfig:LayerwiseNaNs}). The entropic analysis further reveals that the deeper layers suffers from entropy collapse, marked by extremely low entropy values (see blue regions in Figure \ref{subfig:EntHmapSM}), a known indicator of training instability \citep{zhai2023stabilizing}. Moreover, 45\% of total heads exhibit entropy values in the range of $\frac{3\text{max}}{4}$ to ${\tt max}$, with most close to the maximum value (Figure \ref{fig:FractionAttentionHeads}) in the early layers, indicating severe entropic overload.


\section{AERO} \label{sec:AERO}

We introduce AERO, an entropy-guided architectural framework that remove nonlinearities and reduces FLOPs through principled refinements in LLM architectures, without de-stabilizing them. Further, we propose a novel entropy regularization technique to prevent entropic overload, by aligning the model's entropy dynamics closer to the well-behaved attention entropy distribution in MHA.

\subsection{Inference-efficient Substitute for LayerNorm} \label{subsec:StaticNorm}
To address training instability, prior work has predominantly relied on LayerNorm applied to various parts of the network, such as QK-LayerNorm \cite{dehghani2023scaling,wortsman2024smallscale,muennighoff2024olmoe} and FFN-LayerNorm \citep{rybakov2024methods}. Since LayerNorm requires expensive inverse-square-root operations during inference \cite{hou2023ciphergpt},  we explore the weight and other static normalization alternatives  to avoid nonlinear computations at inference. 



{\bf Normalizing weights in FFN linear layers prevent entropy collapse and stabilizes training.} 
We find that weight normalization \cite{salimans2016weight} and spectral normalization \cite{miyato2018spectral} offer inference-efficient alternatives to LayerNorm by normalizing weights rather than activations. When applied {\em strategically}, they can effectively prevent entropy collapse in the deeper layers of Softmax-only LLMs, without incurring additional inference overhead in privacy-preserving computation settings. Notably, the effectiveness of these methods varies with the targeted linear layers: applying normalization within the attention sub-block results in {\em higher} perplexity compared to its application within the FFN (see Appendix \ref{Appendix:WNormSNorm}).


Further, we employ a more simpler approach to stabilize the training of Softmax-only models. During training FFN sub-blocks learn their individual  (down)scaling factors for FFN output and its residual output, as follows (see Eq. \ref{eqn:ffn_mha}):

\vspace{-2em}
\begin{equation} \label{eqn:ScaledFFN}
\mathbf{X}_{\text{out}} = \beta\hat{\mathbf{X}}_{\text{SA}} +\frac{1}{\alpha} (\text{FFN}^{\text{SM}}(\mathbf{X}_{\text{SA}})) \; \text{where} \; \alpha, \beta \in \mathbb{R}^{L}
\end{equation}

\vspace{-1em}

\begin{figure} [t]
\centering
\subfloat[WeightNorm]{\includegraphics[width=.165\textwidth]{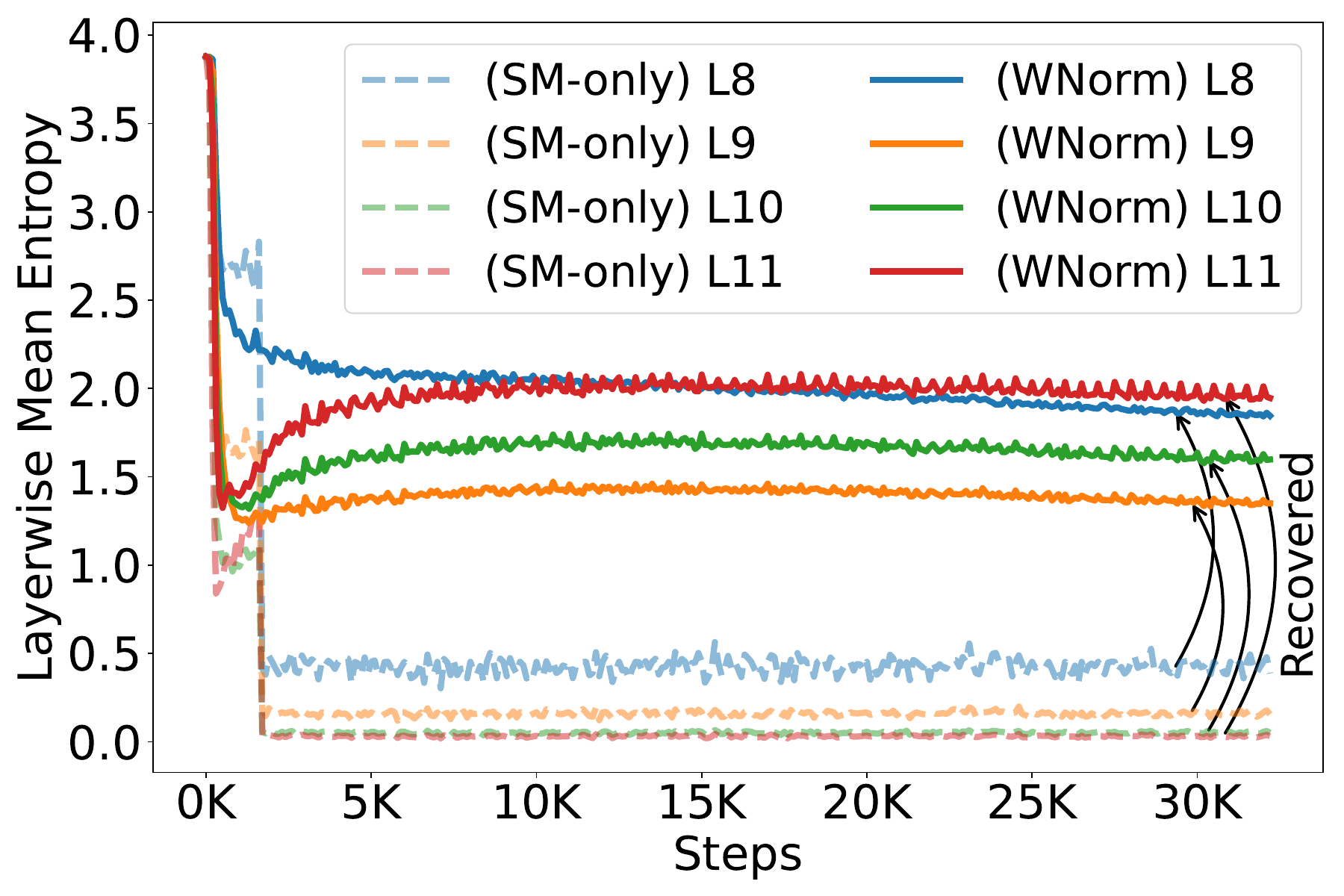}} 
\subfloat[SpectralNorm]{\includegraphics[width=.165\textwidth]{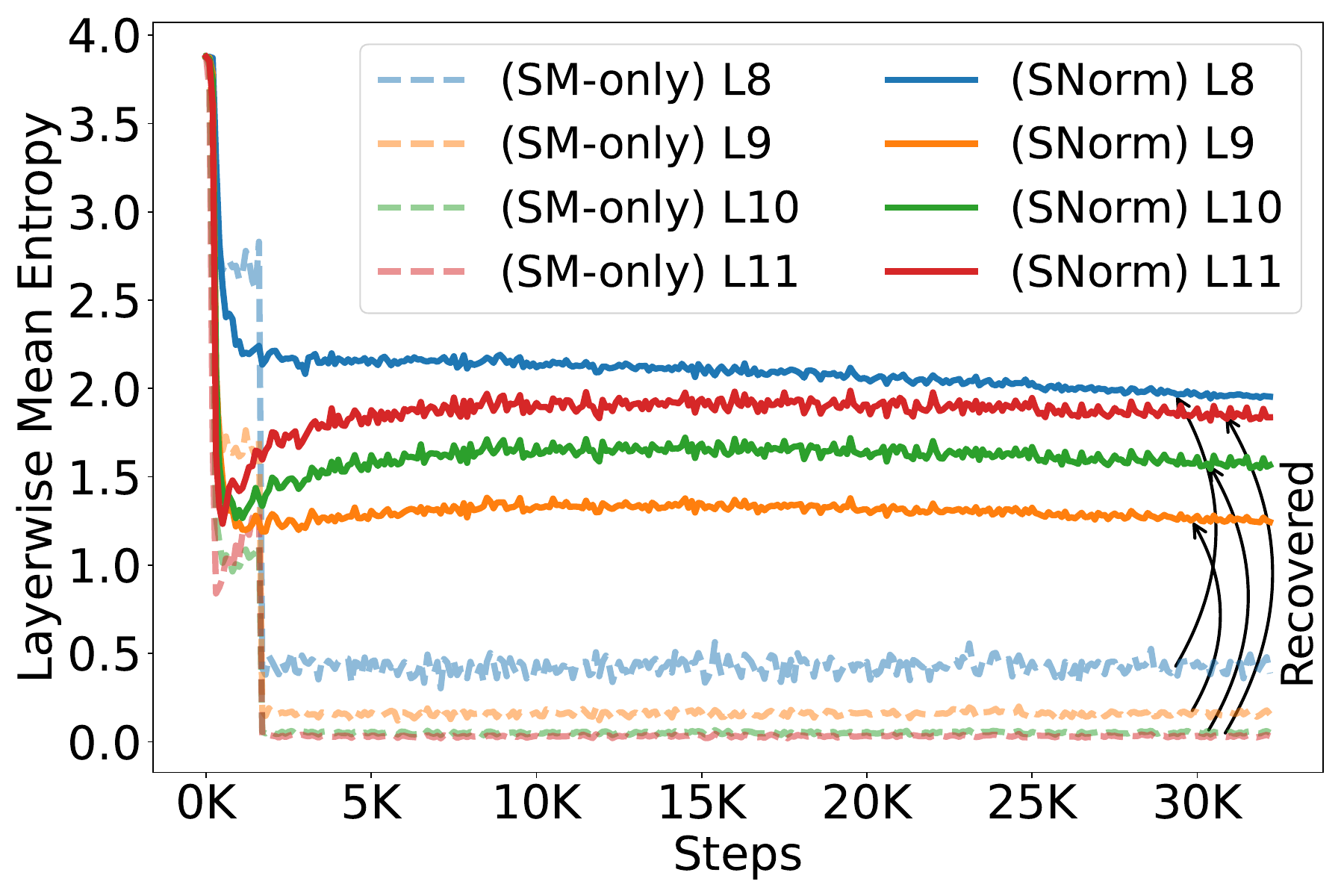}} 
\subfloat[ScaledFFN]{\includegraphics[width=.165\textwidth]{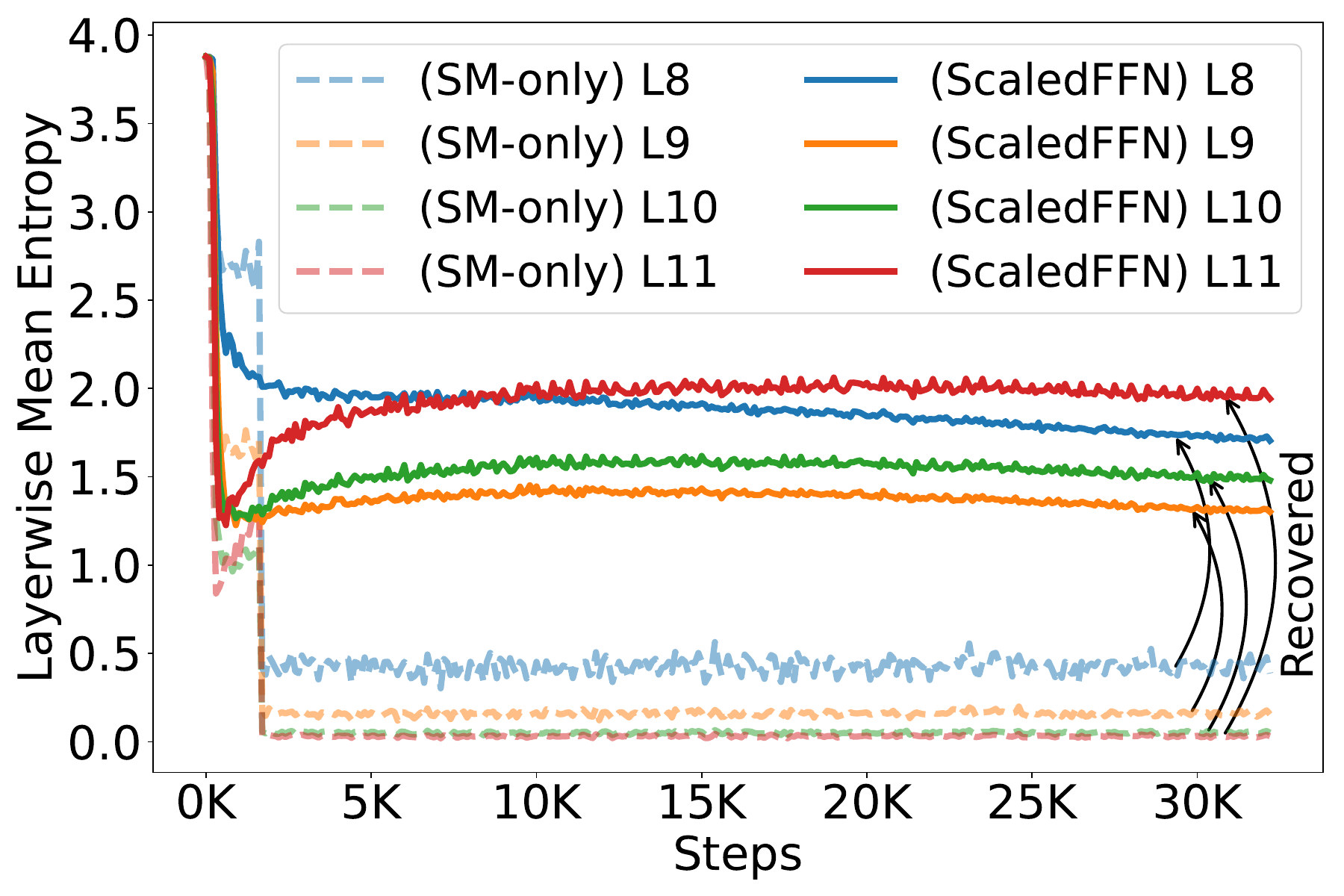}}   \vspace{-.8em}
\caption{Inference-efficient static normalization methods mitigate entropy collapse in the deeper layers of a softmax-only models, showing its utility for LayerNorm substitute for private LLMs.} 
\label{fig:RecoveredEntCollapse}
\end{figure}

Figure \ref{fig:RecoveredEntCollapse} illustrates how these normalization techniques stabilize training by preventing entropy collapse in deeper layers of softmax-only  models. Learnable scaling in FFNs exhibits better performance (lowest perplexity) compared to weight and spectral normalization (Table \ref{tab:SNormVsWNormVsMlpGains}). 


\begin{table} [t]
\caption{Perplexity comparison of inference-efficient static normalization methods employed in FFNs of Softmax-only GPT-2.} 
\label{tab:SNormVsWNormVsMlpGains}
\centering
\begin{tabular}{cccc} \toprule 
& WeightNorm & SpectralNorm & ScaledFFN\\ \toprule 
Eval PPL & 3.640 & 3.624 & 3.478 \\ \bottomrule 
\end{tabular} 
\vspace{-0.5em}
\end{table}



\subsection{Entropy Regularization} \label{subsec:entropyreg}

{\bf Challenges in designing entropy regularization schemes to prevent entropic overload.} 
Previous entropy regularization approaches have primarily aimed at penalizing low-entropy predictions \citep{setlur2022maximizing, pereyra2017regularizing}, based on the principle of maximum entropy \citep{jaynes1982rationale}. Recently, \citet{he2024understanding} introduced entropy regularization in LLM to address extremely low entropy values. 

However, our goal is to regularize higher entropy values, which introduces two key challenges: (1) {\em Head specialization:} Since each attention head captures different aspects of the input, the regularization strength needs to be adjusted for each head individually. (2) {\em Over-regularization:} Some heads naturally exhibit higher entropy, thus, blanket penalization of all higher entropy values could be detrimental, requiring a more flexible approach.

{\bf Key design principles for entropy regularization.} To address aforementioned challenges, our regularization scheme (see Algorithm \ref{Algo:EntRegLossComputation}) is guided by following principles:

\begin{itemize}[noitemsep,nolistsep,leftmargin=0.3cm] \vspace{-0.5em}
\item {\em Balanced entropy distribution with parameterized attention matrix:} Inspired by \citet{miller1996global}, which used temperature parameter as a Lagrangian multiplier to control the entropy of a stochastic system, we parameterized the attention matrix by a learnable temperature \( t \in \mathbb{R}^{H \times T} \) for each softmax operation, allowing the model to adjust the sharpness of the attention scores. A higher temperature (\( t > 1 \)) diffuses attention scores and increases the entropy, while a lower temperature (\( t < 1 \)) sharpen attention scores and reduces the entropy.



\item {\em Dynamic thresholds with head-specific adaptation: } To adapt the regularization strength based on the characteristics of each attention head \citep{voita2019analyzing},  we use headwise learnable threshold parameter $\mathtt{reg\_threshold\_weights} \in \mathbb{R}^H$. Consequently, the threshold for each head is computed as a learnable fraction of the maximum value of entropy ($\mathtt{reg\_threshold\_weights} \times \text{E}_{\text{max}}$), providing the fine-grained control (see Algorithm \ref{Algo:EntRegLossComputation}, line \#\ref{line:LearnableRegThreshold}).

\item {\em Tolerance margin to prevent over-regularization: }
To prevent over-regularization, we allow small deviations from the respective thresholds. Thus, a penalty is imposed only if the deviation from the threshold exceeds the tolerance margin, which is set as a fraction of \(\text{E}_{\text{max}}\) using the hyper-parameter \(\gamma\) (see Algorithm\ref{Algo:EntRegLossComputation}, line \#\ref{line:ToleranceMargin}). 

\vspace{-2em}

\begin{equation} \label{eqn:EntThersDeviation}
\text{penalty}^{(l,h)} \! \!= \! \!\begin{cases} \!\!
\Big(\! \text{deviation}^{(l,h)} \Big)^2 \!\!\!\!\!&\!\!\! \text{iff} \big| \text{deviation}^{(l,h)} \big| \!\! >\!\! \gamma E_{\text{max}} \\ 
0 & \!\!\!\text{otherwise}
\end{cases} 
\end{equation}

The deviation from threshold is computed as $\text{deviation}^{(l,h)} = \text{E}^{(l,h)}(t) -  \theta^{(l,h)} \text{E}_{\text{max}}$, where $\theta^{(l,h)}$ is $\mathtt{reg\_threshold\_weights}$. The hyper-parameter \(\gamma\) ensures that the model is not excessively penalized for minor deviations from the desired entropy threshold, which could impede its capacity to learn effectively. This careful calibration between stringent regularization and desired flexibility improves the model's robustness while maintaining its adaptability to various input distributions.

\item {\em Maximum entropy reference:} We set \(E_{\text{max}} = \log(T)\) as a reference for thresholds and tolerance margins, ensuring consistency across layers and heads. 


\end{itemize}


\begin{figure} [t]
\centering
\includegraphics[width=.49\textwidth]{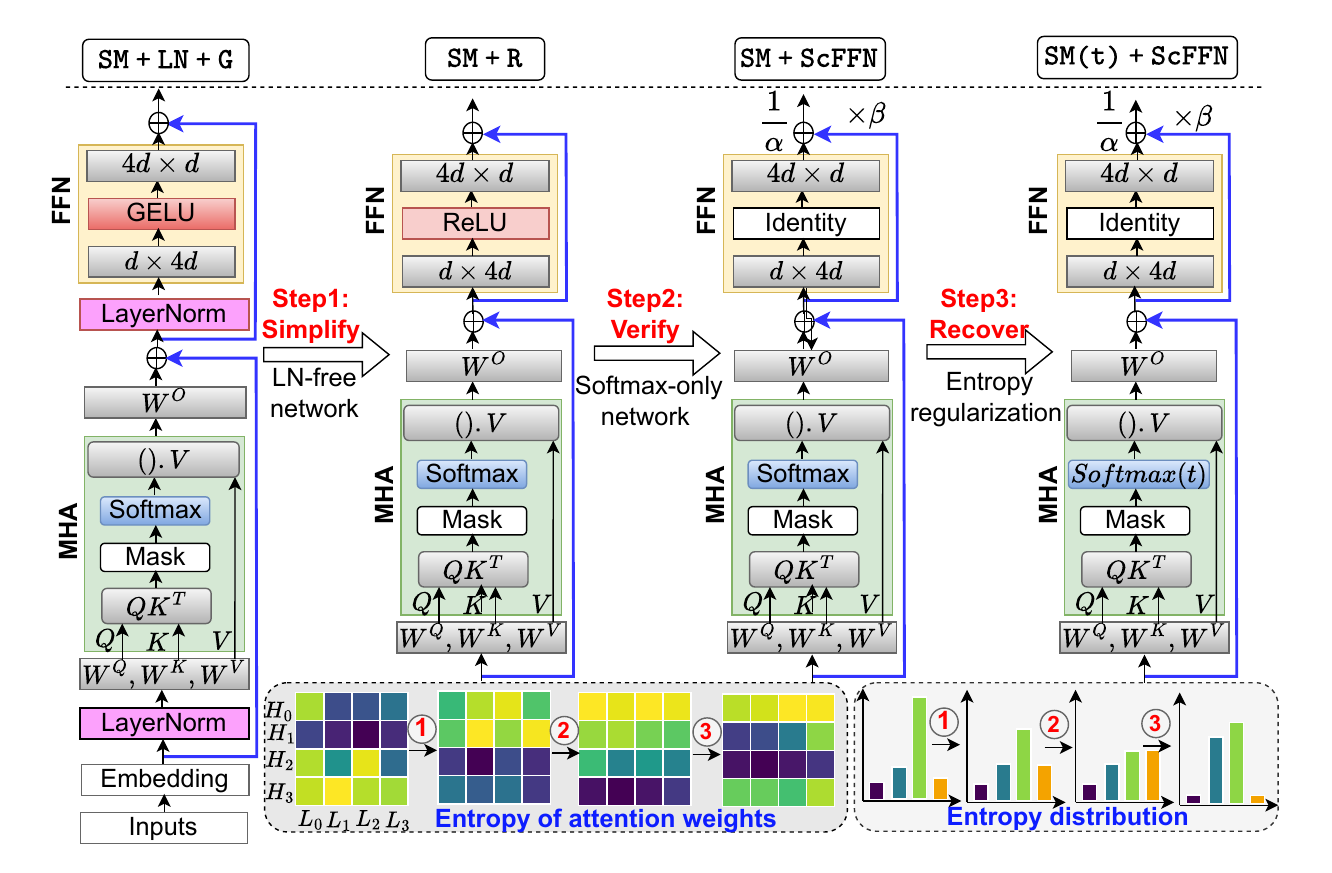} \vspace{-2.5em}
\caption{Overview of the AERO framework for efficient private LLM design. Step 1 simplifies the architecture by removing LayerNorm and substituting GELU with ReLU in FFNs. Step 2 validates our approach through softmax-only design. Step 3 employs entropy regularization to penalize extreme entropy values.} 
\label{fig:EndToEndAERO}
\end{figure}


\subsection{Putting it All Together}

The AERO framework addresses private inference inefficiencies through targeted architectural refinements and algorithmic innovations. Given an input LLM, {\bf Step 1} tackles communication overhead by identifying by identifying {\tt SM+R}, a configuration that strategically removes LayerNorm and substitute GELU with ReLU in FFNs. {\bf Step 2} validates these choices through extreme stress testing: the softmax-only architecture which highlights the practical utility of architectural simplification.  Finally, {\bf Step 3} provides the critical algorithmic component, adaptive entropy regularization that maintains attention diversity by penalizing extreme distributions in the absence of nonlinearities.

\section{Evaluation}  \label{sec:evaluation}

{\bf System setup}
We use a SecretFlow setup \citep{lu2023bumblebee} with the client and server simulated on two physically separate machines, each equipped with an AMD EPYC 7502 server with specifications of 2.5 GHz, 32 cores, and 256 GB RAM. We measure the {\em end-to-end} PI latency, including input embeddings and final output (vocabulary projection) layers, in WAN setting (bandwidth:100Mbps, latency:80ms),  simulated using Linux Traffic Control (tc) commands. The number of threads is set to 32. Following \cite{he2024simplifying,stanic2023languini,geiping2023cramming}, all the models are trained on a single RTX 3090 GPU.

{\bf Models and datasets} We conducted experiments on GPT-2 models (12 and 18 layers) and Pythia, training them from scratch on the CodeParrot and Languini book datasets---standard benchmarks for LLM evaluation \citep{he2024simplifying,he2024understanding}. CodeParrot dataset is sourced from 20 million Python files, contains 8 GB of files, a total of 2.1 billion training tokens. We use a tokenizer with a vocabulary of 50K and train with context lengths of 128 and 256. The Languini book dataset includes 84.5 GB of text, and 23.9 billion tokens with a WikiText-trained vocabulary of 16,384, and we train with context of 512 tokens. 


{\bf Training hyperparameters}
For pre-training on the CodeParrot dataset, we adopt the training settings from \citep{he2024simplifying}. Similarly, for training on the Languini dataset, we follow the settings from \citep{stanic2023languini}. These settings remain consistent across all architectural variations to accurately reflect the impact of the architectural changes. When applying entropy regularization on the CodeParrot dataset, we initialize the learnable temperature to 1e-2 and set $\lambda$ to 1e-5. For the Languini dataset, the temperature is initialized to 1e-1, and $\lambda$ is set to 5e-5.

\subsection{Effectiveness of Entropy Regularization}

\begin{figure} [htbp]
\centering
\subfloat[{\scriptsize WeightNorm(FFN)}]{\includegraphics[width=.165\textwidth]{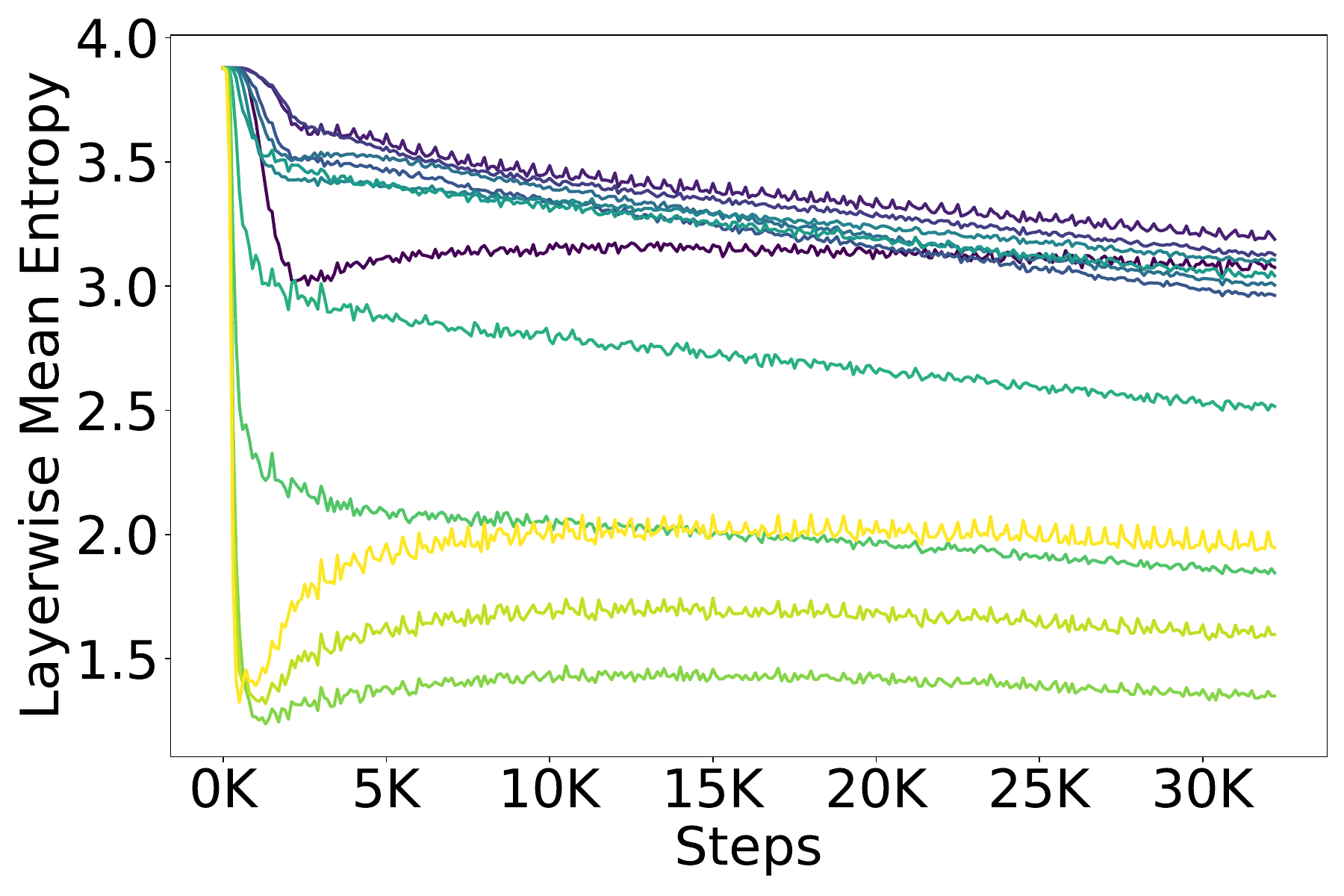}} 
\subfloat[{\scriptsize SpectralNorm(FFN)}]{\includegraphics[width=.165\textwidth]{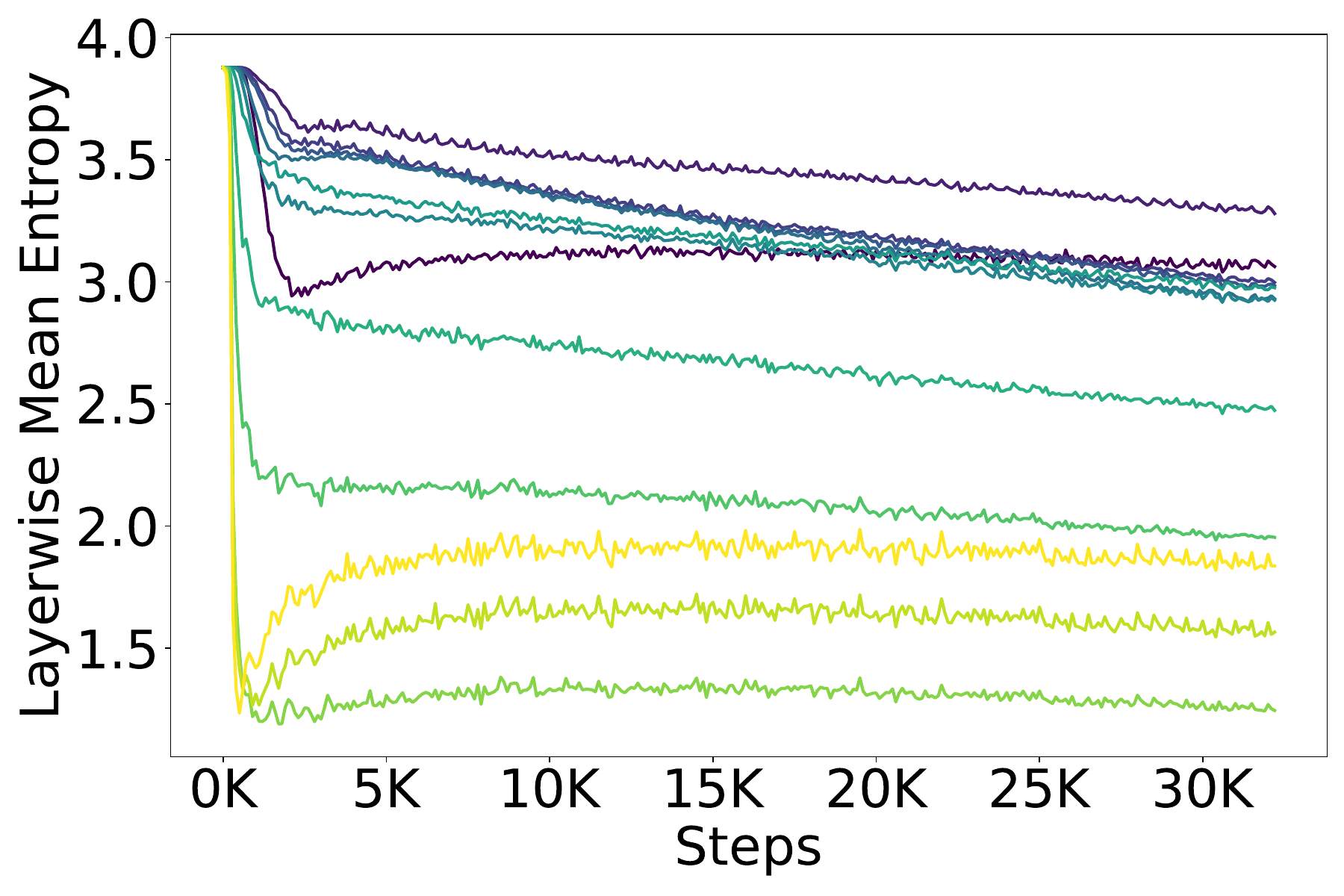}} 
\subfloat[{\scriptsize Scaled-FFN} ]{\includegraphics[width=.165\textwidth]{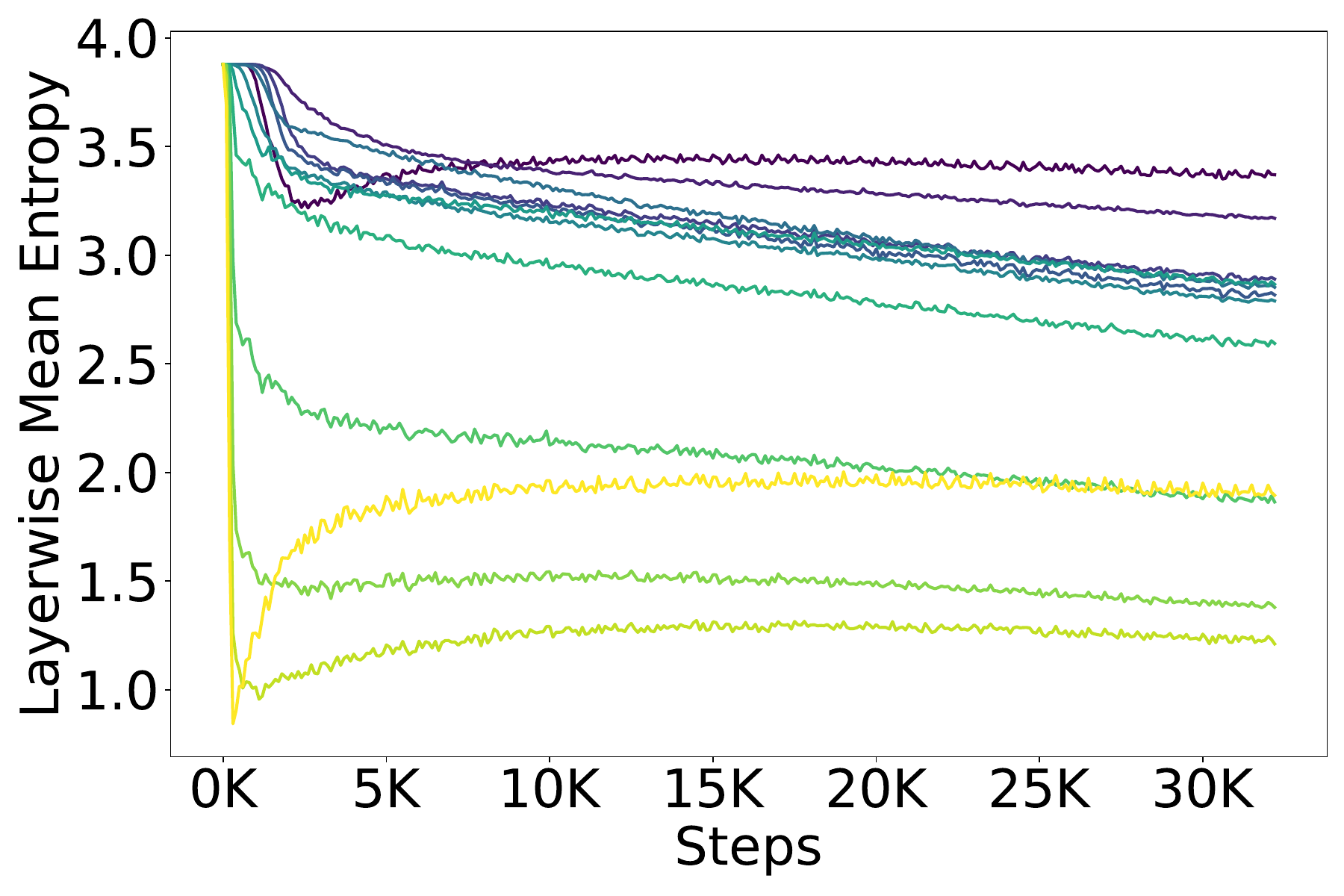}} \\ \vspace{-0.8em}
\subfloat[{\scriptsize EReg(M = 0))}]{\includegraphics[width=.165\textwidth]{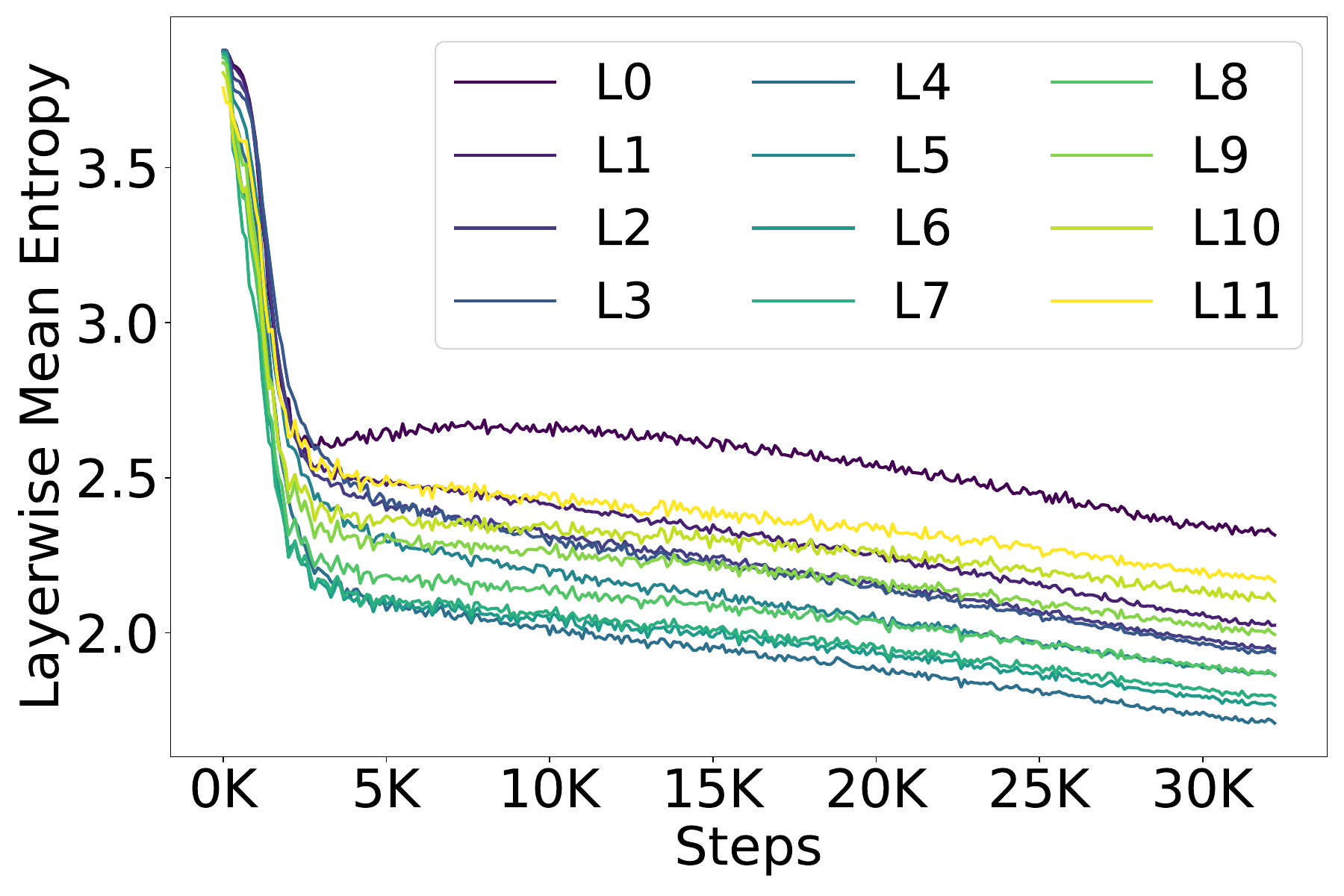}} 
\subfloat[{\scriptsize EReg(M = \(0.1\text{E}_{\text{max}} \))}]{\includegraphics[width=.165\textwidth]{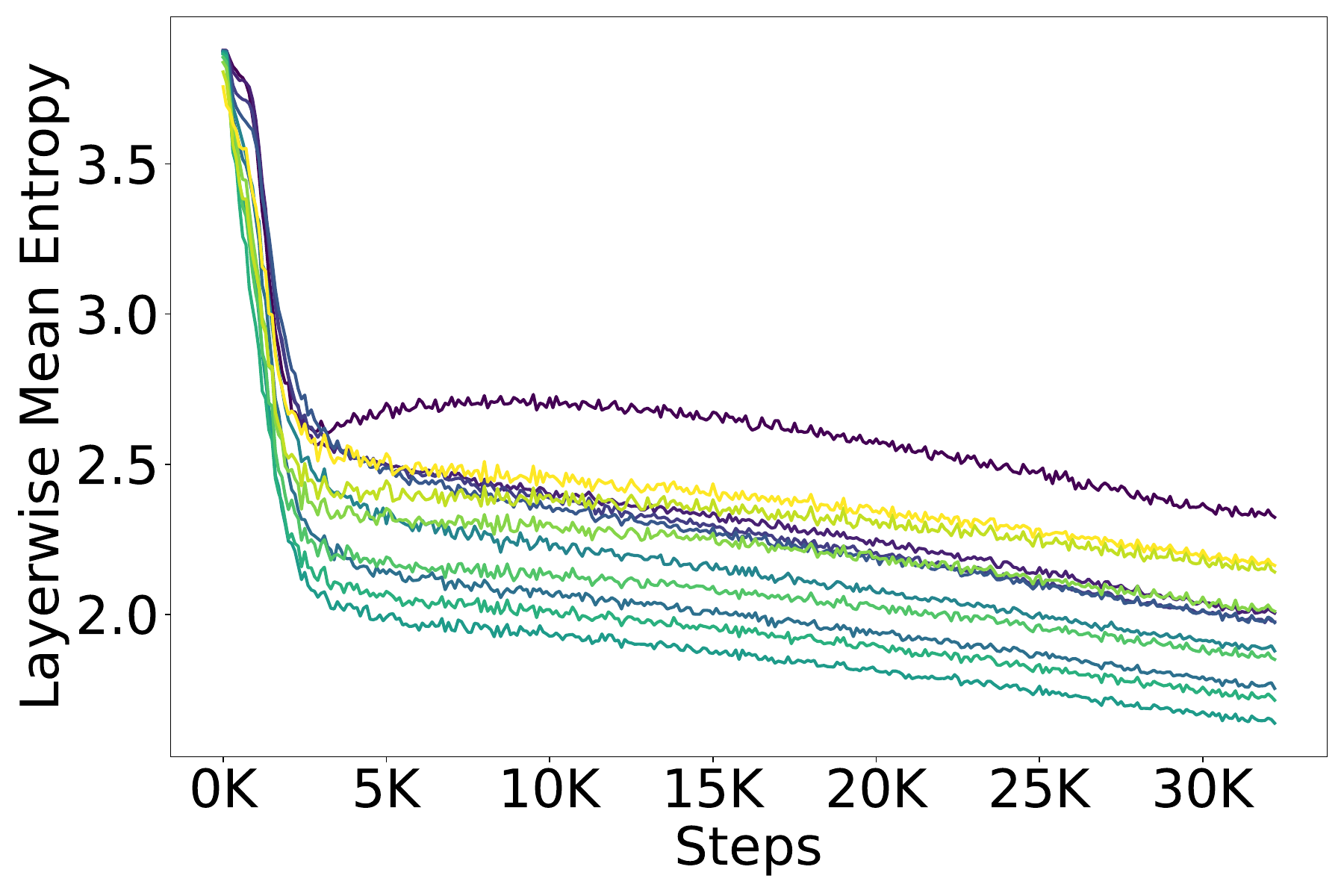}} 
\subfloat[{\scriptsize EReg(M = \(0.2\text{E}_{\text{max}} \))}]{\includegraphics[width=.165\textwidth]{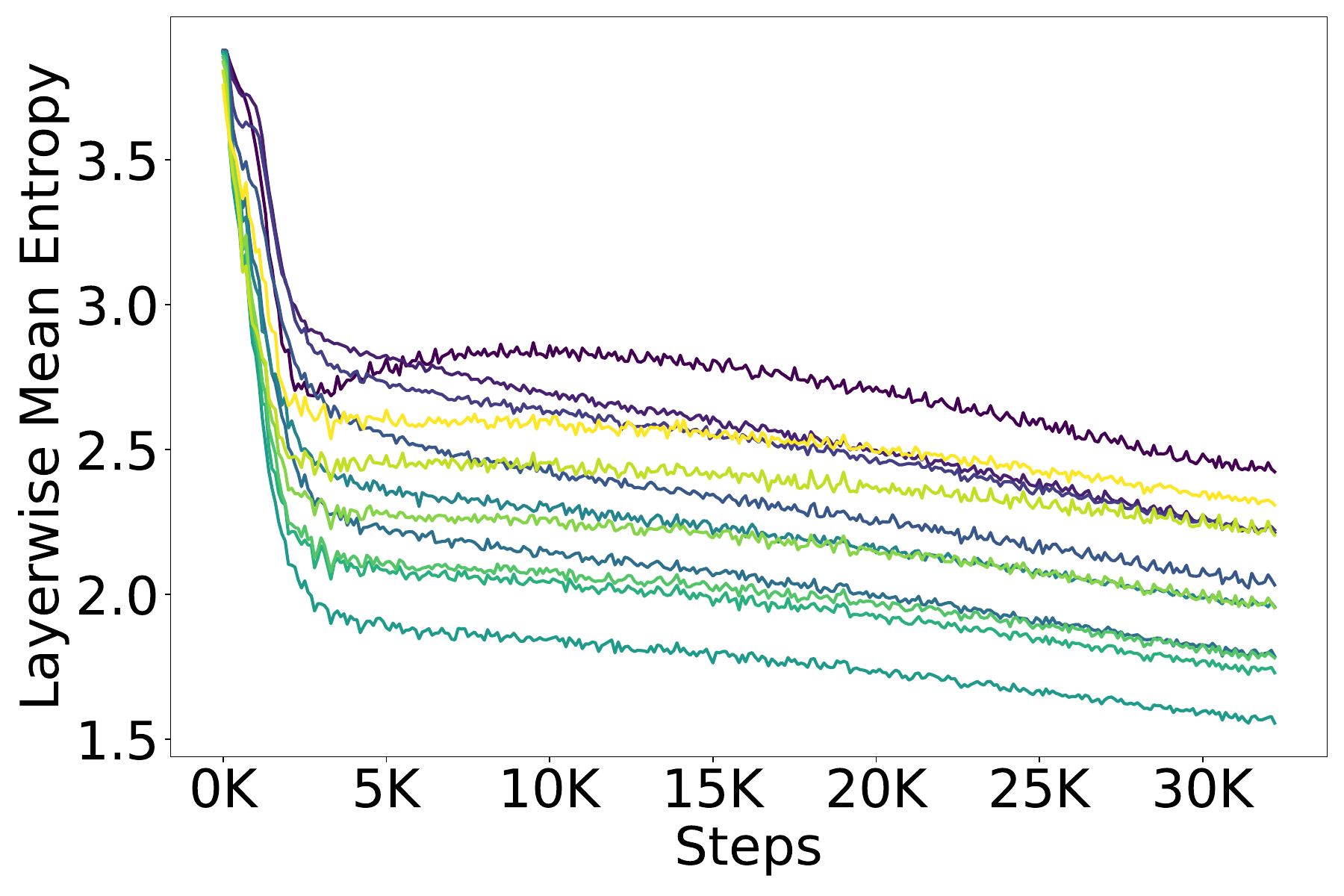}} \vspace{-0.5em}
\caption{Entropy dynamics of a Softmax-only GPT-2 model trained from scratch on 2.1B tokens. {\bf Top:} Weight normalization, spectral normalization, and learnable scaling  in FFNs {\em alone} do not prevent entropic overload, indicated by persistent higher entropy in early layers. {\bf Bottom:} Entropy regularization mitigates this overload, while a non-zero tolerance margin avoids over-regularization.} 
\label{fig:EntropyWSNorm}
\end{figure}

{\bf Entropy regularization prevents entropic overload} 
While  weight/spectral normalization, and learnable scaling normalization in FFNs effectively prevent entropy collapse in the deeper layers and stabilize the training of Softmax-only models, they {\em fail to address the issue of entropic overload}, (Figure \ref{fig:EntropyWSNorm}). In contrast, the entropy regularization scheme penalizes the model to avoid extreme entropy values during training, resulting in a more balanced distribution. Hence, it complements the training stabilizing methods by further mitigating entropic overload in the early layers (Figure \ref{fig:EntropyWSNorm}), improving the attention heads utilization and perplexity.

\begin{figure} [t] \centering
\subfloat[Learned threshold weights \label{fig:EntLearnableThreshold}]{\includegraphics[width=.24\textwidth]{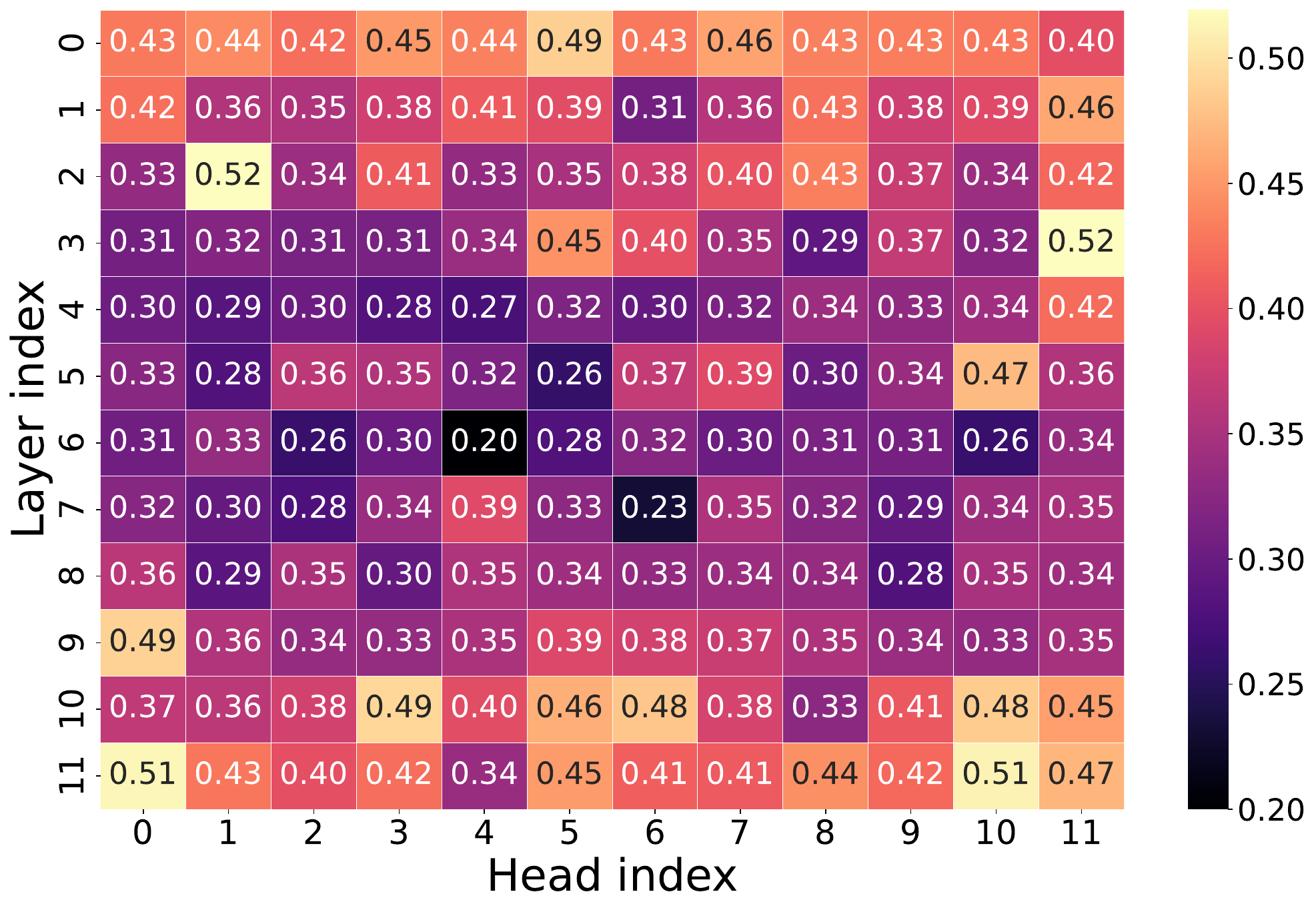}} 
\subfloat[Layerwise mean and variance]{\includegraphics[width=.24\textwidth]{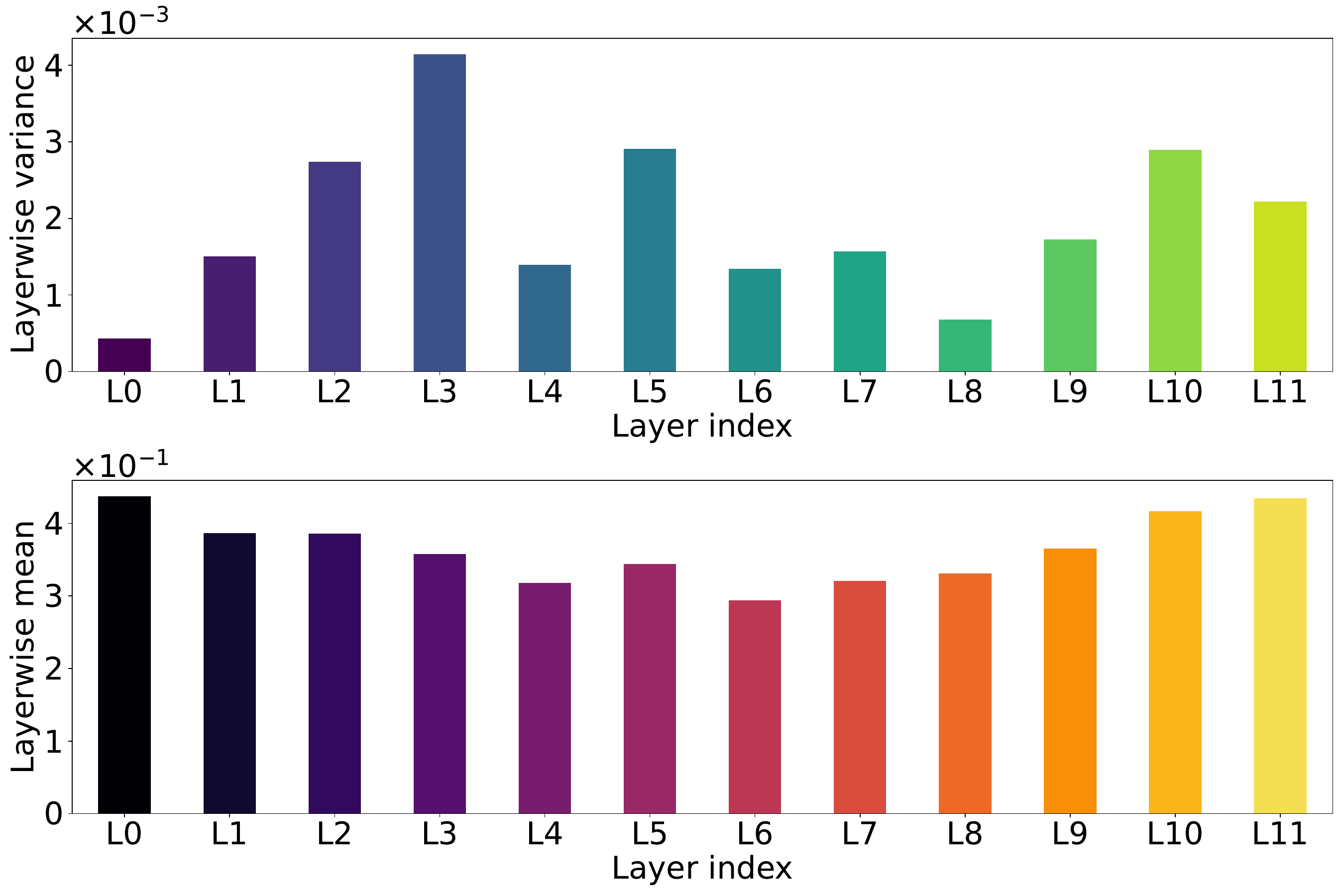}} \\ \vspace{-0.6em}
\caption{Learned threshold weights in entropy regularization (Eq. \ref{eqn:EntThersDeviation}) in a trained softmax-only GPT-2 model: Attention heads adaptively learned nonuniform threshold across heads and layers.} 
\label{fig:LearnableThreshold}
\end{figure}

{\bf Learnable thresholds help preserve the attention head diversity and tolerance margins prevent over-regularization}
Figure \ref{fig:EntLearnableThreshold} depicts threshold parameters ($\mathtt{reg\_threshold\_weights}$) of entropy regularization scheme employed in Softmax-only models, after the model has been trained. They exhibit significant variability,  both across layers and within individual heads of each layers, which reflects the model's ability to dynamically adjust the regularization strength in tailored to the head-specific roles.

Figures \ref{fig:EntropyWSNorm} (bottom) shows the effect of nonzero tolerance margins in entropy regularization. Higher margins allow a small yet crucial fraction of early-layer attention heads to attain higher entropy values, desirable for well-behaved entropy distribution. Specifically, as shown in Figure \ref{fig:EntDistTmargin}, $\gamma$ increases from 0 to 0.15, only 0.7\% of attention heads fall in the highest entropy range, but this fraction rises to 2.08\%, 3.47\%, and 6.25\% at $\gamma = 0.20, 0.25,$ and $0.30$, respectively.

\subsection{Practical Utility of AERO for Private LLM Design}

{\bf Efficiency gains without performance loss} The practical application of AERO hinges on the efficiency gain through nonlinearity reduction while  recovering performance loss through entropy regularization.  Table \ref{tab:GPT2CLen128} shows that AERO enables the {\tt SM+R} configuration to achieve {\bf 3.41$\times$} communication reduction and {\bf 1.42$\times$} speedup while fully recovering performance through entropy regularization (PPL: 2.66 vs 2.69 baseline). However, removing entropy regularization significantly drops the performance (2.94 PPL). Table \ref{tab:SensitivityTMargins} shows the mechanism driving this performance recovery. ReLU variants exhibit strong sensitivity to tolerance margins, improving from 2.924 to 2.688 PPL (at TM = 0.10), an 8\% gain, whereas GELU reduces PPL only by 1.5\%.

\begin{table}[t]
\centering
\caption{Results, and comparison against SOTA \citep{he2024simplifying}, when  GPT-2 ($L$=12, $H$=12, $d$=768) model is trained from scratch on CodeParrot  dataset with context length  128.} 
\label{tab:GPT2CLen128}

\setlength{\tabcolsep}{5pt} 
\renewcommand{\arraystretch}{0.96} 

\resizebox{0.49\textwidth}{!}{
\begin{tabular}{p{0.2cm}lclcccc}
\toprule
& \multirow{2}[4]{*}{Network Arch.} & \multirow{2}[4]{*}{PPL} & \multirow{2}[4]{*}{\#Nonlinear Ops} & \multirow{2}[4]{*}{\shortstack{Comm.\\(GB)}} & \multirow{2}[4]{*}{\shortstack{Lat.\\(min.)}} & \multicolumn{2}{c}{Savings} \\
\cmidrule(lr){7-8}
& & & & & & Comm. & Lat. \\
\midrule
\multirow{6}{*}{  \rotatebox[origin=c]{90}{Baseline} } & \multirow{3}{*}{${\tt SM + LN + G}$} & \multirow{3}{*}{2.69} & SM:$144\times\mathbb{R}^{128\times128}$ & \multirow{3}{*}{25.32} & \multirow{3}{*}{8.21} & \multirow{3}{*}{1$\times$} & \multirow{3}{*}{1$\times$} \\
& & & LN:$24\times\mathbb{R}^{128\times768}$ & & & & \\
& & & G:$12\times\mathbb{R}^{128\times3072}$ & & & & \\ \cline{2-8}
& \multirow{3}{*}{${\tt SM + LN + R}$} & \multirow{3}{*}{2.76} & SM:$144\times\mathbb{R}^{128\times128}$ & \multirow{3}{*}{9.44} & \multirow{3}{*}{6.06} & \multirow{3}{*}{2.68$\times$} & \multirow{3}{*}{1.35$\times$} \\
& & & LN:$24\times\mathbb{R}^{128\times768}$ & & & & \\
& & & R:$12\times\mathbb{R}^{128\times3072}$ & & & & \\
\midrule
\multirow{4}{*}{  \rotatebox[origin=c]{90}{Norm-free} } & \multirow{2}{*}{${\tt SM + G}$} & \multirow{2}{*}{3.20} &  SM:$144\times\mathbb{R}^{128\times128}$ & \multirow{2}{*}{23.31} & \multirow{2}{*}{7.86} &  \multirow{2}{*}{1.09$\times$} & \multirow{2}{*}{1.04$\times$} \\
& & & G:$12\times\mathbb{R}^{128\times3072}$ & & & &  \\ \cline{2-8}
&\multirow{2}{*}{${\tt SM + R}$}  & \multirow{2}{*}{2.94} &  SM:$144\times\mathbb{R}^{128\times128}$ & \multirow{2}{*}{7.43} & \multirow{2}{*}{5.78}  & \multirow{2}{*}{3.41$\times$} & \multirow{2}{*}{1.42$\times$}\\
& & & R:$12\times\mathbb{R}^{128\times3072}$ & & & & \\ 
\midrule
\multirow{6}{*}{  \rotatebox[origin=c]{90}{SOTA} } & \multirow{3}{*}{${\tt SM + G}$} & \multirow{3}{*}{2.93} &  SM:$144\times\mathbb{R}^{128\times128}$ & \multirow{3}{*}{23.19} & \multirow{3}{*}{7.40} &  \multirow{3}{*}{1.09$\times$} & \multirow{3}{*}{1.11$\times$} \\
& & & G:$12\times\mathbb{R}^{128\times3072}$ & & & & \\
& & & LN: $1\times\mathbb{R}^{128\times768}$ & & & & \\ \cline{2-8}
&\multirow{3}{*}{${\tt SM + R}$}  & \multirow{3}{*}{3.03} &  SM:$144\times\mathbb{R}^{128\times128}$ & \multirow{3}{*}{7.30} & \multirow{3}{*}{5.55}  & \multirow{3}{*}{3.47$\times$} & \multirow{3}{*}{1.48$\times$}\\
& & & R:$12\times\mathbb{R}^{128\times3072}$ & & & & \\
& & & LN: $1\times\mathbb{R}^{128\times768}$ & & & & \\ 
\midrule
\multirow{4}{*}{  \rotatebox[origin=c]{90}{\bf AERO} } & \multirow{2}{*}{$\text{EReg}({\tt SM + G})$} & \multirow{2}{*}{3.07} &  SM:$144\times\mathbb{R}^{128\times128}$ & \multirow{2}{*}{23.31} & \multirow{2}{*}{7.86} &  \multirow{2}{*}{1.09$\times$} & \multirow{2}{*}{1.04$\times$} \\
& & & G:$12\times\mathbb{R}^{128\times3072}$ & & & &  \\ \cline{2-8}
&\multirow{2}{*}{$\text{EReg}({\tt SM + R})$}  & \multirow{2}{*}{\bf 2.66} &  SM:$144\times\mathbb{R}^{128\times128}$ & \multirow{2}{*}{7.43} & \multirow{2}{*}{5.78}  & \multirow{2}{*}{\bf 3.41$\times$} & \multirow{2}{*}{\bf 1.42$\times$}\\
& & & R:$12\times\mathbb{R}^{128\times3072}$ & & & & \\ 

\bottomrule
\end{tabular}}
\end{table}

\begin{table}[t]
\centering
\caption{ReLU exhibits higher sensitivity to tolerance margins than GELU in LayerNorm-free GPT-2 models (8\% vs 1.5\% variation)
} \label{tab:SensitivityTMargins}
\resizebox{0.49\textwidth}{!}{
\begin{tabular}{lccc|ccc}
\toprule
& \multicolumn{3}{c}{EReg(SM+G)} & \multicolumn{3}{c}{EReg(SM+R)} \\
\cmidrule(lr){2-4} \cmidrule(lr){5-7}
& TM=0 & TM=0.10 & TM=0.20 & TM=0 & TM=0.10 & TM=0.20 \\
\midrule
Eval PPL & 3.113 & 3.100 & {\bf 3.067} & 2.924 & {\bf 2.658} & 2.886 \\
\bottomrule
\end{tabular}}
\end{table}


When compared to state-of-the-art LayerNorm-free method \citep{he2024simplifying}, AERO achieves comparable communication savings without retaining any LayerNorm layers and delivers {\bf 14\%} lower perplexity (2.66 vs 3.03). Notably, their ReLU variant yields higher perplexity, suggesting that the underlying optimization is not tailored for PI efficiency.


\subsection{Scalability of AERO}

{\bf Scalability across training scales}
We perform experiments on the Languini dataset across varying token counts (1.2B to 4.8B). We observed consistent improvements in perplexity across the training scale (Table \ref{tab:LanguiniGPT2}), when entropy regularization is employed on most-stringent nonlinearity configuration (Softmax-only), whereas SOTA exhibits training instability even at the lower training scale (1.2B).

\begin{table}[t]
\centering
\caption{GPT-2 ($L$=12, $H$=12, $d$=768) trained from scratch on Languini \citep{stanic2023languini} dataset with context length  512. NaNs indicate training instability in SOTA \citep{he2024simplifying}.}
\label{tab:LanguiniGPT2}

\setlength{\tabcolsep}{4pt} 
\renewcommand{\arraystretch}{0.95} 

\resizebox{0.49\textwidth}{!}{
\begin{tabular}{p{0.2cm}lccclcc}
\toprule
& \multirow{2}[4]{*}{Network Arch.} & \multicolumn{3}{c}{Eval PPL} & \multirow{2}[4]{*}{\#Nonlinear Ops} & \multirow{2}[4]{*}{\shortstack{Comm.\\(GB)}} & \multirow{2}[4]{*}{\shortstack{Lat.\\(min.)}} \\
\cmidrule(lr){3-5}
& & 1.2B & 2.4B & 4.8B & & & \\
\midrule
\multirow{6}{*}{\rotatebox[origin=c]{90}{Baseline}} & \multirow{3}{*}{${\tt SM + LN + G}$} & \multirow{3}{*}{25.71} & \multirow{3}{*}{23.32} & \multirow{3}{*}{21.29} & SM:$144\times\mathbb{R}^{512\times512}$ & \multirow{3}{*}{145.24} & \multirow{3}{*}{30.74} \\
& & & & & LN:$24\times\mathbb{R}^{512\times768}$ & & \\
& & & & & G:$12\times\mathbb{R}^{512\times3072}$ & & \\ \cline{2-8}
& \multirow{3}{*}{${\tt SM + LN + R}$} & \multirow{3}{*}{26.06} & \multirow{3}{*}{23.55} & \multirow{3}{*}{21.58} & SM:$144\times\mathbb{R}^{512\times512}$ & \multirow{3}{*}{81.71} & \multirow{3}{*}{23.54} \\
& & & & & LN:$24\times\mathbb{R}^{512\times768}$ & & \\
& & & & & R:$12\times\mathbb{R}^{512\times3072}$ & & \\
\midrule
\multirow{2}{*}{\rotatebox[origin=c]{90}{\scriptsize SOTA}} & \multirow{2}{*}{${\tt SM + ScFFN}$} & \multirow{2}{*}{NaNs} & \multirow{2}{*}{NaNs} & \multirow{2}{*}{NaNs} & SM:$144\times\mathbb{R}^{512\times512}$ & \multirow{2}{*}{72.10} & \multirow{2}{*}{21.56} \\
& & & & & LN: $1\times\mathbb{R}^{512\times768}$ & & \\ 
\midrule
\multirow{2}{*}{\rotatebox[origin=c]{90}{\scriptsize {\bf AERO}}} &${\tt SM + ScFFN}$ & 33.91 & 31.12 & 28.89 & SM:$144\times\mathbb{R}^{512\times512}$ & 71.76 & 21.51 \\	
&$\text{EReg}({\tt SM(t) + ScFFN})$ & \cellcolor{green!15}31.65 & \cellcolor{green!15}28.74 & \cellcolor{green!15}26.61 & \cellcolor{green!15}SM:$144\times\mathbb{R}^{512\times512}$ & \cellcolor{green!15}71.76 & \cellcolor{green!15}21.51 \\
\bottomrule
\end{tabular}}
\end{table}

\begin{table}[t]
\centering

\setlength{\tabcolsep}{4pt} 
\renewcommand{\arraystretch}{0.95} 

\caption{Comparison of AERO with SOTA \citep{he2024simplifying} across model depths and context sizes using Softmax-only variants, trained on  CodeParrot dataset. NaNs indicate training instability. Refer to Table \ref{tab:GPT2Scalability} for baseline performance and their PI overheads} 
\label{tab:GPT2Scalability}
\resizebox{0.49\textwidth}{!}{
\begin{tabular}{@{}p{0.2cm}lclcccc@{}}
\toprule
& \multirow{2}[4]{*}{Network Arch.} & \multirow{2}[4]{*}{PPL} & \multirow{2}[4]{*}{\#Nonlinear Ops} & \multirow{2}[4]{*}{\shortstack{Comm.\\(GB)}} & \multirow{2}[4]{*}{\shortstack{Lat.\\(min.)}} & \multicolumn{2}{c}{Savings} \\
\cmidrule(lr){7-8}
& & & & & & Comm. & Lat. \\  \midrule
\multicolumn{7}{c} { GPT-2 ($L$=12, $H$=12, $d$=768), $T$=128; Baseline PPL: GELU=2.69, ReLU=2.76} \\ \midrule
\multirow{2}{*}{\rotatebox[origin=c]{90}{\scriptsize SOTA}} & \multirow{2}{*}{${\tt SM + ScFFN}$} & \multirow{2}{*}{4.00} &  SM:$144\times\mathbb{R}^{128\times128}$ & \multirow{2}{*}{6.83} & \multirow{2}{*}{5.31} &  \multirow{2}{*}{3.71$\times$} & \multirow{2}{*}{1.55$\times$} \\
& & & LN: $1\times\mathbb{R}^{128\times768}$ & & & & \\ \cline{2-8}

\multirow{2}{*}{\rotatebox[origin=c]{90}{\scriptsize {\bf AERO}}} &${\tt SM + ScFFN}$ & 3.50 &  SM:$144\times\mathbb{R}^{128\times128}$ & 6.95 & 5.68  & 3.64$\times$ & 1.45$\times$ \\			
&$\text{EReg}({\tt SM(t) + ScFFN})$ & \cellcolor{green!15}3.24 & \cellcolor{green!15} SM:$144\times\mathbb{R}^{128\times128}$ & \cellcolor{green!15}6.43 & \cellcolor{green!15}4.76  & \cellcolor{green!15}3.94$\times$ & \cellcolor{green!15}1.72$\times$ \\
\midrule \midrule
\multicolumn{7}{c} { GPT-2 ($L$=12, $H$=12, $d$=768), $T$=256; Baseline PPL: GELU=2.56, ReLU=2.63} \\ \midrule
\multirow{2}{*}{\rotatebox[origin=c]{90}{\scriptsize SOTA}} & \multirow{2}{*}{${\tt SM + ScFFN}$} & \multirow{2}{*}{3.47} &  SM:$144\times\mathbb{R}^{256\times256}$ & \multirow{2}{*}{21.52} & \multirow{2}{*}{11.42} &  \multirow{2}{*}{2.72$\times$} & \multirow{2}{*}{1.45$\times$} \\
& & & LN: $1\times\mathbb{R}^{256\times768}$ & & & & \\ \cline{2-8}

\multirow{2}{*}{\rotatebox[origin=c]{90}{\scriptsize {\bf AERO}}} &${\tt SM + ScFFN}$ & 3.04 &  SM:$144\times\mathbb{R}^{256\times256}$ & 21.77 & 11.91  & 2.69$\times$ & 1.39$\times$ \\			
&$\text{EReg}({\tt SM(t) + ScFFN})$ & \cellcolor{green!15}2.96 & \cellcolor{green!15} SM:$144\times\mathbb{R}^{256\times256}$ & \cellcolor{green!15}20.72 & \cellcolor{green!15}10.45 & \cellcolor{green!15}2.82$\times$ & \cellcolor{green!15}1.59$\times$ \\
\midrule \midrule

\multicolumn{7}{c} { GPT-2 ($L$=18, $H$=12, $d$=768), $T$=128; Baseline PPL: GELU=2.35, ReLU=2.41} \\ \midrule
\multirow{2}{*}{\rotatebox[origin=c]{90}{\scriptsize SOTA}} & \multirow{2}{*}{${\tt SM + ScFFN}$} & \multirow{2}{*}{NaNs} &  SM:$216\times\mathbb{R}^{128\times128}$ & \multirow{2}{*}{9.39} & \multirow{2}{*}{6.75} &  \multirow{2}{*}{3.96$\times$} & \multirow{2}{*}{1.60$\times$} \\
& & & LN: $1\times\mathbb{R}^{128\times768}$ & & & & \\ 
\midrule
\multirow{2}{*}{\rotatebox[origin=c]{90}{\scriptsize {\bf AERO}}} &${\tt SM + ScFFN}$ & 3.26 &  SM:$216\times\mathbb{R}^{128\times128}$ & 9.62 & 7.23  & 3.86$\times$ & 1.49$\times$ \\			
&$\text{EReg}({\tt SM(t) + ScFFN})$ & \cellcolor{green!15}3.15 & \cellcolor{green!15} SM:$216\times\mathbb{R}^{128\times128}$ & \cellcolor{green!15}8.83 & \cellcolor{green!15}6.07 & \cellcolor{green!15}4.21$\times$ & \cellcolor{green!15}1.77$\times$ \\
\bottomrule
\end{tabular}}
\end{table}

{\bf Scalability across model depths and context lengths}
To demonstrate the scalability of AERO, particular the our loss recovery approach, entropy regularization, we employ AERO on Softmax-only GPT-2 variants across multiple depths and context lengths (Table \ref{tab:GPT2Scalability}). While the Softmax-only models show severe performance degradation compared to their baselines, entropy regularization consistently improves perplexity by {\bf 6\%}–{\bf 8\%}. However, SOTA often exhibits training instability under these conditions. This performance gap can be attributed to the critical role of MHA in model pre-training performance, particularly in the absence of FFN operations \citep{lu2024does}. We hypothesize that SOTA's aggressive optimization of attention FLOPs, unlike AERO's more balanced approach, leads to both inferior predictive performance and training instability.

\section{Conclusion} \label{Sec:Conclusion}
We perform an information-theoretic analysis for understanding the role of nonlinearities in pre-training learning dynamics, and developed 
entropy-guided framework (AERO) to strategically remove the costlier nonlinearities, achieving 3.41$\times$ communication reduction and 1.42$\times$ speedup without inuring performance loss. Although our study focuses on sub-billion parameter LLMs, the insights are likely to be relevant for larger models as well \cite{wortsman2024smallscale}.

{\bf Limitations.}
This work focuses on pre-training performance through perplexity as a primary metric, and does not evaluate transfer learning or few-shot learning capabilities. Future work will extend AERO to large-scale LLMs with broader experimental evaluation.(see Appendix \ref{Appendix:FutureWork}).



\bibliography{MyRef}
\bibliographystyle{mlsys2025}


\appendix

\clearpage
\section{Extended Related Work} \label{AppendixSec:extnd_related_work}

{\bf The Pitfalls of LayerNorm in LLM}
While LayerNorm serves as a critical source of nonlinearity in both convolutional neural networks \citep{ni2024on} and transformer-based models \citep{wu2024role,zhao2023tuning,joudaki2023impact}, it poses several challenges in the development and deployment of LLMs across various applications.
In private LLMs, LayerNorm poses challenges due to the difficulty of inverse-square root computations and the complexity of polynomial approximations caused by its unusually wide range of variance values \cite{zimerman2023converting}. 

Furthermore, LayerNorm’s trainable parameters are strongly linked to outlier features \citep{he2024understanding,bondarenko2024quantizable,wei2022outlier,kovaleva2021bert}, posing significant challenges for low-precision training and quantization. In particular, the LayerNorm scaling parameters amplify outlier activations, making it difficult to train models with fewer bits of precision \citep{wei2022outlier}.

LayerNorm also complicates mechanistic interpretability by making the residual stream more intricate and harder to analyze \citep{nanda2023attribution}. Finally, from a signal propagation perspective, LayerNorm can degrade trainability in deeper LLMs \citep{he2024simplifying,he2023deep}.

{\bf Entropy Regularization}
Entropy regularization has been widely applied in various areas of machine learning, serving multiple purposes: improving model robustness through penalizing low entropy predictions \citep{pereyra2017regularizing} or maximizing prediction entropy \citep{setlur2022maximizing}, improving adversarial robustness \citep{jagatap2022adversarially}, avoiding poor initialization and local minima \citep{miller1996global}, and optimizing layer-wise information flow \citep{peer2022improving}. 

In the context of attention mechanism, entropy regularization specifically targets attention score distributions. \cite{bao2024self} leverage a signal propagation framework to regulate attention weight localization by penalizing the QK-eigenspectrum scale, thereby preventing signal vanishing and ensuring stable training dynamics.

In reinforcement learning, entropy-regularization is used for balancing exploration-exploitation trade-offs and promoting action diversity \citep{wang2024diffusion,ahmed2019understanding,lu2019information,neu2017unified,mnih2016asynchronous}. Also, it is used in domain generalization \citep{zhao2020domain}.

\section{Integrations of Entropy Regularization in Loss Function} \label{Appendix:EntropyRegImplementation}

\subsection{Entropy Regularization Algorithm}
We present our entropy regularization in Algorithm \ref{Algo:EntRegLossComputation}, incorporating (i) head-wise learnable threshold for adaptable regularization,  and 2) tolerance margin to prevent excessive regularization, allowing a handful attention head to possess higher attention values.

\begin{algorithm*} [t]
\caption{Entropy Regularization Loss Computation} \label{Algo:EntRegLossComputation}

\textbf{Inputs:} $\text{attentions}$: List of attention matrices,  $\Theta(L,H)$= $\text{reg\_threshold\_weights}$, $T$: Sequence length, $\lambda$: Regularization loss weightage, $\gamma$: Hyper-parameter for  Tolerance margin

\textbf{Output:} $\mathcal{L}_{\text{total}}$: Total loss including entropy regularization

\begin{algorithmic}[1] 
\STATE $\mathcal{L}_{\text{entropy}} \leftarrow 0$
\STATE $\text{E}_{\text{max}} \leftarrow \log(T)$ \COMMENT{Theoretical maximum value of entropy}
\STATE $\text{Tol}_{\text{margin}} \leftarrow \gamma \text{E}_{\text{max}}$ \COMMENT{Tolerance margin is set as a small fraction of $\text{E}_{\text{max}}$} \label{line:ToleranceMargin}
\FOR{each layer $l$ in layers}
    \STATE $\mathcal{L}_{\text{layer}} \leftarrow 0$
    \STATE $\text{A}(t) \leftarrow \text{attentions}[l]$ \COMMENT{Attention matrix with learnable temperature for each query position}
    \STATE $\text{E}(t) \leftarrow -\frac{1}{T} \sum_{i=1}^{T} \sum_{j=1}^{T} \text{A}_{ij}(t) \log(\text{A}_{ij}(t))$ \COMMENT{Compute entropy, averaged over query length}
    \FOR{each head $h$ in heads}
        \STATE $E^{(l,h)} \leftarrow \text{Slice}(\text{E}(t), h)$ \COMMENT{Entropy for head $h$}
        \STATE $\theta^{(l, h)} \leftarrow \text{Slice}(\Theta{(L, H), h})$ \COMMENT{Learnable threshold weight head $h$}
        \STATE $\delta^{(l,h)} \leftarrow \text{E}^{(l,h)}(t) - \theta^{(l,h)} \text{E}_{\text{max}}$\COMMENT{Deviation from head-specific threshold} \label{line:LearnableRegThreshold}
        \STATE $\text{penalty}^{(l,h)} \leftarrow (\delta^{(l,h)})^2 \mathds{1}(|\delta^{(l,h)}| > \text{Tol}_{\text{margin}})$ \COMMENT{Penalize iff deviation exceeds Tolerance}
        \STATE $\mathcal{L}_{\text{layer}} \leftarrow \mathcal{L}_{\text{layer}} + \text{penalty}^{(l,h)}$
    \ENDFOR
    \STATE $\mathcal{L}_{\text{layer}} \leftarrow \frac{\mathcal{L}_{\text{layer}}}{\text{num\_heads}}$\COMMENT{Average over heads}
    \STATE $\mathcal{L}_{\text{entropy}} \leftarrow \mathcal{L}_{\text{entropy}} + \mathcal{L}_{\text{layer}}$
\ENDFOR
\STATE $\mathcal{L}_{\text{entropy}} \leftarrow \frac{\mathcal{L}_{\text{entropy}}}{\text{len(attentions)}}$ \COMMENT{Average over layers}
\STATE $\mathcal{L}_{\text{total}} \leftarrow \mathcal{L}_{\text{CE}} + \lambda \mathcal{L}_{\text{entropy}}$
\STATE \textbf{return} $\mathcal{L}_{\text{total}}$
\end{algorithmic}
\end{algorithm*}

\begin{figure*} [htbp]
\centering
\subfloat[SM + LN + G \label{subfig:sm_ln_g}]{\includegraphics[width=.33\textwidth]{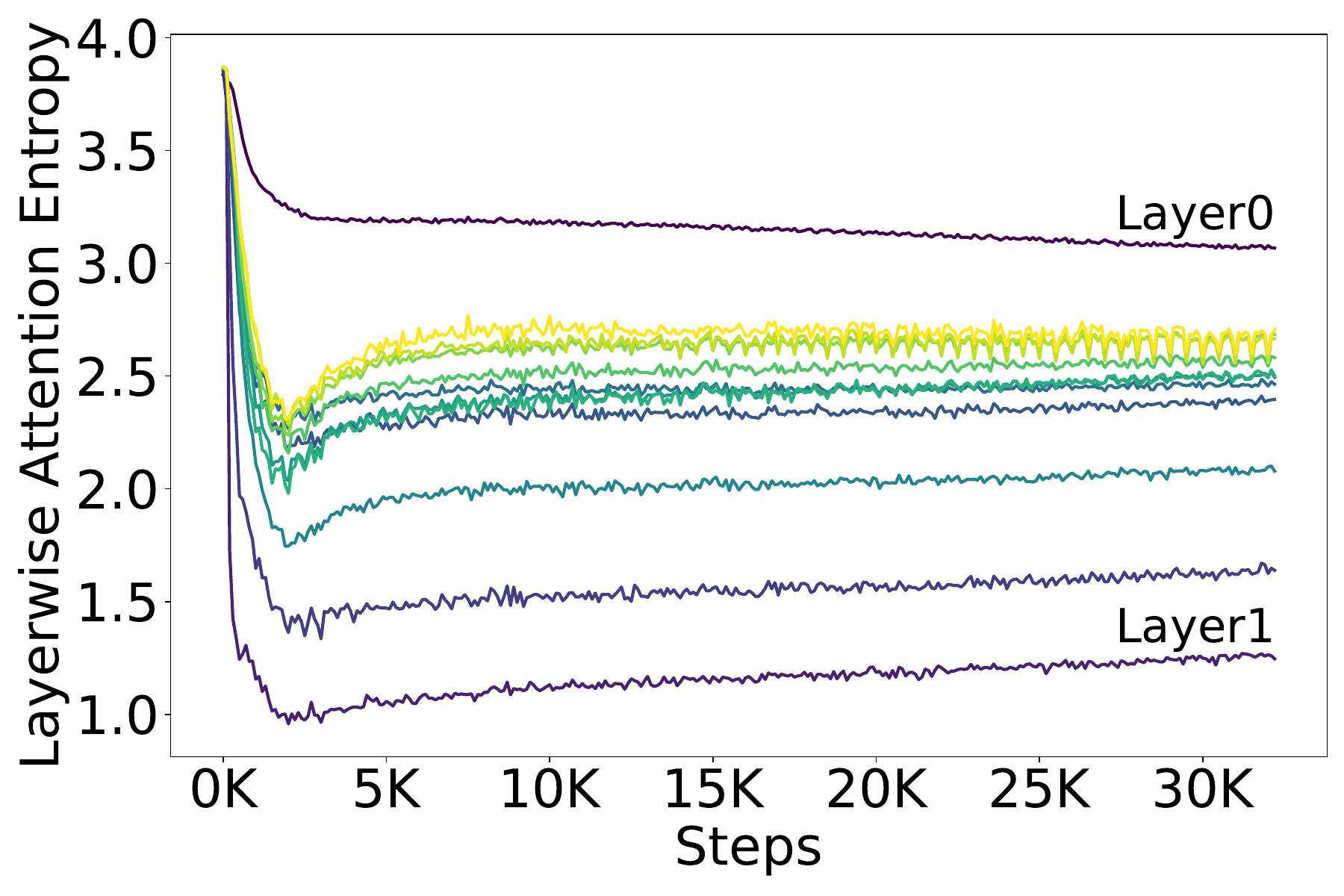}} 
\subfloat[SM + LN + R \label{subfig:sm_ln_r}]{\includegraphics[width=.33\textwidth]{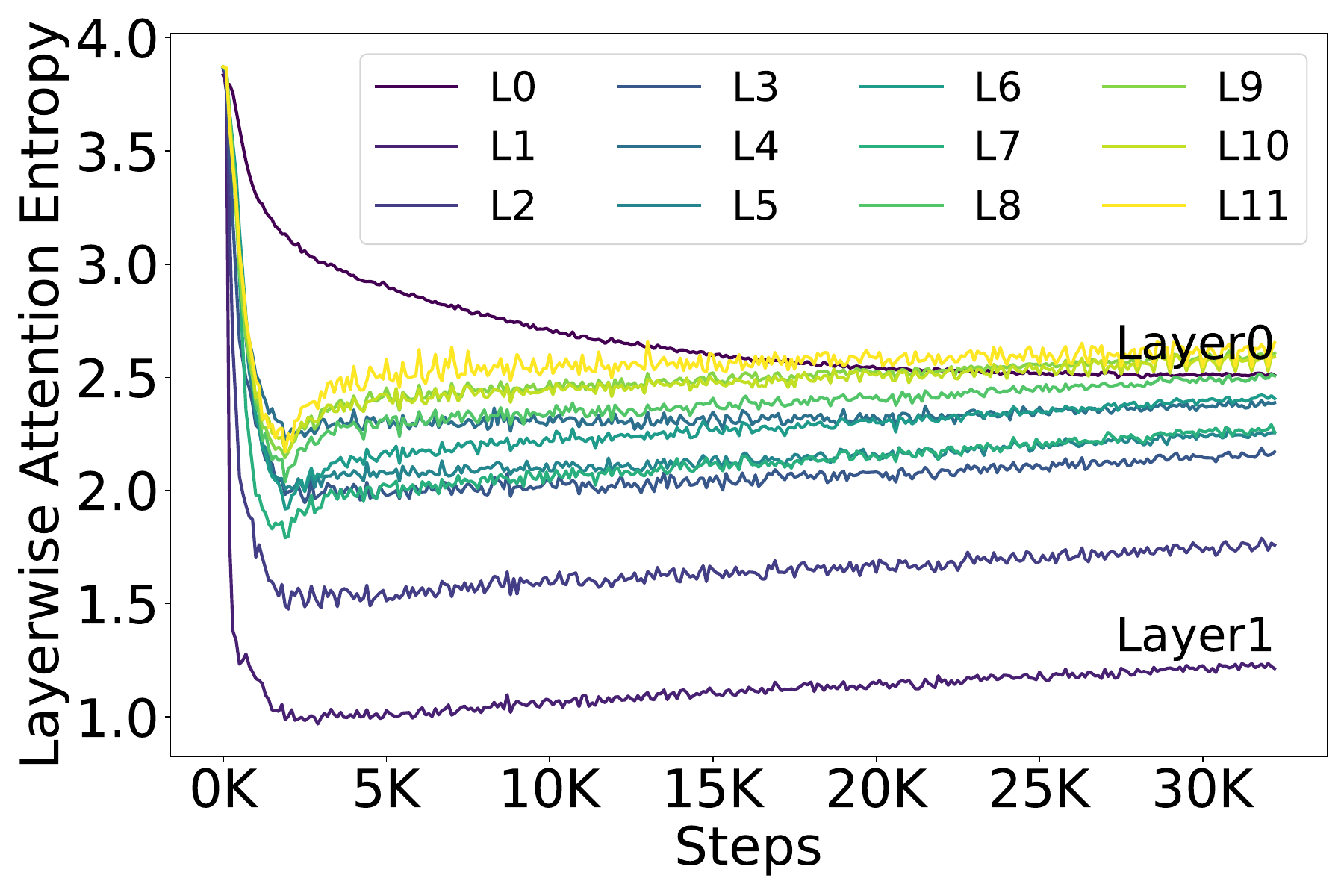}} 
\subfloat[SM + LN \label{subfig:sm_ln}]{\includegraphics[width=.33\textwidth]{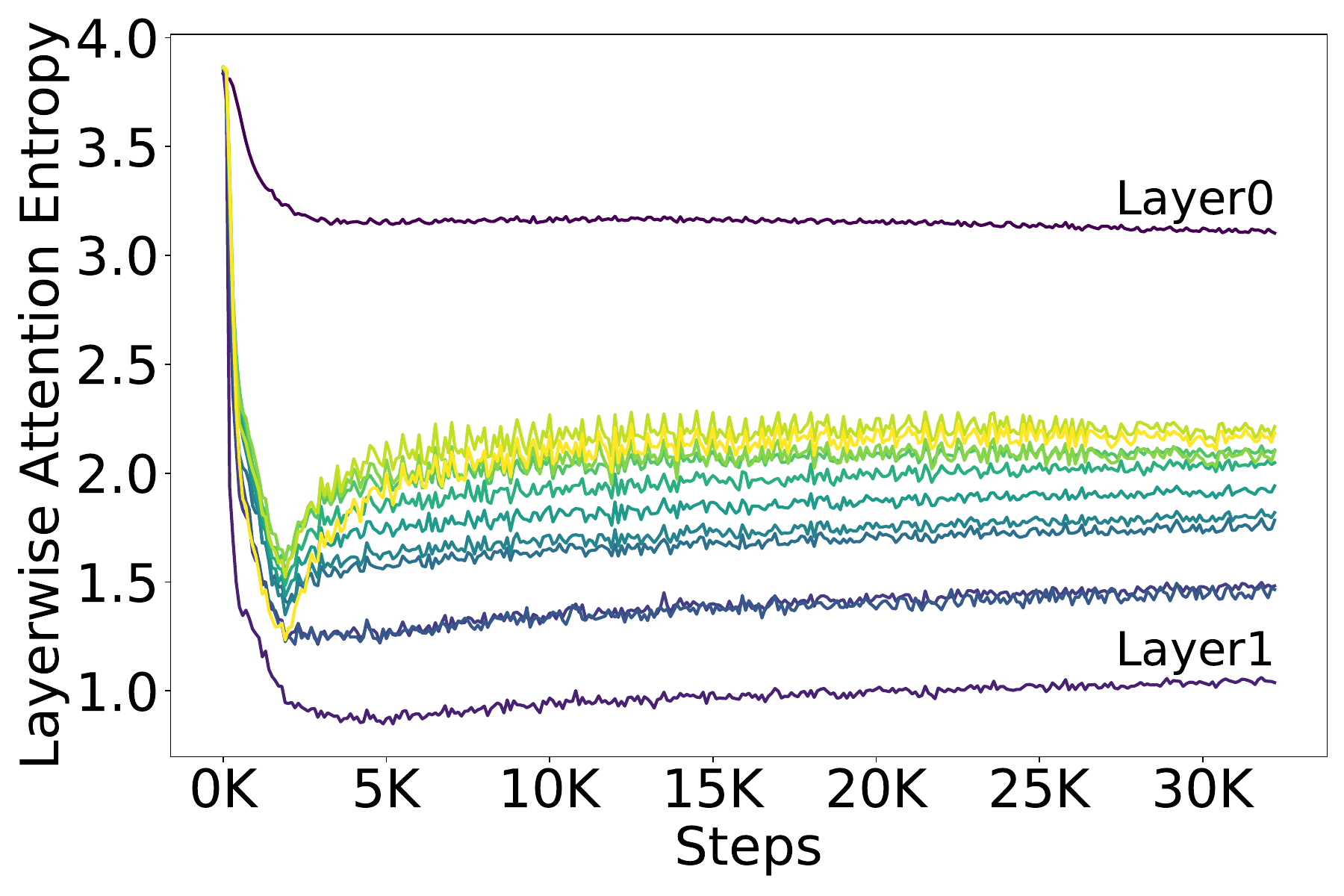}}  \\ \vspace{-1em}
\subfloat[SM + G \label{subfig:sm_g}]{\includegraphics[width=.33\textwidth]{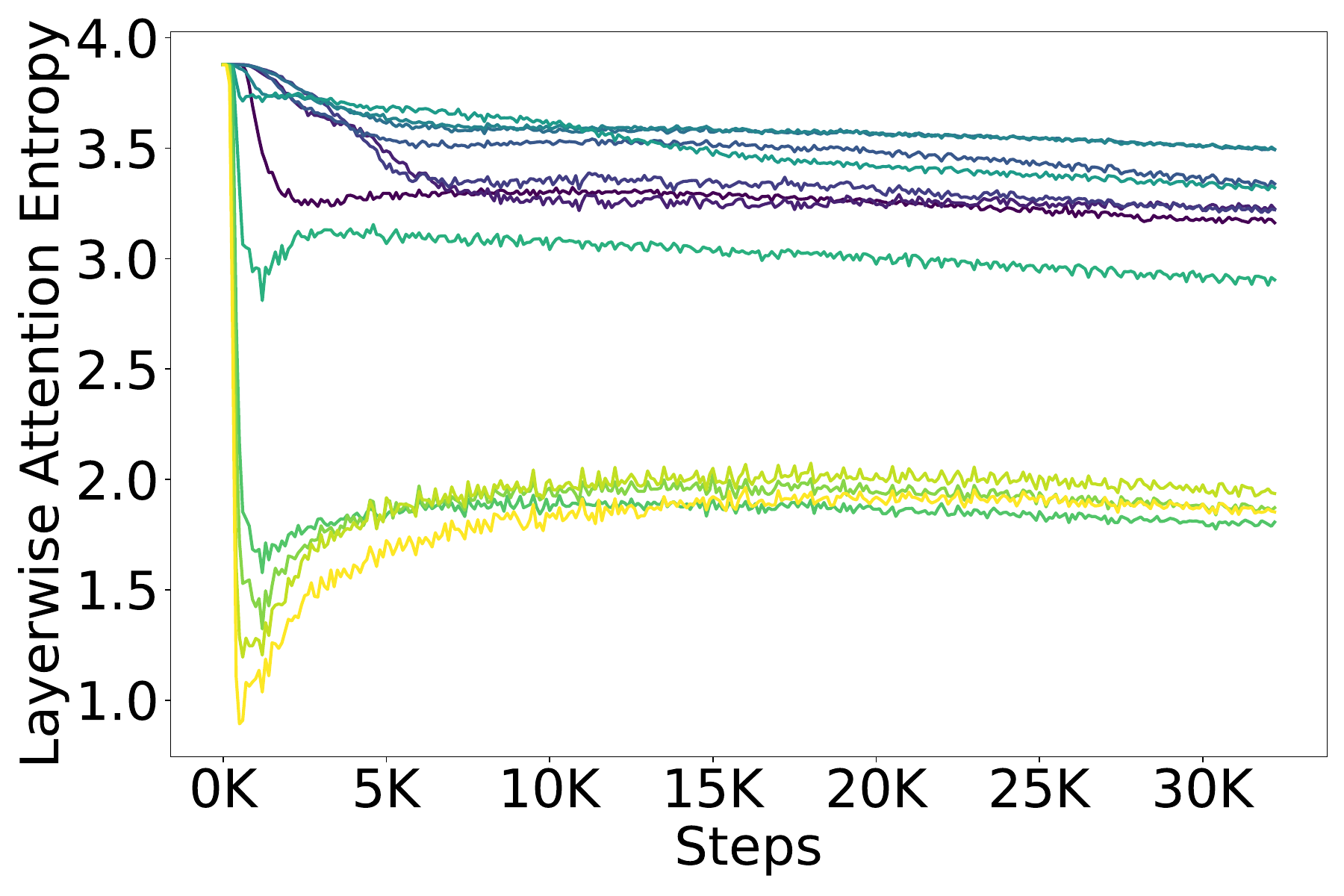}} 
\subfloat[SM + R \label{subfig:sm_r}]{\includegraphics[width=.33\textwidth]{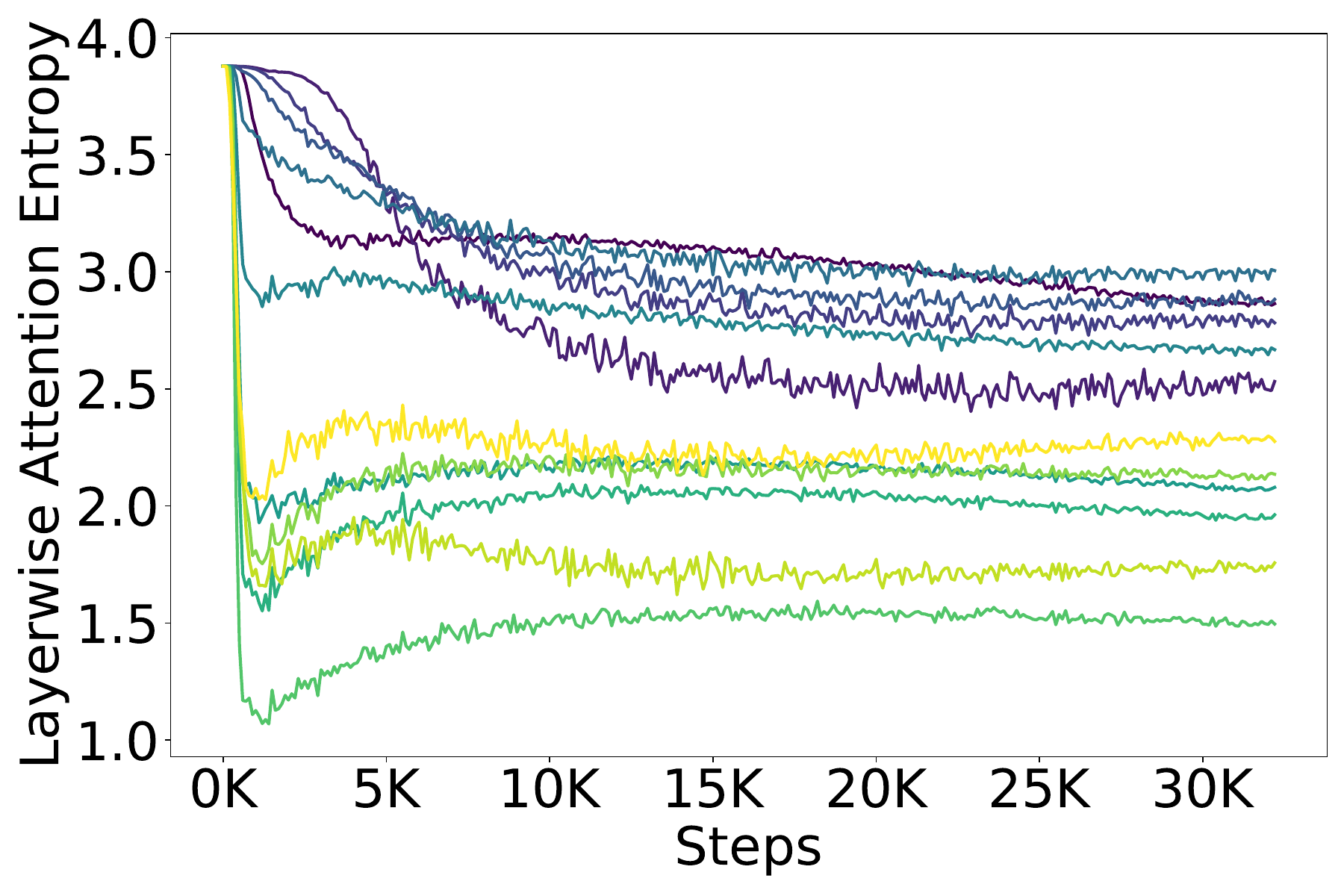}} 
\subfloat[SM \label{subfig:sm}]{\includegraphics[width=.33\textwidth]{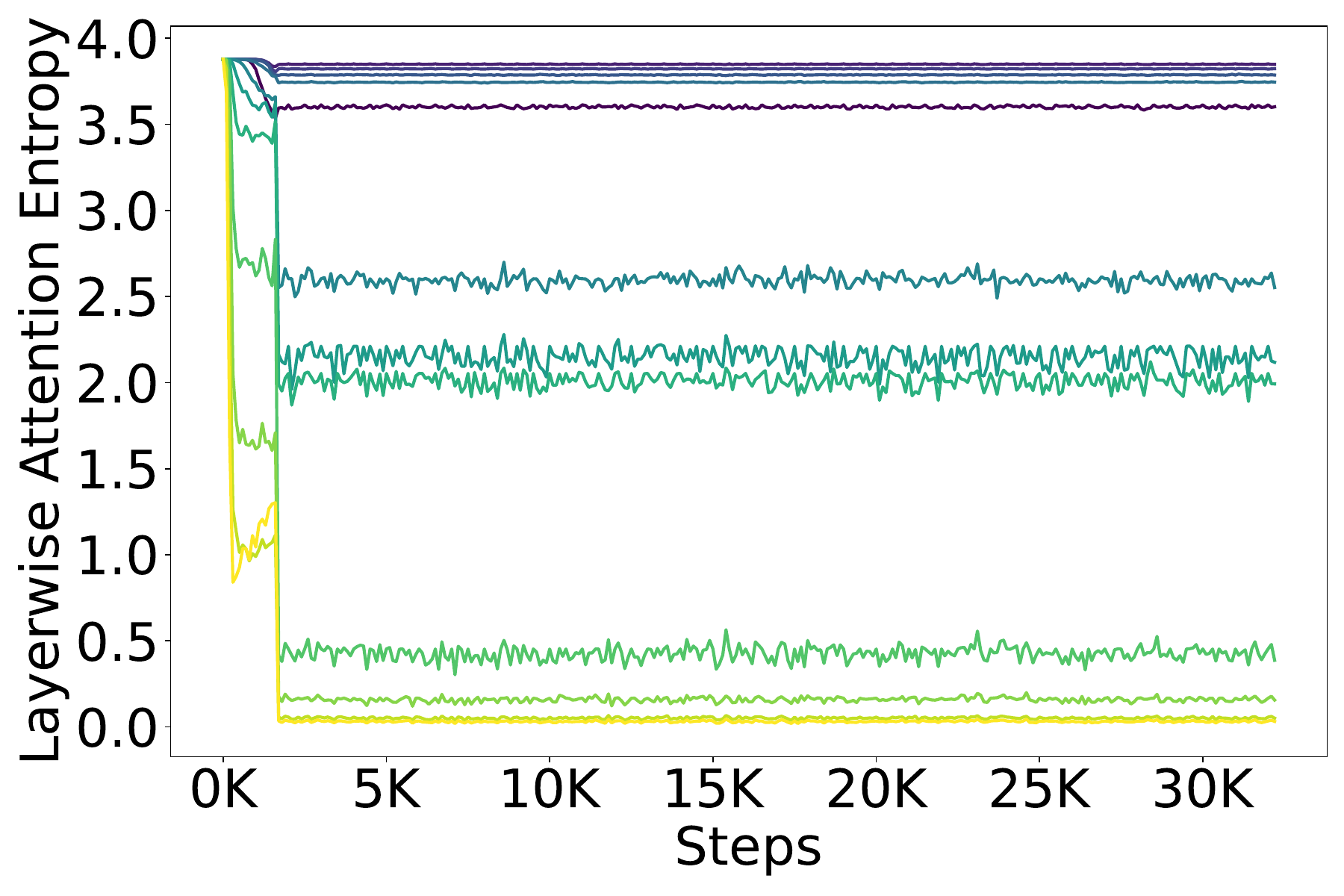}}  \vspace{-0.5em}
\caption{Entropy dynamics of GPT-2 (12$L$, 12$H$, 768$d$) models, each trained from scratch on 2.1B CodeParrot tokens with progressively fewer nonlinearities to isolate their effects. Normalization-free GELU (SM+G) and Softmax-only (SM) models show entropic overload in early layers; the latter also exhibits entropy collapse in deeper layers } 
\label{fig:LayerwiseEntropy}
\end{figure*}

\subsection{PyTorch Implementation} \label{subsecAppendix:EntRegPythonCode}
The PyTorch implementation, shown in Figure \ref{code:entropy-reg}, computes the entropy regularization loss for attention weights in a transformer model. This regularization ensures a balanced attention distribution, preventing it from becoming overly concentrated or too diffuse.

\newtcblisting{pythoncode}{
    arc=1pt,
    boxrule=0pt,
    colback=gray!5,
    listing only,
    left=20pt,
    enhanced,
    overlay={},
    listing options={
        basicstyle=\ttfamily\footnotesize,
        columns=fullflexible,
        breaklines=true,
        numbers=left,
        numbersep=5pt,
        numberblanklines=false,
        numberstyle=\tiny\color{gray},
        keywordstyle=\color{blue},
        commentstyle=\color{teal},
        stringstyle=\color{red},
        showstringspaces=false,
        frame=none,
        language=Python
    }
}

\begin{figure*}[t]  
\begin{pythoncode}
import torch

def calculate_entropy_reg_loss(attentions, blocks, seq_len):
    """
    Calculate the entropy regularization loss.
    
    Parameters:
    attentions (list): A list of attention matrices from different layers.
    blocks (list): A list of transformer blocks.
    seq_len (int): The length of the sequence (context length).
    
    Returns:
    float: The entropy regularization loss.
    """
    entropy_reg_loss = 0
    max_entropy = torch.log(torch.tensor(seq_len))  # Theoretical maximum entropy 
    fraction = 0.10  # Design hyper-parameter for tolerance margin
    tolerance_margin = fraction * max_entropy  # Set tolerance margin as fraction of the maximum entropy
    
    for layer_idx, (block, attn_mat) in enumerate(zip(blocks, attentions)):
        reg_threshold_weights = block.attn.reg_threshold_weights # Head-wise learnable parameters to set head-specific threshold
        ent_val = -torch.sum(attn_mat * torch.log(attn_mat + 1e-9), dim=-1)  # Compute entropy averaged over sequence length
        layer_entropy_reg_loss = 0
        
        for head_idx in range(block.attn.num_heads):
            head_entropy = ent_val[:, head_idx, :]  # Get head-specific entropy
            threshold = reg_threshold_weights[head_idx] * max_entropy
            deviation = torch.abs(head_entropy - threshold)
            penalty = torch.square(torch.where(deviation > tolerance_margin, deviation, torch.zeros_like(deviation)))
            layer_entropy_reg_loss += penalty.sum()
        
        layer_entropy_reg_loss /= block.attn.num_heads
        entropy_reg_loss += layer_entropy_reg_loss
    
    entropy_reg_loss /= len(attentions)
    return entropy_reg_loss
    
# Calculate the total loss including entropy regularization
lambda_reg = 1e-5  # Hyperparameter for entropy regularization weight
entropy_regularization = calculate_entropy_reg_loss(attentions, blocks, seq_len) 
total_loss = ce_loss + lambda_reg * entropy_regularization
\end{pythoncode}
\caption{PyTorch implementation of entropy regularization loss calculation.}
\label{code:entropy-reg}
\end{figure*}


\section{ Weight and Spectral Normalization} \label{Appendix:WNormSNorm}


Weight normalization reparameterizes the weight vectors as $\mathbf{W}_{\text{normalized}} = \frac{\mathbf{V}}{\|\mathbf{V}\|_2} g$, where $\mathbf{V}$ is reparameterized weight vector, $\|\mathbf{V}\|_2$ is Euclidean norm, and $g$ is a learnable scaling factor. Spectral normalization, in contrast,   normalizes the weight matrix $\mathbf{W}$ by its largest singular value $\sigma(\mathbf{W})$, yielding $\mathbf{W}_{\text{normalized}} = \frac{\mathbf{W}}{\sigma(\mathbf{W})}$. While weight normalization regulates weight magnitude, spectral normalization constrains the Lipschitz constant of linear layers.
We employed these normalizations in the FFN of the softmax-only model which transform $\text{FFN}^{\text{SM}}(\mathbf{X}) = (\mathbf{X} \mathbf{W}_{\text{in}}^{\text{ffn}})\mathbf{W}_{\text{out}}^{\text{ffn}}$ as follows:

\vspace{-2em}

\begin{equation}
\begin{aligned}
\text{FFN}_{\text{WNorm}}^{\text{SM}}(\mathbf{X}) &= \left(\mathbf{X} \frac{\mathbf{V}_{\text{in}}^{\text{ffn}}}{\|\mathbf{V}_{\text{in}}\|_2} g_{\text{in}}\right) \frac{\mathbf{V}_{\text{out}}^{\text{ffn}}}{\|\mathbf{V}_{\text{out}}\|_2} g_{\text{out}} \; \text{and} \\
\text{FFN}_{\text{SNorm}}^{\text{SM}}(\mathbf{X}) &= \left(\mathbf{X}  \frac{\mathbf{W}_{\text{in}}^{\text{ffn}}}{\sigma(\mathbf{W}_{\text{in}}^{\text{ffn}})}\right) \frac{\mathbf{W}_{\text{out}}^{\text{ffn}}}{\sigma(\mathbf{W}_{\text{out}}^{\text{ffn}})}
\end{aligned}
\end{equation}

\vspace{-1em}

Table \ref{tab:SNormVsWNorm} compares the performance of weight and spectral normalization applied to the linear layers within the attention and FFN sub-blocks in Softmax-only model. Results show that applying these normalization techniques to the attention blocks yields diminishing returns compared to their application in the FFN linear layers.

\begin{table}[htbp]
\caption{Comparison of weight normalization \cite{salimans2016weight} and spectral normalization \cite{miyato2018spectral} when employed in Softmax-only GPT-2 ($L$=12, $H$=12, $d$=768) models, and trained from scratch ($T$=128) on CodeParrot dataset. 
}
\label{tab:SNormVsWNorm}
\centering 
\begin{tabular}{lcc} \toprule 
Linear layers & Eval PPL(WNorm.) & Eval PPL(SNorm.) \\ \toprule 
QK & 3.89 & 4.25 \\
\rowcolor{green!20} FFN &  3.64 & 3.63 \\
QK+FFN& 3.88 & 4.23 \\
QKV+FFN& 3.93 & 4.26 \\
QKVO+FFN& 3.98 & 4.34 \\ \bottomrule  
\end{tabular} 
\end{table}


\section{Entropy Dynamics}
Figure \ref{fig:LayerwiseEntropy} presents the entropy dynamics of the GPT-2 ($L$=12, $H$=12, $d$=3072) models as nonlinearities are progressively
removed. Each model is  trained from scratch on 2.1B tokens from CodeParrot datatset.


\clearpage

\section{Learnable Scaling in Feed-Forward Networks} \label{Appendix:LearnableScalingFFN}
\begin{figure} [!htbp]
\centering
\subfloat[FFN scaling factor $\alpha$ \label{subfig:ffn_alpha}]{\includegraphics[width=.245\textwidth]{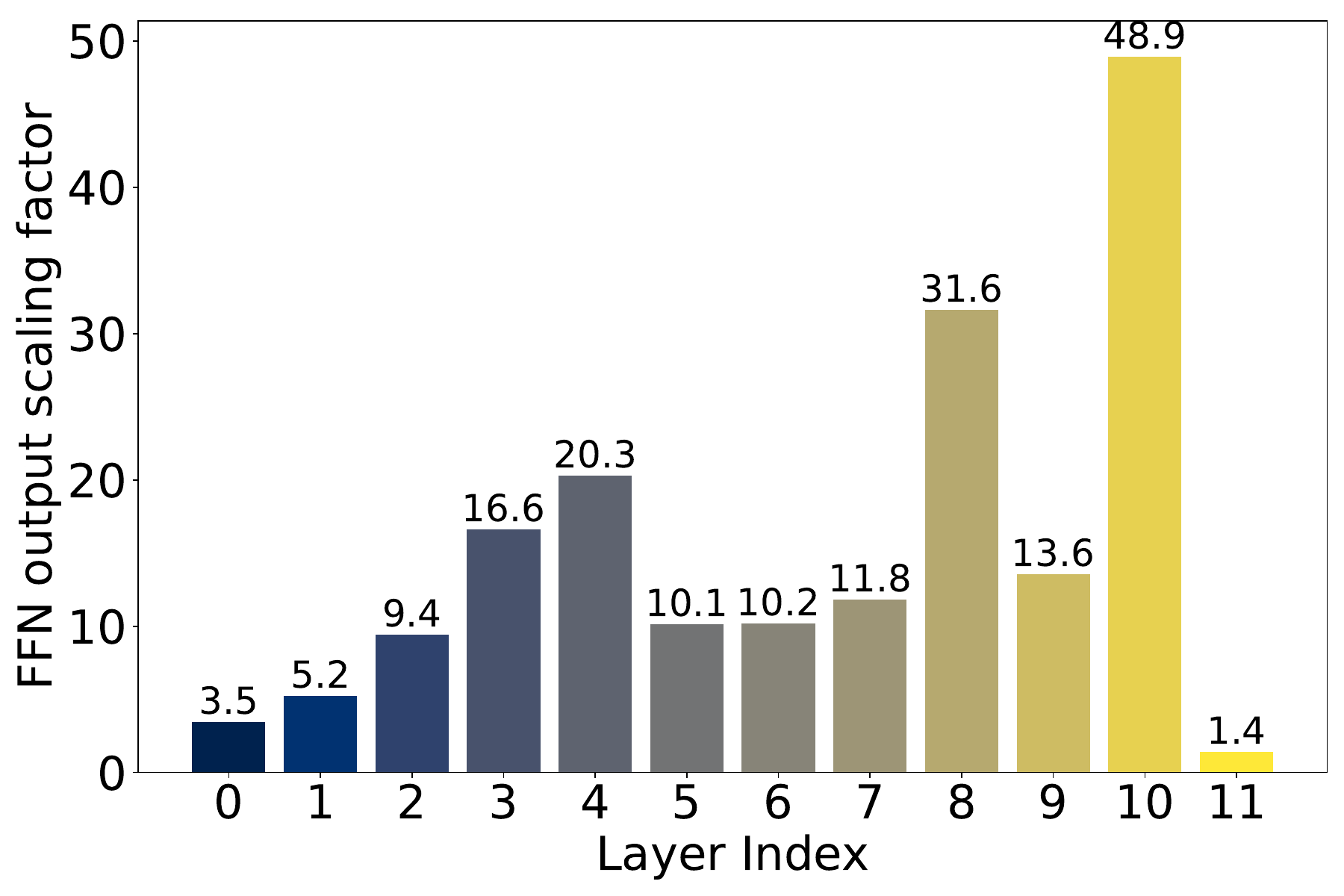}} 
\subfloat[FFN residual scaling factor $\beta$ \label{subfig:ffn_beta}]{\includegraphics[width=.245\textwidth]{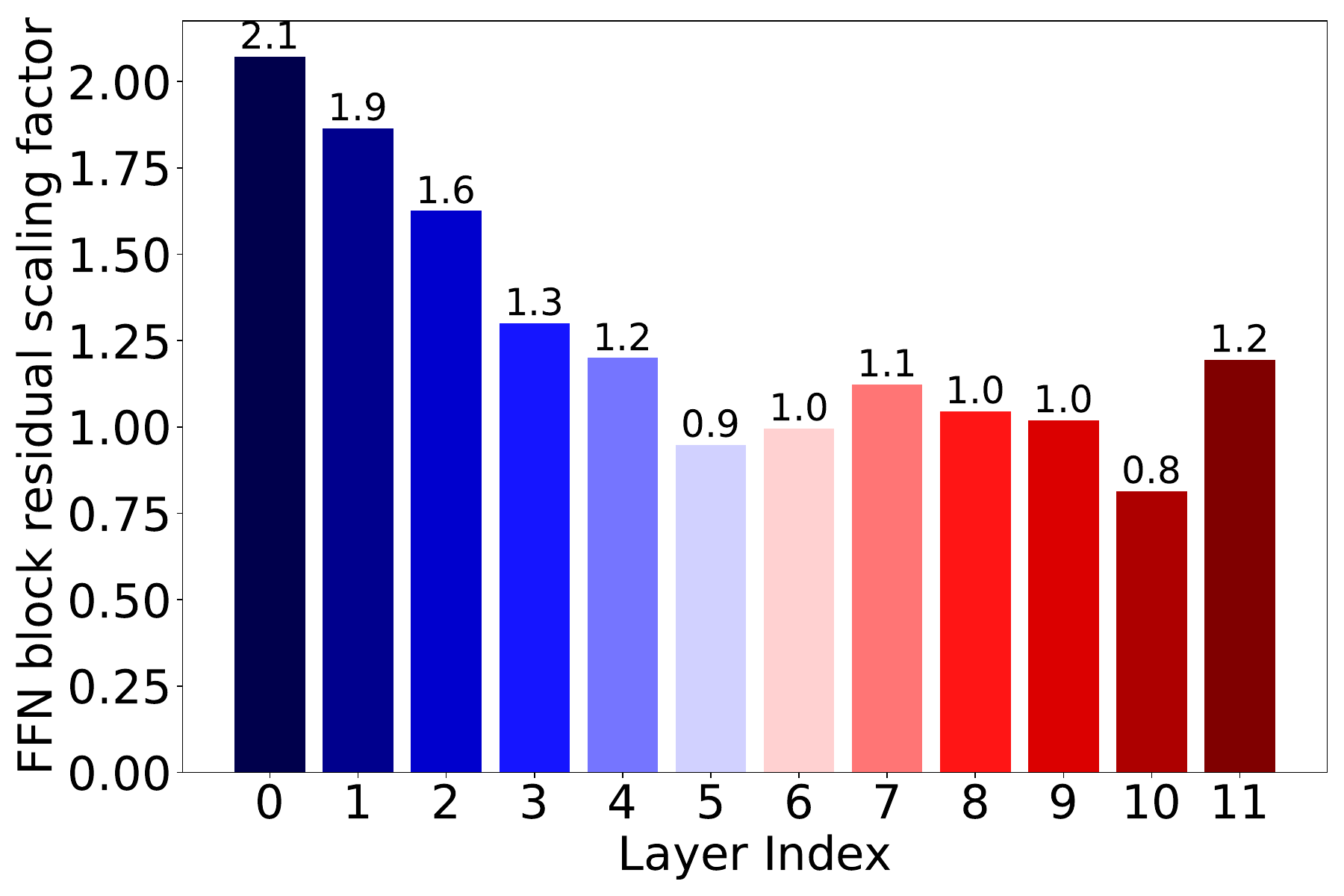}} \vspace{-1em}
\caption{Learned scaling factors $\alpha$ and $\beta$ in Eq. \ref{eqn:ScaledFFN} across different layers in the Softmax-only GPT-2 model ($L$=12, $H$=12, $d$=768). The values were plotted after full model training to observe the modulation of FFN outputs and residual connections in each layer.} 
\label{fig:LearnableScalingFactor}
\end{figure}

We plot the values of FFN scaling factors $\alpha$ and $\beta$ (see Eq. \ref{eqn:ScaledFFN}) learned across the layers in the full-trained softmax-only GPT-2 model, and made the following observations from the Figure \ref{fig:LearnableScalingFactor}:

\begin{itemize} [noitemsep,nolistsep,leftmargin=0.5cm] \vspace{-18em}
\item {\bf Significant increase in $\alpha$ with layer depth:} The scaling factor $\alpha$ increases substantially in the deeper layers of the model, with particularly high values observed in $L$10 (Figure \ref{subfig:ffn_alpha}). This indicates that as the network goes deeper, the FFN outputs are heavily scaled down by $\alpha$. This downscaling is essential to prevent the FFN outputs from dominating the activations, which could otherwise lead to numerical instability, as evidenced by the NaNs observed early in training in Figure \ref{subfig:LayerwiseNaNs}. The large $\alpha$ values in deeper layers suggest that this downscaling becomes increasingly critical as the model progresses through its layers, effectively stabilizing the training process by keeping the FFN outputs in check.
\item {\bf Balancing $\beta$ values:} The  scaling factors $\beta$, which modulate the residual connections within the FFN block by upscaling their output, start higher in the earlier layers and decreases gradually, with some fluctuation (Figure \ref{subfig:ffn_beta}). The moderate up-scaling provided by $\beta$ ensures that the residual connections are not overshadowed by the scaled-down FFN outputs. This balance between the strong downscaling by $\alpha$ and the corresponding upscaling by $\beta$ is crucial for stabilizing activations across layers, particularly in deeper layers where instability is more likely to occur. 
\end{itemize}


These observations underscore the critical role that the learnable scaling factors $\alpha$ and $\beta$ play in stabilizing the training of softmax-only GPT-2 models. By dynamically adjusting the contributions of the FFN sub-block outputs, $\alpha$ and $\beta$ prevent the numerical issues that arise in deeper layers, ensuring stable and effective training. This fine-tuned balance is key to avoiding entropy collapse and other forms of instability that would otherwise derail the training process.

\section{Mitigating Over-Regularization Through Tolerance Margin}


\begin{figure} [htbp]
\centering
\includegraphics[width=0.49\textwidth]{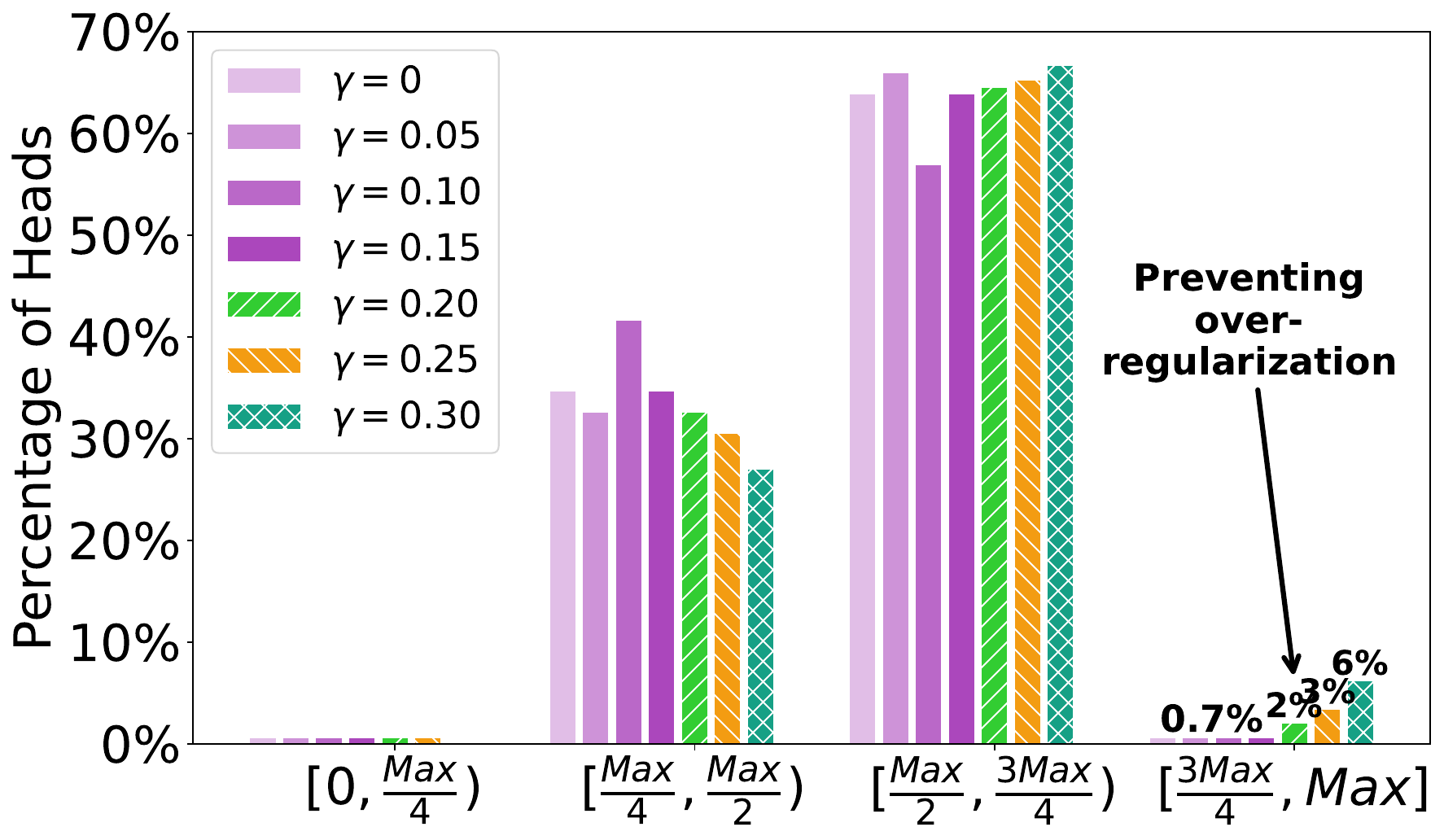} \vspace{-2em}
\caption{Headwise entropy distribution in Softma-only GPT-2 model (12$L$, 12$H$, 768$d$) when entropy regularization is applied with varying threshold margin, controlled by hyperparameter $\gamma$.  } 
\label{fig:EntDistTmargin}
\end{figure}


\begin{figure*} [t]
\centering
\subfloat[\(\text{Tol}_{\text{margin}} = 0 \)]{\includegraphics[width=.33\textwidth]{plots/ent_layerwise_Tmargin_0Per}} 
\subfloat[\(\text{Tol}_{\text{margin}} = 0.05\text{E}_{\text{max}} \)]{\includegraphics[width=.33\textwidth]{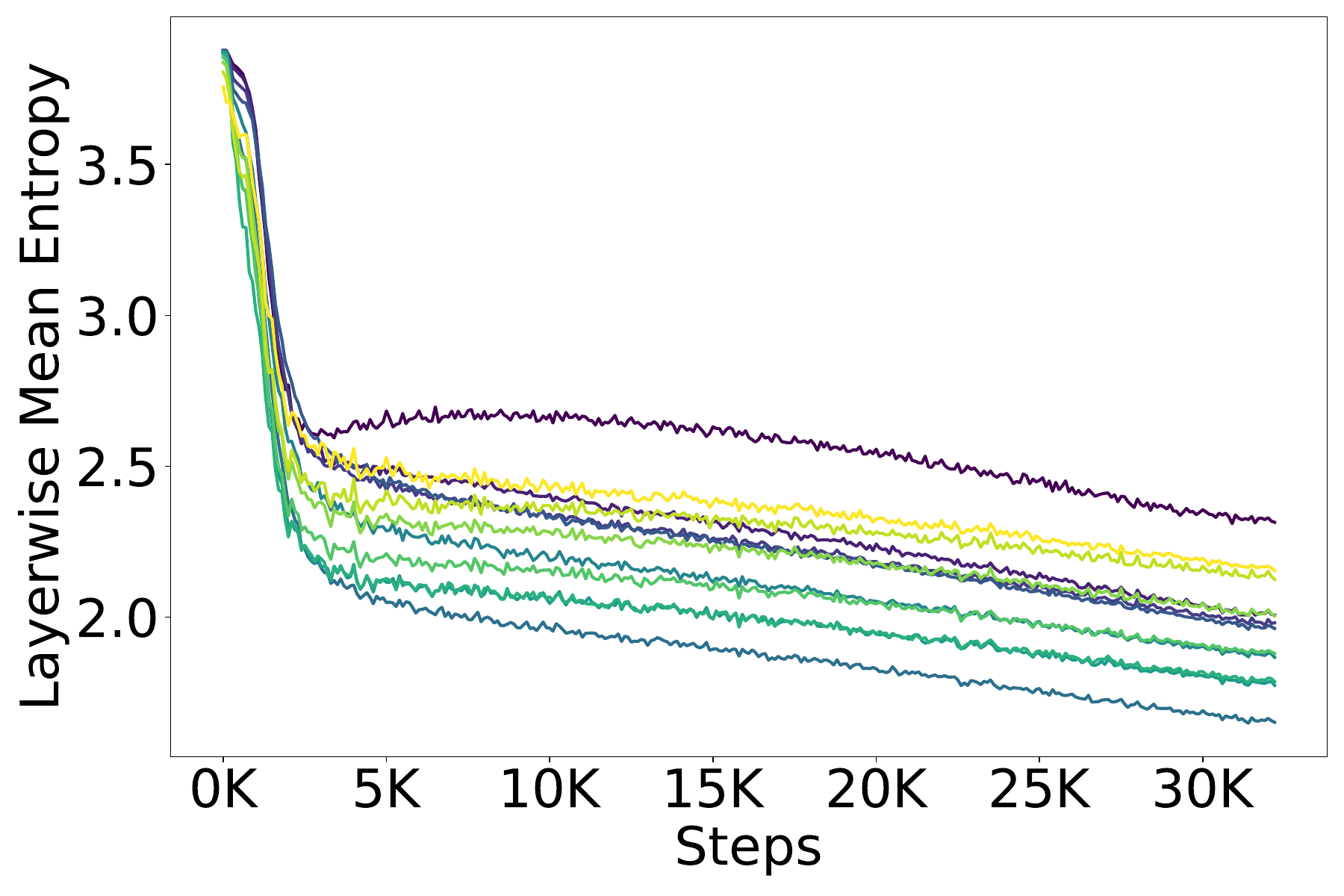}} 
\subfloat[\(\text{Tol}_{\text{margin}} = 0.10\text{E}_{\text{max}} \)]{\includegraphics[width=.33\textwidth]{plots/ent_layerwise_Tmargin_10Per}} \\ \vspace{-1em}  
\subfloat[\(\text{Tol}_{\text{margin}} = 0.15\text{E}_{\text{max}} \)]{\includegraphics[width=.33\textwidth]{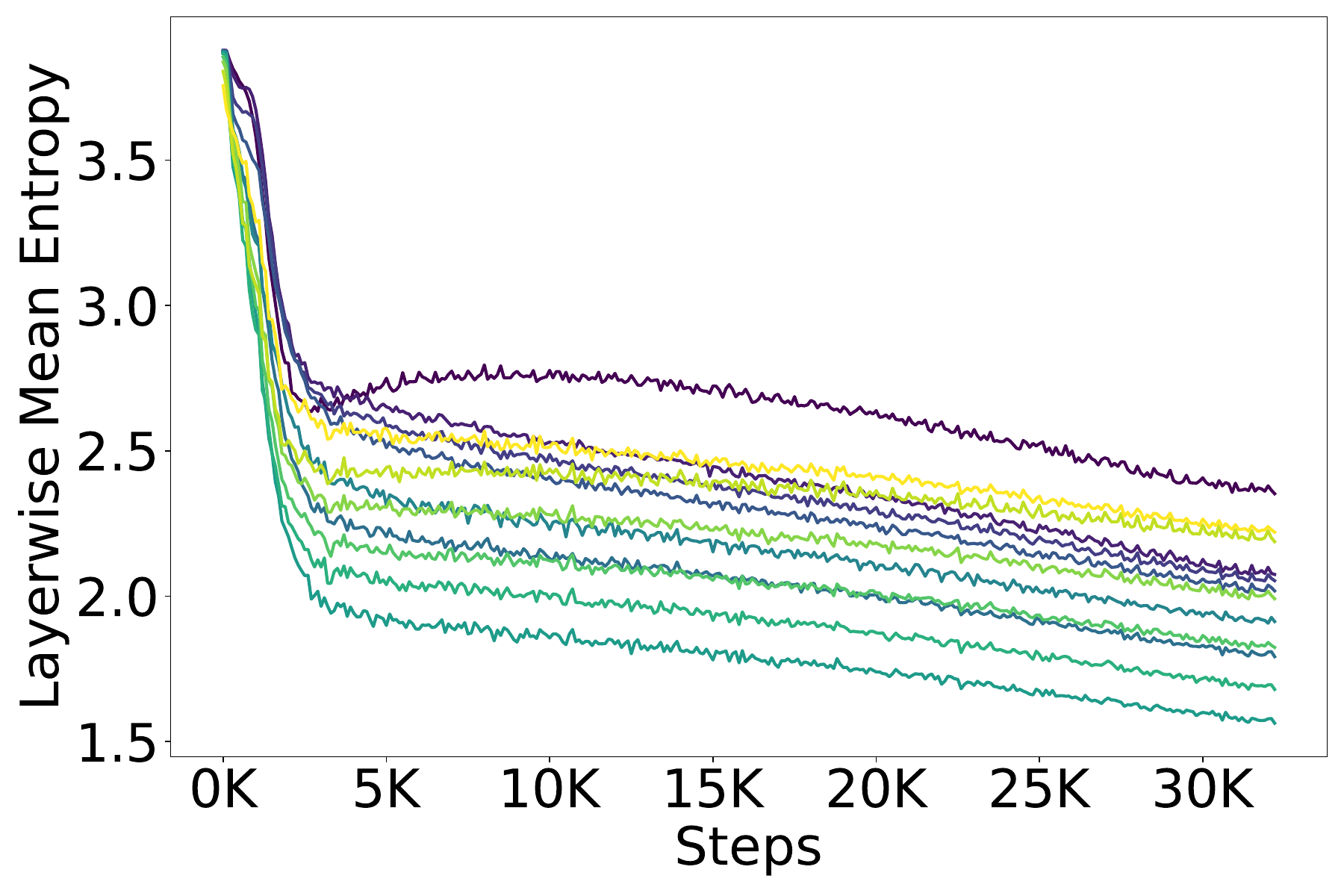}}  
\subfloat[\(\text{Tol}_{\text{margin}} = 0.20\text{E}_{\text{max}} \)]{\includegraphics[width=.33\textwidth]{plots/ent_layerwise_Tmargin_20Per}}  
\subfloat[\(\text{Tol}_{\text{margin}} = 0.25\text{E}_{\text{max}} \)]{\includegraphics[width=.33\textwidth]{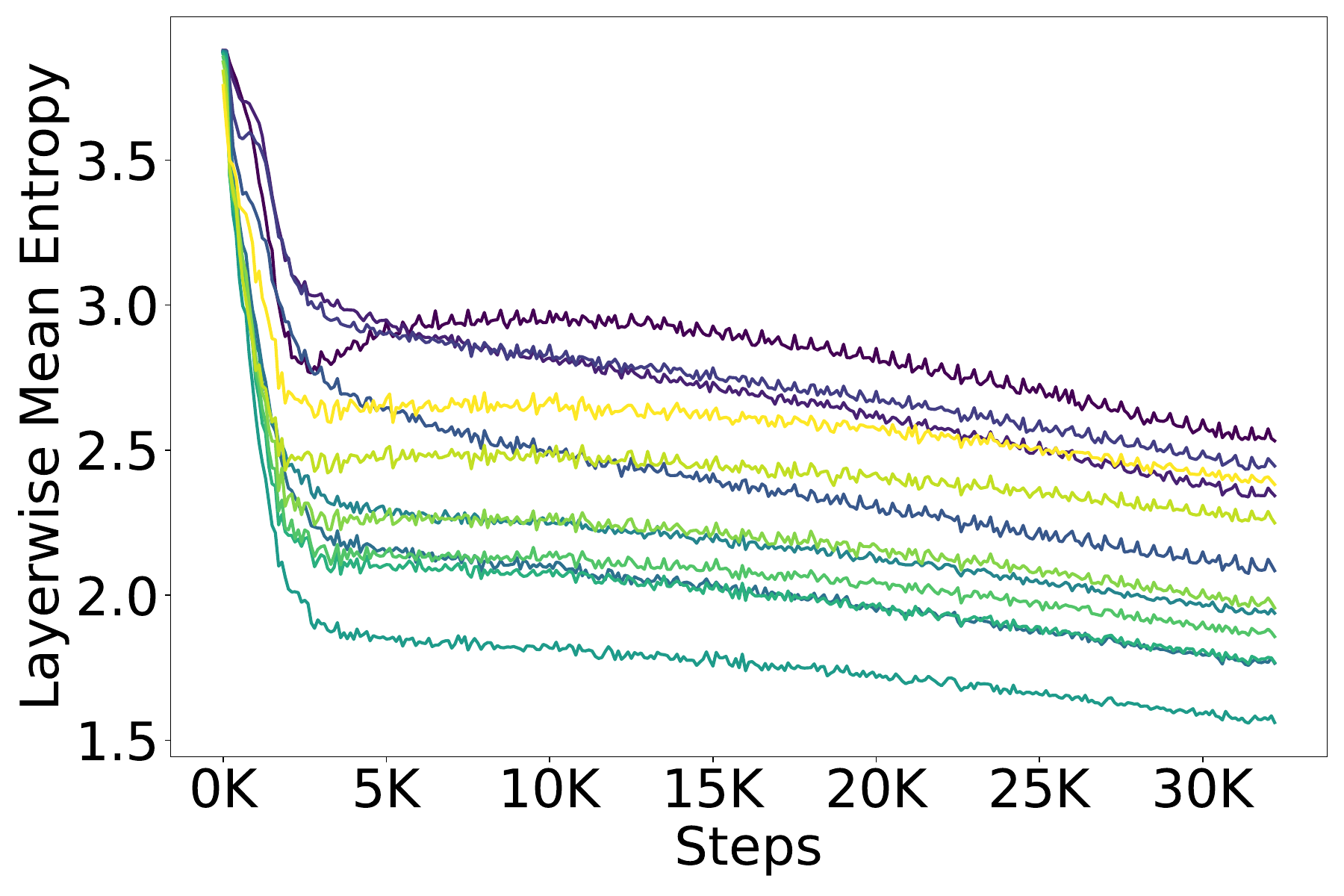}}  
\caption{Layerwise entropy dynamics when entropy regularization is employed with increasing threshold margin, defined as \(\text{Tol}_{\text{margin}} = \gamma \text{E}_{\text{max}} \) (see Algorithm\ref{Algo:EntRegLossComputation}, line \#\ref{line:ToleranceMargin}). At higher $\gamma$, the mean entropy of the early layers increases.}
\label{fig:EntropyDynamicsTmargins}
\end{figure*}

Figure \ref{fig:EntDistTmargin} illustrates the effect of $\gamma$ on the headwise entropy distribution.  The hyperparameter $\gamma$  employed to adjust the threshold margin in entropy regularization, defined as \(\text{Tol}_{\text{margin}} = \gamma \text{E}_{\text{max}} \) (Algorithm\ref{Algo:EntRegLossComputation}, line \#\ref{line:ToleranceMargin}), effectively preventing over-regularization by ensuring that a sufficient fraction of heads maintains entropy values in the upper range  $\frac{\tt 3Max}{4}$ to ${\tt Max}$. As $\gamma$ increases from 0 to 0.15, only a small proportion of attention heads (0.7\%) are situated in the highest entropy range. However, as $\gamma$ is increased beyond 0.15, the fraction of heads in this upper range starts increasing, reaching 2.08\%, 3.47\%, and 6.25\% at $\gamma$=0.20, 0.25, and 0.30, respectively. This fine-grained control on the population of attention heads in the higher entropy range highlights the ability of entropy regularization to prevent over-regularization and maintain the attention heads' diversity. We find that $\gamma$=0.2  yields slightly better performance in terms of lower perplexity compared to higher $\gamma$ values, and thus, we adopt this value in the final entropy regularization scheme.

Furthermore, Figure \ref{fig:EntropyDynamicsTmargins} illustrates the layer-wise entropy dynamics during training as the tolerance margin is progressively increased. Evidently, at higher $\gamma$, a greater fraction of attention heads exhibit higher entropy values, as reflected in the increased mean entropy of the early layers. This prevents over-regularization while preserving the attention head diversity.


\section{Additional Results} 

\subsection{Baseline GPT-2 Models} \label{AppendixSec:BaselinePerf}

Table \ref{tab:AddtionalBaselines} shows the detailed breakdown of private inference (communication and latency) overheads of baseline GPT-2 models
We compare two architectural variants for each baseline: the traditional GELU-based architecture (${\tt SM + LN + G}$) and the PI-friendly ReLU  variant (${\tt SM + LN + R}$). 

\begin{table}[htbp]
\centering
\small
\setlength{\tabcolsep}{4pt}
\renewcommand{\arraystretch}{0.95}
\caption{Performance evaluation of baseline GPT-2 models trained from scratch on the CodeParrot dataset (2.1B tokens). $L$, $H$, $d$, and $T$ represent model depth, head counts, embedding dimension, and input context length during training, respectively.} 
\label{tab:AddtionalBaselines}
\resizebox{0.49\textwidth}{!}{
\begin{tabular}{@{}l>{\centering\arraybackslash}c>{\centering\arraybackslash}cl>{\centering\arraybackslash}c>{\centering\arraybackslash}c@{}}
\toprule
Config. & Network Arch. & Eval PPL & \#Nonlinear Ops & \shortstack{Comm. \\(GBs)} & \shortstack{Latency \\(minutes)} \\
\midrule

\multirow{6}{*}{\shortstack[l]{$L$=18\\$H$=12\\$d$=768\\$T$=128}} 
& \multirow{3}{*}{${\tt SM + LN + G}$} & \multirow{3}{*}{2.56} & SM:$216\times\mathbb{R}^{128\times128}$ & \multirow{3}{*}{37.17} & \multirow{3}{*}{10.77} \\
& & & LN:$36\times\mathbb{R}^{128\times768}$ & & \\
& & & G:$18\times\mathbb{R}^{128\times3072}$ & & \\ \cline{2-6}
& \multirow{3}{*}{${\tt SM + LN + R}$} & \multirow{3}{*}{2.63} & SM:$216\times\mathbb{R}^{128\times128}$ & \multirow{3}{*}{13.34} & \multirow{3}{*}{8.04} \\
& & & LN:$36\times\mathbb{R}^{128\times768}$ & & \\
& & & R:$18\times\mathbb{R}^{128\times3072}$ & & \\ \midrule

\multirow{6}{*}{\shortstack[l]{$L$=12\\$H$=12\\$d$=768\\$T$=256}} 
& \multirow{3}{*}{${\tt SM + LN + G}$} & \multirow{3}{*}{2.35} & SM:$144\times\mathbb{R}^{256\times256}$ & \multirow{3}{*}{58.51} & \multirow{3}{*}{16.57} \\
& & & LN:$24\times\mathbb{R}^{256\times768}$ & & \\
& & & G:$12\times\mathbb{R}^{256\times3072}$ & & \\ \cline{2-6}
& \multirow{3}{*}{${\tt SM + LN + R}$} & \multirow{3}{*}{2.41} & SM:$144\times\mathbb{R}^{256\times256}$ & \multirow{3}{*}{26.73} & \multirow{3}{*}{12.59} \\
& & & LN:$24\times\mathbb{R}^{256\times768}$ & & \\
& & & R:$12\times\mathbb{R}^{256\times3072}$ & & \\
\bottomrule
\end{tabular}}
\end{table}

The ReLU-based design significantly improve PI efficiency, with a  marginal  increase in perplexity (PPL). Notably, communication overhead is reduced by up to 2.79$\times$ (from 37.17GB to 13.34GB in the 18-layer model), while inference latency decreases by $\sim$1.3$\times$ across all configurations.

\section{Future Work} \label{Appendix:FutureWork}

To further reduce non-linear operations, off-the-shelf head pruning techniques \cite{voita2019analyzing,michel2019sixteen,jo2020roles,ma2021contributions,li2022accelerating} can be applied on top of AERO. 
Another approach is to explore linear softmax operations. However, these linear softmax operations sometimes introduce additional normalization layers or complex activation functions in the FFN \cite{zhang2024the}, which could increase the PI overheads, counteracting the intended efficiency improvements.

Additionally, incorporating weight and activation quantization \cite{wu2023understanding,xiao2023smoothquant,dettmers2023case} could further enhance the efficiency of private inference in our architecture.

Orthogonally, performance improvement techniques such as knowledge distillation (KD) can be employed to complement these optimizations \cite{ko2024distillm,liang2023less,gu2023minillm,hsieh2023distilling,li2023symbolic}.

Looking ahead, scaling AERO to more complex and deeper LLMs can be achieved by strategically combining techniques such as weight normalization, spectral normalization, and FFN output scaling. These methods can be applied selectively, with different layers using different techniques---for instance, employing spectral normalization in early layers and FFN output scaling in deeper layers. This tailored approach could lead to better stability and efficiency in larger models.


\section{AERO Beyond Private Inference: Broader Impact and Promises} \label{Appendix:BroaderImpact}
While AERO is developed for enabling efficient private inference in transformer-based language models, its architectural innovations and insights have far-reaching implications beyond privacy-preserving computations. Its principled approach for architectural simplifications and techniques for maintaining model stability offer valuable insights into the broader field of language model design and optimization. Here, we discuss how AERO's principles and entropy-guided solutions can impact both architectural design choices and practical deployment considerations.

{\bf Plaintext efficiency}
Our finding that ReLU naturally emerges as the preferred activation function in normalization-free architectures aligns well with plaintext efficiency goals. ReLU's ability to induce sparsity in activations accelerates the plaintext inference by reducing the data traffic between CPU and GPU \cite{mirzadeh2024relu}.

{\bf Low-precision training and quantization for resource-constrained applications}
The deployment of LLMs in resource-constrained environments is significantly hindered by outliers that complicate low-bitwidth quantization. These outliers emerge from two primary sources: softmax-based attention mechanism \cite{hu2024outlierefficient} and LayerNorm layers \cite{he2024understanding,wei2022outlier,kovaleva2021bert}. LayerNorm-induced outliers are particularly {\em severe} as their scaling parameters amplify activation extremes, demanding higher bit-widths for accurate representation.

Our normalization-free design inherently addresses the major source of outliers by eliminating problematic scaling parameters. While empirical validation remains necessary, AERO’s architectural choices offer promising directions for enabling low-precision training and quantization, where effective outlier management is crucial.

{\bf Entropy-guided framework for tackling softmax-inherent challenges in attention mechanism}
The softmax function, fundamental to transformer-based attention mechanisms, inherently assigns non-zero probabilities to all tokens due to its normalized exponential structure. This leads to two primary issues inherent to softmax: disproportionate emphasis on specific tokens (known as attention sink) \cite{xiao2024efficient,cancedda2024spectral,gu2024attention}; and non-zero scores for irrelevant tokens (known as attention noise). These challenges can result in undesirable effects such  as  hallucinations \cite{ye2024differential}, outlier activations \cite{hu2024outlierefficient}, and  rank collapse \cite{bao2024self}. 

While prior work proposed various strategies to mitigate these issues \cite{yin2024stablemask,yu2024unveiling,bao2024self}, our entropy regularization offers a fine-grained control of attention entropy distribution. By penalizing extreme entropy entropy values using headwise learnable threshold parameters, each attention head adaptively determine its optimal degree of focus, preventing over-diffusion while {\em preserving} the mathematical properties of softmax.

{\bf Insights into mechanistic interpretability and  disentangling polysemantic neurons}
By stripping away complex nonlinearities, such as GELU and LayerNorm, AERO makes it easier to dissect how individual neurons, attention mechanisms, and linear FFNs (in Softmax-only architecture) contribute to the model's internal behavior, facilitating a more {\em granular} understanding of LLM internal dynamics. In standard transformer architectures, nonlinearities like GELU induce complex feature interactions, often resulting in polysemantic neurons---neurons that encode overlapping or unrelated information \cite{gurnee2023finding,mu2020compositional}. By simplifying the FFN to a purely linear structure, AERO could minimizes feature entanglement, fosters monosemantic neuron behavior, and produces a more disentangled and interpretable neuron space \cite{pearce2024weightbased}



\end{document}